\documentclass[a4paper, abstracton, bibliography=totoc, numbers=noendperiod]{scrartcl}

\usepackage[utf8]{inputenc}

\usepackage[table]{xcolor}
\usepackage{amsmath}
\usepackage{amssymb}
\usepackage{mathtools}
\usepackage{tikz}
\usepackage{amsthm}
\usepackage{pgfplots}
\usepackage{abstract}
\usepackage{floatrow}
\usepackage{subfig}
\usepackage{natbib}
\usepackage{tabularx, booktabs}
\usepackage{scrextend}
\usepackage{url}
\usepackage{arydshln}
\usepackage{relsize}
\usepackage{multirow,bigdelim}
\usepackage{mdframed}
\usepackage{bbm}
\usepackage{thmtools}
\usepackage{xifthen}
\usepackage{adjustbox}
\usepackage[thinlines]{easytable}
\usepackage{calc}
\usepackage[pass]{geometry}
\usepackage{varwidth}
\usepackage{pdfpages}
\usepackage[ruled, vlined, linesnumbered]{algorithm2e}
\usepackage[toc]{appendix}
\usepackage[subfigure]{tocloft}
\tocloftpagestyle{empty}

\usepackage[font=small, labelfont=bf,format=plain]{caption}
\usepackage[color=orange!60]{todonotes}
\usepackage{lipsum}

\usepackage{listings}
\usepackage{color}
\usepackage{hhline}
\usepackage[pdftex,breaklinks=true]{hyperref}
\usepackage{breakcites}

\definecolor{confred}{rgb}{1,0.48,0.48}
\definecolor{confgreen}{rgb}{0.6,0.9,0.6}

\definecolor{dkgreen}{rgb}{0,0.6,0}
\definecolor{gray}{rgb}{0.5,0.5,0.5}
\definecolor{mauve}{rgb}{0.58,0,0.82}

\definecolor{verbgray}{gray}{0.94}

\lstnewenvironment{code}{%
  \lstset{backgroundcolor=\color{verbgray},
  frame=single,
  framerule=0pt,
  basicstyle=\ttfamily,
  columns=fullflexible}}{}

\definecolor{shadecolor}{rgb}{.9, .9, .9}

\usetikzlibrary{positioning,fit,calc,arrows,shapes,decorations.pathmorphing,patterns}

\makeatletter
\pgfarrowsdeclare{X}{X}
{
  \pgfutil@tempdima=0.3pt%
  \advance\pgfutil@tempdima by.25\pgflinewidth%
  \pgfutil@tempdimb=5.5\pgfutil@tempdima\advance\pgfutil@tempdimb by.5\pgflinewidth%
  \pgfarrowsleftextend{+-\pgfutil@tempdimb}
  \pgfarrowsrightextend{0pt}
}
{ 
  \pgfutil@tempdima=0.3pt%
  \advance\pgfutil@tempdima by.25\pgflinewidth%
  \pgfsetdash{}{+0pt}
  \pgfsetroundcap
  \pgfsetmiterjoin
  \pgfpathmoveto{\pgfqpoint{-5.5\pgfutil@tempdima}{-6\pgfutil@tempdima}}
  \pgfpathlineto{\pgfqpoint{5.5\pgfutil@tempdima}{6\pgfutil@tempdima}}
  \pgfpathmoveto{\pgfqpoint{-5.5\pgfutil@tempdima}{6\pgfutil@tempdima}}
  \pgfpathlineto{\pgfqpoint{5.5\pgfutil@tempdima}{-6\pgfutil@tempdima}}
  \pgfusepathqstroke 
}
\makeatother

\newtheoremstyle{dotless}{}{}{}{}{\bfseries}{}{10pt}{}
\theoremstyle{dotless}

\declaretheorem[qed=$\triangle$,numberwithin=section]{definition}
\declaretheorem[qed=$\triangle$,sibling=definition]{example}

\newcommand\restr[2]{\ensuremath{#1|_{#2}}}

\newcommand{\ch}[2][]{
  \ifthenelse{\isempty{#1}}%
    {\ensuremath{\text{ch}({#2})}}
    {\ensuremath{\text{ch}_{#1}({#2})}}
}

\newcommand{\pa}[2][]{%
  \ifthenelse{\isempty{#1}}%
    {\ensuremath{\text{pa}({#2})}}
    {\ensuremath{\text{pa}_{#1}({#2})}}
}

\makeatletter
\renewenvironment{abstract}{%
    \if@twocolumn
      \section*{\abstractname}%
    \else 
      \begin{center}%
        {\bfseries \sffamily \Large\abstractname\vspace{\z@}}
      \end{center}%
      \quotation
    \fi}
    {\if@twocolumn\else\endquotation\fi}
\makeatother

\newcommand\anno[1]{\textsf{\smaller{#1}}}

\newcommand\pfun{\mathrel{\ooalign{\hfil$\mapstochar\mkern5mu$\hfil\cr$\to$\cr}}}

\newcommand{\ct}{\ensuremath{C_{t{\normalfont\text{AMR}}}}} 
\newcommand{\cs}{\ensuremath{c_{s{\normalfont\text{AMR}}}}} 
\newcommand{\cf}{\ensuremath{c_{f{\normalfont\text{AMR}}}}} 

\newcommand\mydots{\hbox to 1em{.\hss.\hss.}}
\newcommand\squarebox[1]{\parbox[c][1.2cm]{0.9cm}{\centering {#1}}}

\DeclareMathOperator*{\argmin}{arg\,min}
\DeclareMathOperator*{\argmax}{arg\,max}
\DeclareMathOperator*{\argop}{arg\,op}

\def\tbarraystretch{1.15} 
\def\tblinepad{1.5mm} 
\def\tbcolumnpad{3mm} 

\definecolor{plot1}{RGB}{55,126,184}
\definecolor{plot2}{RGB}{228,26,28}

\newcolumntype{Z}{>{\raggedright\let\newline\\\arraybackslash\hspace{0pt}}X}

\hyperbaseurl{http://}
\hypersetup{extension = }

\begin{document}

\title{Transition-Based Generation from \\ Abstract Meaning Representations\footnote{This is a slightly modified version of my Master's thesis ``Transition-Based Generation from Abstract Meaning Representations'' at Technische Universität Dresden.}}
\author{Timo Schick}
\date{}

\maketitle

\begin{abstract}

\noindent This work addresses the task of generating English sentences from Abstract Meaning Representation (AMR) graphs. To cope with this task, we transform each input AMR graph into a structure similar to a dependency tree and annotate it with syntactic information by applying various predefined actions to it. Subsequently, a sentence is obtained from this tree structure by visiting its nodes in a specific order. We train maximum entropy models to estimate the probability of each individual action and devise an algorithm that efficiently approximates the best sequence of actions to be applied. Using a substandard language model, our generator achieves a Bleu score of 27.4 on the LDC2014T12 test set, the best result reported so far without using silver standard annotations from another corpus as additional training data.

\end{abstract}

\section{Introduction}
\label{INTRODUCTION}

Semantic representations of natural language are of great interest for various aspects of natural language processing (NLP). For example, semantic representations may be useful for challenging tasks such as information extraction \citep{palmer2005propbank}, question answering \citep{shen2007using}, natural language generation \citep{langkilde1998generation} and machine translation \citep{jones2012semantics}.


To provide a coherent framework for semantic representations, \citet{banarescu2013abstract} introduced \emph{Abstract Meaning Representation} (AMR), a semantic representation language that encodes the meanings of natural language sentences as directed acyclic graphs with labels assigned to both vertices and edges. Within this formalism, vertices represent so-called \emph{concepts} and edges encode \emph{relations} between them. As AMR abstracts away various kinds of information, each graph typically corresponds to not just one, but a number of different sentences. An exemplary AMR graph
can be seen in Figure~\ref{fig:initial-amr-a}; several sentences corresponding to this graph are listed in Figure~\ref{fig:initial-amr-b}.  
For AMR to be useful in solving the above-mentioned tasks, one must of course be able to convert sentences into AMR graphs and vice versa. Therefore, two important domain-specific problems are \emph{(text-to-AMR) parsing}, the task of finding the graph corresponding~to~a given natural language sentence, and \emph{(AMR-to-text) generation}, the inverse task of finding a good natural language realization for a given AMR graph. To give a simple example of how solutions to these tasks may be beneficial for NLP, a parser and a generator can easily be combined into a machine translation system \citep{jones2012semantics}.

While many approaches have been proposed for the text-to-AMR parsing task \citep[see][]{flanigan2014discriminative,peng2015synchronous,pust2015using,wang2015transition,puzikov2016m2l,zhou2016amr,buys2017robust,van2017neural,konstas2017neural}, the number of currently published AMR-to-text generators is comparably small \citep[see][]{flanigan2016generation,pourdamghani2016generating,song2016amr,song2017amr,konstas2017neural}. 

In this work, we tackle the problem of natural language generation from AMR by successively transforming input AMR graphs into structures that resemble dependency trees. To this end, we define a set of actions (\emph{transitions}) such as the deletion, merging and swapping of edges and vertices. After applying these transitions to the input, we turn the obtained tree structure into a sentence by visiting its vertices in a specific order.
We embed the different kinds of required actions into a \emph{transition system}, a formal framework that, in the context of NLP, is often used for dependency parsing \citep[see][]{nivre2008algorithms}.
To predict the correct sequence of transitions to be applied for each input, we train maximum entropy models \citep{berger1996maximum} from a corpus of AMR graphs and corresponding realizations. As is done in all previous works on this topic, we restrict ourselves to generating English sentences; we do so simply because no reasonably large corpus for any other natural language is available to date. However, we are confident that our results can be transferred to many other languages with some effort.

Our transition-based approach is to a large extent inspired by the likewise transition-based parser CAMR \citep{wang2015transition}. In fact, this parser may be seen as the direct inverse of our system: While we turn AMR graphs into ordered trees which, in turn, are converted into sentences, the parser by \citet{wang2015transition} generates dependency trees from sentences and subsequently transforms these trees into AMR graphs. Accordingly, several transitions used by CAMR have a direct counterpart in our generator. 

In a way, the task performed by our system is simpler than its inverse. This is because we are not required to transform input AMR graphs into actual dependency trees; any tree is sufficient as long as the sentence obtained from it is a good realization of the input. For this very reason, there is also no need for us to assign dependency labels as they have no representation in the generated sentence.
In other respects, however, the transformation from AMR graphs to suitable trees is much more challenging than going the opposite way. For example, we have to somehow cope with the fact that AMR graphs, in contrast to dependency trees, are unordered. Furthermore, AMR abstracts away tense, number and voice as well as function words such as articles, pronouns and prepositions; all this information must somehow be retrieved. 
Finally, the inclusion of a language model into our generation pipeline -- which is indispensable to obtain competitive results -- makes it very difficult to efficiently determine the best sequence of transitions for a given input.

We address these challenges in various ways. For instance, we devise a set of special transitions to establish an order on the vertices of our input. We try to compensate for lacking syntactic information by training several maximum entropy models to estimate this very information; this idea is formalized by introducing the concept of \emph{syntactic annotations}. To actually implement our system, we develop a novel generation algorithm that incorporates a language model but is still sufficiently efficient.

We proceed as follows: After giving a succinct overview of previous work on AMR-to-text generation and related tasks in Section~\ref{RELATEDWORK}, we discuss basic notation and other preliminaries such as the AMR formalism, transition systems and maximum entropy models in Section~\ref{PRELIMINARIES}. We introduce our generator in Section~\ref{GENERATION}, which constitutes the core of this work. This section includes a detailed definition of all required transitions as well as a thorough derivation of our generation algorithm and an explanation of the required training procedure. In Section~\ref{IMPLEMENTATION}, we discuss our Java-based implementation of the generator. Results obtained with this implementation are reported in Section~\ref{EXPERIMENTS}; for a quick overview on the performance of our generator and a comparison with all other currently published approaches, we refer to Table~\ref{tab:comparison} of Section~\ref{EXPERIMENTS}. We conclude with a concise summary of our work and an outlook on future research topics in Section~\ref{CONCLUSION}.

\begin{figure}[t]
\centering
\subfloat[]{
\scalebox{0.8} {
\begin{tikzpicture}
\tikzstyle{amr-node}=[shape=ellipse,draw, inner sep=0.2, minimum width=1.5cm, minimum height=0.8cm, text height=1.5ex, text depth=.25ex]
\tikzstyle{dep-node}=[shape=ellipse,draw, inner sep=0.2, minimum height=0.8cm, minimum width=1.5cm, text height=1.5ex, text depth=.25ex]
\tikzstyle{text-node}=[text height=1.5ex, text depth=.25ex, align=left]
\tikzstyle{amr-align}=[-latex,color=red]
\tikzstyle{dep-align}=[-latex,color=red]
\tikzstyle{frame}=[draw, dashed, inner sep=0.3cm]

    \node[amr-node] (amr-possible) at (0,0) {possible};
    \node[amr-node] (amr-close) at (-3.6,-1.3) {close-01};
    \node[amr-node] (amr-boy) at (-4,-3.6) {boy};
    \node[amr-node] (amr-eye) at (-0.25,-2.6) {eye};
    \node[amr-node] (amr-minus) at (2,-1.7) {$-$};

    \path [-latex](amr-possible) edge node[fill=white, inner sep=0.5mm] {domain} (amr-close);
    \path [-latex](amr-possible) edge node[fill=white] {polarity} (amr-minus); 
    \path [-latex](amr-close) edge node[fill=white] {ARG0} (amr-boy); 
    \path [-latex](amr-close) edge node[fill=white] {ARG1} (amr-eye);
    \path [-latex](amr-eye) edge node[fill=white] {part-of} (amr-boy);       
\end{tikzpicture}
}
\label{fig:initial-amr-a}
}
\subfloat[]{
\scalebox{0.8} {
\begin{tikzpicture}[]
\tikzstyle{text-node}=[text height=1.5ex, text depth=.25ex, align=left]

    \node[text-node] (textbox) {
    \large{
        \begin{varwidth}{7.9cm}\begin{itemize}
        			\item It is not possible for the boy to close his eyes.
                    \item The boy is unable to close his own eyes.
                    \item The boys couldn't close their eyes.
                    \item There was no possibility for the boy to close his eyes.
                \end{itemize}\end{varwidth}
     }};
     \node[below = 1.7cm of textbox] (PAD) { };
\end{tikzpicture}
}
\label{fig:initial-amr-b}
}
\caption{Visualization of an AMR graph and corresponding sentences}
\label{fig:initial-amr}
\end{figure}

\clearpage


\section{Related Work}
\label{RELATEDWORK}

In this section, we give a short overview of previous work on AMR-related tasks, but we restrict ourselves to only such work that is closely related to the generation of natural language sentences from AMR. For a general introduction to AMR, we refer to Section~\ref{PRELIMINARIES:AMR} of this work and to \citet{banarescu2013abstract}.

\paragraph{Alignments} Both generation and parsing methods are often trained using an \emph{AMR corpus}, a large set of AMR graphs and corresponding reference sentences. For such training procedures, it is useful to somehow link vertices of each AMR graph $G$ to corresponding words of its reference sentence $s$. These links are commonly referred to as an \emph{alignment}; several methods have been proposed for automatically generating such alignments. 

The methods described by \citet{jones2012semantics} and \citet{pourdamghani2014aligning} both bijectively convert an AMR graph $G$ into a string $s_G$ through a simple breadth first search and depth first search, respectively.\footnote{The aligner by \citet{pourdamghani2014aligning} is available at \url{isi.edu/~damghani/papers/Aligner.zip}; the aligner by \citet{jones2012semantics} is not publicly available.} Then, a string-to-string alignment between $s_G$ and $s$ is obtained using one of the models described in \citet{brown1993mathematics}; these models originate from the field of machine translation and are commonly referred to as \emph{IBM Models}. The obtained alignment can then easily be converted into the desired format by retransforming $s_G$ into $G$. 

A fundamentally different approach is proposed by \citet{flanigan2014discriminative}, where a set of alignment rules is defined by hand; these rules are then greedily applied in a specified order.\footnote{The aligner by \citet{flanigan2016generation} is available at \url{github.com/jflanigan/jamr}.} An example of such a rule is the \emph{Minus Polarity Tokens} rule, which aligns the words ``no'', ``not'' and ``non'' to vertices with the label ``$-$''; this label is used in AMR to indicate negative polarity. The set of all rules used by this rule-based aligner can be found in~\citet{flanigan2014discriminative}.

\paragraph{Parsing} Many approaches for parsing English sentences into AMR graphs have been proposed. However, as the subject of this work is generation, we consider here only the transition-based parser CAMR introduced by \citet{wang2015transition}.\footnote{The CAMR parser by \citet{wang2015transition} is available at \url{github.com/c-amr/camr}.} We consider this specific parser because several of its transitions are either equal or inverse to the transitions used by our generator. The idea behind CAMR is to make use of the fact that AMR graphs and dependency trees share some structural similarities. Therefore, given a sentence $s$, CAMR relies on some dependency parser to first generate the dependency tree $D_s$ corresponding to $s$. Subsequently, several transitions are applied to $D_s$ in order to successively turn it into the desired AMR graph $G$. These transitions include, for example, deleting and renaming both vertices and edges, swapping vertices or merging them into a single one as well as adding new edges. After each application of a transition, the transition to be applied next is determined using a linear classifier which, in turn, is trained with the aid of the alignment method described in \citet{flanigan2014discriminative}.    

\paragraph{Generation} The first system for generating English strings from AMR graphs was published by \citet{flanigan2016generation}.\footnote{The generator by \citet{flanigan2016generation} is available at \url{github.com/jflanigan/jamr/tree/Generator}.} The core idea of this system is to convert AMR graphs into trees and to train a special kind of tree-to-string transducer \citep[see][]{huang2006statistical} on these trees. To obtain rules for the transducer, the greedy rule-based aligner of \citet{flanigan2014discriminative} is used and several rule extraction mechanisms are tried out. 
An obvious problem with this approach is that the conversion of an AMR graph into a tree in general requires us to remove edges from it; the information encoded by these edges is therefore lost.

\citet{song2016amr} treat AMR generation as a variant of the traveling salesman problem (TSP).\footnote{The generator by \citet{song2016amr} is available at \url{github.com/xiaochang13/AMR-generation}.} Input AMR graphs are first partitioned into several disjoint subgraphs and for each subgraph, a corresponding English phrase is determined using a set of rules extracted from a training set. Afterwards, an order among all subgraphs is specified. To this end, a \emph{traveling cost} for visiting one subgraph after another is learned and the cost of each order is set to the sum of all traveling costs of adjacent subgraphs. For the final output, the order with the lowest score is determined using a TSP solver and the extracted phrases are concatenated in this very order.

The core idea of \citet{pourdamghani2016generating} is to convert AMR graphs into strings, a process referred to as \emph{linearization}, and then train a string-to-string translation model on the so-obtained pairs of linearized AMR graphs and corresponding sentences. For the linearization process, a simple depth first search is performed. However, since there is no order among vertices of an AMR graph, siblings can be visited in any order. As it may be helpful for the string-to-string translation model if the linearized AMR graph resembles English word order, a linear classifier is trained to decide for each pair of sibling vertices $(v_1, v_2)$ whether $v_1$ should be visited before $v_2$ or vice versa. The actual string-to-string translation is then performed using a phrase-based model implemented in \emph{Moses} \citep{koehn2007moses}. 

Another approach that requires AMR graphs to be linearized is proposed by \citet{konstas2017neural}. Their generator uses a sequence-to-sequence model built upon a \emph{long short-term memory} (LSTM) neural network architecture. As this architecture requires a large set of training data to achieve good results, \citet{konstas2017neural} use a text-to-AMR parser to automatically annotate millions of unlabeled sentences before training their system; the so-obtained AMR graphs are then used as additional training data.

Yet another approach is to tackle the problem of AMR generation using \emph{synchronous node replacement grammars} \citep{song2017amr}. A synchronous node replacement grammar is a rewriting formalism primarily defined by a set of rules that simultaneously produce graph fragments and phrases. Through repeated application of such rules, AMR graphs and corresponding sentences can be obtained; a sequence of rule applications is called a \emph{derivation}. Given an AMR graph $G$, the approach of \citet{song2017amr} is to assign scores to all possible derivations which produce $G$ and to take the sentence produced by the highest-scoring such derivation as the output of the generator.


\clearpage


\section{Preliminaries}
\label{PRELIMINARIES}

\subsection{Basic Notation}

\paragraph{Set theory} Let $A$ and $B$ be sets. We write $a \in A$ if an object $a$ is an element of $A$. The cardinality of $A$ is denoted by $|A|$. If $A$ is a subset of $B$, we write $A \subseteq B$ and $A \subset B$ if $A \neq B$. The Cartesian product of $A$ and $B$, their union, intersection and difference are written $A \times B$, $A \cup B$, $A \cap B$ and $A \setminus B$, respectively. For $n \in \mathbb{N}$, the $n$-fold Cartesian product of $A$ with itself is written~$A^n$. The power set of $A$ is denoted by $\mathcal{P}(A)$. We denote the empty set as $\emptyset$, the set $\{0, 1, 2, \ldots \}$ of natural numbers as $\mathbb{N}$ and $\mathbb{N}\setminus\{0\}$ as $\mathbb{N}^+$. In an analogous manner, we write the set of integers as $\mathbb{Z}$, the set of real numbers as $\mathbb{R}$, the set of nonnegative reals as $\mathbb{R}^+_0$ and the set of positive reals as $\mathbb{R}^+$. For $n \in \mathbb{N}$, $[n]$ denotes the set $\{1, 2, \ldots, n\}$  and  $[n]_0$ denotes $[n] \cup \{0\}$. 

\paragraph{Binary relations} Let $A$, $B$ and $C$ be sets. A \emph{binary relation between $A$ and $B$} is a set $R \subseteq A \times B$. If $A=B$, we call $R$ a \emph{binary relation on $A$}. We sometimes denote $(a,b) \in R$ as $a\, R \, b$. The \emph{inverse} of a relation $R \subseteq A \times B$, denoted by $R^{-1}$, is the relation $\{ (b,a) \mid (a,b) \in R \} \subseteq B \times A$. The \emph{domain of $R$} is the set $\text{dom}(R) = \{ a \in A \mid \exists b \in B : (a,b) \in R \}$. For relations $R_1 \subseteq A \times B$ and $R_2 \subseteq B \times C$, their \emph{composition} is defined as 
\[R_1R_2 = \{ (a,c) \in A \times C \mid \exists b \in B : (a,b) \in R_1 \wedge (b,c) \in R_2 \}\,.\]

In the following, let $R$ be a binary relation on $A$ and let $A' \subseteq A$. $R$ is called \emph{irreflexive} if for all $a \in A$, $(a,a) \notin R$ and \emph{transitive} if for all $a,b,c\in A$, $(a,b) \in R \wedge (b,c) \in R \Rightarrow (a,c) \in R$. The \emph{transitive closure} of $R$, denoted by $R^+$, is the smallest relation on $A$ such that $R \subseteq R^+$ and $R^+$ is transitive. We call a relation that is both irreflexive and transitive a \emph{strict order}. $R$ is a \emph{total order on $A'$} if $R$ is a strict order and for all $a, b \in A'$, $(a,b) \in R$ or $(b,a) \in R$. If $A'$ is a finite set with $n$ elements and $R$ is a total order on $A'$, the \emph{$A'$-sequence induced by $R$} is the uniquely determined sequence $(a_1, \ldots, a_n)$ where for all $i \in [n-1]$, $(a_i, a_{i+1}) \in R \cap A' \times A'$.

\paragraph{Functions} Let $A$, $B$ and $C$ be sets. We call a binary relation $f$ between $A$ and $B$ a \emph{partial function from $A$ to $B$} and write $f: A \pfun B$ if for all $a \in A$, there is at most one $b \in B$ such that $(a,b) \in f$; we also denote $b$ by $f(a)$. If $\text{dom}(f) = A$, we call $f$ a \emph{(total) function} and write $f: A \rightarrow B$. We call $f : A \rightarrow B$ a \emph{bijective} function or \emph{bijection} if for all $b \in B$, there is exactly one $a \in A $ such that $f(a) = b$. For $f: A \pfun B$, $a \in A$ and $b \in B$, the function $f[a \mapsto b] : \text{dom}(f) \cup \{ a \} \rightarrow B$ is defined by 
\[
f[a \mapsto b] (x) = 
\begin{cases} 
b & \text{if } x = a \\
f(x) & \text{otherwise}
\end{cases}
\] 
for all $x \in \text{dom}(f) \cup \{ a \}$. Let $f: A \pfun B$, $a_1, \ldots, a_n \in A$, $b_1, \ldots, b_n \in B$, $n \in \mathbb{N}$. We write $f[a_1 \mapsto b_1, \ldots, a_n \mapsto b_n]$ as a shorthand for $(\ldots(f[a_1 \mapsto b_1]) \ldots ) [a_n \mapsto b_n]$. For $f: A \pfun (B \pfun C)$,  $a_1, \ldots, a_n \in A$, $b_1, \ldots, b_n \in B$, $c_1, \ldots, c_n \in C$, we write 
\[
f[a_1(b_1) \mapsto c_1, \ldots, a_n(b_n) \mapsto c_n]
\] 
as a shorthand for $
f[a_1 \mapsto f(a_1)[b_1 \mapsto c_1], \ldots, a_n \mapsto f(a_n)[b_n \mapsto c_n]]
$.

For $g\colon A \rightarrow \mathbb{R}$ and $\text{op} \in \{ \text{min}, \text{max} \}$, $\argop_{x \in A} g(x)$ usually denotes the set
\[
S_\text{op} = \{ x \in A \mid \nexists x' \in A \colon g(x') \mathbin{\Diamond} g(x) \} \text{ where } \Diamond = \begin{cases} {>} & \text{if } \text{op} = \text{max} \\ < & \text{if } \text{op} = \text{min}\,. \end{cases}
\]
However, we are often just interested in one arbitrary $x \in S_\text{op}$. We therefore identify $\argop_{x \in A}g(x)$ with some element of the set $S_\text{op}$ for the rest of this work.

\paragraph{Formal languages} An \emph{alphabet} $\Sigma$ is a nonempty set of distinguishable symbols.\footnote{While alphabets are commonly defined as \emph{finite} sets, we explicitly allow them to be of infinite size.} A \emph{string} over $\Sigma$ is a finite sequence of symbols from $\Sigma$; $\Sigma^*$ denotes the set of all such strings. The concatenation of two strings $a, b \in \Sigma^*$ is written $a \cdot b$ or $ab$. We abbreviate the $n$-fold concatenation of the same symbol $a \in \Sigma$ by $a^n$. 
Let $w = (w_1, \ldots, w_n)$ be a string over some alphabet $\Sigma$ with $w_i \in \Sigma$ for all $i \in [n]$. We denote $w_i$ also by $w(i)$. We sometimes write $w_1 \ldots w_n$ as an abbreviation for $(w_1, \ldots, w_n)$. If we are only interested in the first $m \leq n$ symbols of $w$, we also denote $w$ as $w_1{:}w_2{:}\ldots{:}w_m{:}w'$ with $w' = (w_{m+1}, \ldots, w_m)$. The length of $w$ is written $|w|$, $\varepsilon$ denotes the empty string. For $\Sigma' \subseteq \Sigma$, we define $w \setminus \Sigma'$ to be the sequence $w_1' \cdot \ldots \cdot w_n'$ with
\[
w_i' = \begin{cases}
w_i & \text{if } w_i \notin \Sigma' \\
\varepsilon & \text{otherwise}
\end{cases}
\]
for all $i \in [n]$, i.e. $w \setminus \Sigma'$ is obtained from $w$ by removing from it all $w_i \in \Sigma'$.

An alphabet frequently used throughout this work is the set of all English words, hereafter denoted by $\Sigma_\text{E}$. We define $\Sigma_\text{E}$ to contain not only all English words and word forms, but also punctuation marks, numbers and special characters. Notwithstanding the above definitions, we always separate symbols from $\Sigma_\text{E}$ by spaces. That is, we write ``$\text{the house}$'' rather than ``$(\text{the},\text{house})$'' or ``$\text{the}\cdot\text{house}$''. \label{los:sigma_e}

\paragraph{Probability theory} 

Let $\Omega$ be a countable set. A \emph{probability measure on $\Omega$} is a function $P\colon \mathcal{P}(\Omega) \rightarrow [0,1]$ such that $P(\Omega) = 1$ and 
\[
P\left( \bigcup_{i=1}^{\infty} A_i \right) = \sum_{i=1}^\infty P(A_i)
\]
for every countable sequence $A_1, A_2, \ldots$ of pairwise disjoint sets $A_i \subseteq \Omega$ (i.e. $A_i \cap A_j = \emptyset$ for all $i, j \in \mathbb{N}$ with $i \neq j $). For $\omega \in \Omega$ and $A, B \subseteq \Omega$, we abbreviate $P(\{\omega \})$ by $P(\omega)$ and $P(A \cap B)$ by $P(A,B)$.

Let $A, B \subseteq \Omega$. For $P(B) \neq 0$, the \emph{conditional probability of A given B} is defined as
\[
P(A \mid B) = P(A,B) \cdot P(B)^{-1}\,.
\] For some $C \subseteq \Omega$ with $P(C) \neq 0$, we say that $A$ and $B$ are \emph{conditionally independent given $C$} if 
$P(A, B \mid C) = P(A \mid C) \cdot P(B \mid C)$.
Let $n \in \mathbb{N}$, $A_i \subseteq \Omega$ for $i \in [n]$ and $(B_i \mid i \in I)$ be a countable partition of $\Omega$. We will make frequent use of the following two identities: 
\begin{align*}
P(A_1, \ldots, A_n) & = P(A_1, \ldots, A_{n-1}) \cdot P(A_n \mid A_1, \ldots, A_{n-1}) & & & \text{(General product rule)} \\
P(A) & = \sum\nolimits_{i \in I} P(A, B_i) & & & \text{(Law of total probability)}
\end{align*}

Let $X$ be a countable set. A \emph{random variable} is a function $\mathbb{X} \colon \Omega \rightarrow X$. For $x \in X$, we use $\mathbb{X} = x$ as an abbreviation for the set $\{ \omega \in \Omega \mid \mathbb{X}(\omega) = x \}$. Thus, \[
P(\mathbb{X} = x) = \sum_{\omega \in \Omega \colon \mathbb{X}(\omega) = x} P(\omega)\,.
\]Throughout this work, we drop random variables $\mathbb{X}$ from our notation whenever they are clear from the context, i.e. we simply write $P(x)$ instead of $P(\mathbb{X} = x)$.

Let $X$ and $Y$ be countable sets. A \emph{probability distribution of $X$} is a function $p \colon X \rightarrow [0,1]$ such that $\sum_{x \in X} p(x) = 1$. A \emph{conditional probability distribution of $X$ given $Y$} is a function $q \colon Y \rightarrow (X \rightarrow [0,1])$ such that for all $y \in Y$, $\sum_{x \in X} q(z)(x) = 1$. We denote $q(z)(x)$ also by $q(x \mid z)$.

\subsection{Labeled Ordered Graphs}

\begin{definition}[Labeled ordered graph]
\label{def:graph}
Let $L_E$ and $L_V$ be two sets (\emph{edge labels} and \emph{vertex labels}). A \emph{(labeled ordered) $(L_E,L_V)$-graph} is a tuple $G = (V,E,L,\prec)$ where $V \neq \emptyset$ is a finite set of \emph{vertices} (or \emph{nodes}), $E \subseteq V \times L_E \times V$ is a finite set of labeled \emph{edges}, $L: V \rightarrow L_V$ is a \emph{vertex labeling} and ${\prec} \subseteq V \times V$ is a strict order.
\end{definition}

If we are not interested in the particular sets of edge and vertex labels, we refer to a $(L_E,L_V)$-graph simply as \emph{graph}.
In the following, let $G = (V,E,L,\prec)$ be a graph. For each $v \in V$, $L(v)$ is called the \emph{label of $v$} and for each $e = (v_1, l, v_2) \in E$, $l$ is called the \emph{label of $e$}. We define a \emph{walk in $G$} to be a sequence of vertices $w = (v_0, \ldots, v_n)$, $n \in \mathbb{N}^+$ such that for all $i \in [n]$, there is some $l_i \in L_E$ with $(v_{i-1},l_i, v_{i}) \in E$. A \emph{cycle} is a walk $(v_0, \ldots, v_n)$ where $v_0 = v_n$ and $v_i \neq v_j$ for all other $i,j \in [n]_0$ with $i \neq j$. We call $G$ \emph{cyclic} if it contains at least one cycle and \emph{acyclic} otherwise. For each node $v \in V$, we denote by 
\begin{align*}
\text{in}_G(v) & = \{ e \in E \mid \exists v' \in V, l \in L_E : e = (v', l, v) \} \\
\text{out}_G(v) & = \{ e \in E \mid \exists v' \in V, l \in L_E : e = (v, l, v') \}
\end{align*}
the set of its \emph{incoming edges} and \emph{outgoing edges}, respectively. Correspondingly,
\begin{align*}
\pa[G]{v} & = \{v' \in V \mid \exists l \in L_E : (v',l,v) \in E \} \\
\ch[G]{v} & = \{v' \in V \mid \exists l \in L_E : (v,l,v') \in E\}
\end{align*} 
denote the set of $v$'s \emph{parents} and \emph{children}. If $G$ is acyclic, the sets of \emph{successors} and \emph{predecessors} of $v$ are defined recursively as 
\begin{align*}
\text{succ}_G(v) = \ch[G]{v} \cup \bigcup_{v' \in \ch[G]{v}} \text{succ}_G(v') \qquad
\text{pred}_G(v) = \pa[G]{v} \cup \bigcup_{v' \in \pa[G]{v}} \text{pred}_G(v')\,.
\end{align*}
From the above notations, we sometimes drop the subscript if the corresponding graph is clear from the context; for example, we often simply write $\pa{v}$ and $\ch{v}$ instead of $\pa[G]{v}$ and $\ch[G]{v}$. 
We call $v \in V$ a \emph{root of $G$} if $\pa[G]{v} = \emptyset$. If $V$ contains exactly one root, $G$ is called a \emph{rooted} graph; we denote this vertex by $\text{root}(G)$. $G$ is called a \emph{tree} if it is rooted, acyclic and $|\text{in}_G(v)| = 1$ for all $v \in V \setminus \{ \text{root}(G) \}$. We say that $G$ is \emph{totally ordered} if for all $v \in V$, $\prec$ is a total order on $\ch[G]{v} \cup \{v\}$. 

Throughout this work, we often represent a graph $G = (V, E, L, \prec)$ graphically. In such a visualization, each vertex $v \in V$ is represented by an ellipse inscribed either with $L(v)$ or $v\,{:}\,L(v)$. Each edge $(v_1, l, v_2) \in E$ is represented by an arrow line connecting the graphical representations of $v_1$ and $v_2$; this line is inscribed with $l$. We do not depict the order $\prec$ in this visualization, but whenever $\prec$ is of relevance, we explicitly specify it. 

\begin{definition}[$v$-Subgraph]
Let $G = (V,E,L,\prec)$ be an acyclic graph. For $v \in V$, the \emph{$v$-subgraph of $G$}, denoted by $\restr{G}{v}$, is the graph $(V', E', L', \prec')$ where
\begin{align*}
V' &= \text{succ}(v) \cup \{v\} & E' &= \{ (v_1, l, v_2) \in E \mid v_1, v_2 \in V' \} \\
L' &= \{ (v, l) \in L  \mid v \in V' \} & \prec'\, &= \{ (v_1, v_2) \in\, \prec\, \mid v_1, v_2 \in V' \} \text{\,.} \qedhere
\end{align*}
\end{definition}

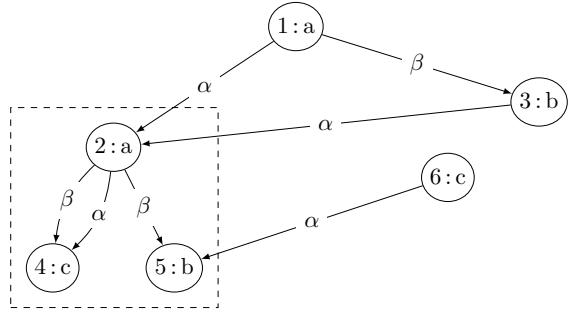
\begin{figure}[t]
\centering
\scalebox{0.8} {
\begin{tikzpicture}
\tikzstyle{amr-node}=[shape=ellipse,draw, inner sep=0.2, minimum height=0.8cm, text height=1.5ex, text depth=.25ex]

    \node[amr-node] (1) at (0,0) {1\,:\,a};
    \node[amr-node] (2) at (-3,-2) {2\,:\,a};
    \node[amr-node] (3) at (4,-1.25) {3\,:\,b};
    \node[amr-node] (4) at (-4,-4) {4\,:\,c};
    \node[amr-node] (5) at (-2,-4) {5\,:\,b};
    \node[amr-node] (6) at (2.5,-2.5) {6\,:\,c};
    
    \node[draw, dashed, fit=(2)(4)(5), inner sep=0.25cm] (subgraph) {};

    \path [-latex](1) edge node[fill=white] {$\alpha$} (2);
    \path [-latex](1) edge node[fill=white] {$\beta$} (3);
    \path [-latex](3) edge node[fill=white] {$\alpha$} (2);
    \path [-latex, bend angle=20, bend left](2) edge node[fill=white] {$\alpha$} (4);
    \path [-latex, bend angle=20, bend right](2) edge node[fill=white] {$\beta$} (4);
    \path [-latex](2) edge node[fill=white] {$\beta$} (5);
    \path [-latex](6) edge node[fill=white] {$\alpha$} (5);

\end{tikzpicture}
}
\caption{Graphical representation of the graph $G_0 = (V_0,E_0,L_0,\prec_0)$ as described in Example~\ref{example:graph}. Each node $v \in V_0$ is inscribed with $v$\,:\,$L_0(v)$. $\restr{G_0}{2}$ is framed by dashed lines.}
\label{fig:graph}
\end{figure}

\begin{example}
\label{example:graph}
Let $L_E = \{ \alpha, \beta \}$ be a set of edge labels and $L_V = \{a, b, c\}$ be a set of vertex labels. The $(L_E, L_V)$-graph $G_0 = (V_0, E_0, L_0, \prec_0)$ where
\begin{align*}
V_0 = &\ \{1,2,3,4,5,6\} \\ 
E_0 = &\ \{ (1, \alpha, 2), (1, \beta, 3), (3, \alpha, 2), (2, \alpha, 4), (2, \beta, 4), (2, \beta, 5), (6, \alpha, 5) \} \\
L_0 = &\ \{(1, a), (2,a), (3, b), (4,c), (5,b), (6,c)\} \\
\prec_0\,= &\ \{(v_1,v_2) \in V_0 \times V_0 \mid v_1 <_\mathbb{N} v_2 \}
\end{align*}
is acyclic and totally ordered, but not rooted. The $2$-subgraph of $G_0$ is the rooted graph $\restr{G_0}{2} = (\{2,4,5\}, \{ (2, \alpha, 4), (2, \beta, 4), (2, \beta, 5) \}, \{ (2,a), (4,c), (5,b) \}, \{ (2,4), (2,5), (4,5) \})$. A graphical representation of both $G_0$ and $\restr{G_0}{2}$ can be found in Figure~\ref{fig:graph}.
\end{example}

\begin{definition}[Yield]
\label{def:yield}
Let $G = (V,E,L,\prec)$ be an acyclic and totally ordered graph. Furthermore, let $\Sigma$ be an alphabet, $V'$ be a set with $V \subseteq V'$ and $\rho: V' \rightarrow \Sigma^*$. The function $\text{yield}_{(G,\rho)}: V \rightarrow \Sigma^*$ is defined for each $v \in V$ as
\[ 
\text{yield}_{(G,\rho)}(v) := \text{yield}_{(G,\rho)}(c_1) \cdot \ldots \cdot \text{yield}_{(G,\rho)}(c_k) \cdot \rho(v) \cdot \text{yield}_{(G,\rho)}(c_{k+1}) \cdot \ldots \cdot \text{yield}_{(G,\rho)}(c_{|\ch{v}|}) 
\]
where $(c_1, \ldots, c_k, v, c_{k+1}, \ldots, c_{|\ch{v}|})$, $k \in [|\ch{v}|]_0$ is the $(\ch{v} \cup \{v\})$-sequence induced by $\prec$. If $G$ is rooted, we write $\text{yield}_\rho(G)$ as a shorthand for $\text{yield}_{(G,\rho)}(\text{root}(G))$.
\end{definition}

Let $G = (V,E,L,\prec)$ and $\rho$ be defined as above. We observe that for all $u, v, w \in V$, if $u$ is a successor of $v$ and the term $\rho(w)$ occurs in $\text{yield}_\rho(G)$ between the terms $\rho(u)$ and $\rho(v)$, then $w$ must also be a successor of $v$; in analogy to a similar property studied in the context of dependency trees \citep[see][]{nivre2008algorithms}, we refer to this property of yield as \emph{projectivity}.

\begin{example}
Let $\Sigma_0 = \{x, y, z\}$ and let $\rho_0 = \{ (1,x), (2,y), (3,x), (4,z), (5,x), (6,y) \}$. We consider the graph $G_0 = (V_0, E_0, L_0, \prec_0)$ defined in Example~\ref{example:graph}. All of the following statements are true:
\begin{align*}
\text{yield}_{(G_0, \rho_0)}(2) & = \rho_0(2) \cdot \rho_0(4) \cdot \rho_0(5) = yzx \\
\text{yield}_{(G_0, \rho_0)}(3) & = \text{yield}_{(G_0, \rho_0)}(2) \cdot \rho_0(3) = yzx\cdot x \\
\text{yield}_{(G_0, \rho_0)}(1) & = \rho_0(1) \cdot \text{yield}_{(G_0, \rho_0)}(2) \cdot \text{yield}_{(G_0, \rho_0)}(3) = x \cdot yzx \cdot yzxx \\
\text{yield}_{(G_0, L_0)}(6) & = L_0(5) \cdot L_0(6) = bc \text{\,.} \qedhere
\end{align*}
\end{example}

\begin{definition}[Bottom-up traversal]
Let $G = (V, E, L, \prec)$ be an acyclic graph. We call a sequence of vertices $s \in V^*$ a \emph{bottom-up traversal of $G$} if there is some total order $\lessdot$ on $V$ such that for all $v \in V$ and $v' \in \ch[G]{v}$ it holds that $v' \lessdot v$ and $s$ is the $V$-sequence induced by $\lessdot$.
\end{definition}

\begin{example}
We consider once more the graph $G_0 = (V_0, E_0, L_0, \prec_0)$ defined in Example~\ref{example:graph}. The sequences
\[
s_1 = (4, 5, 6, 2, 3, 1) \quad s_2 = (4,5,2,3,1,6) \quad s_3 = (5,4,2,6,3,1)
\]
are bottom-up traversals of $G_0$. In contrast, $(4,5,6,3,2,1)$ is not a bottom-up traversal of $G_0$ because the corresponding order $\lessdot = \{ (4,5), (5,6), (6,3), (3,2), (2,1) \}^+$ does not contain the tuple $(2,3)$ although $2 \in \ch[G_0]{3}$.
\end{example}

\subsection{Abstract Meaning Representation}
\label{PRELIMINARIES:AMR}

\emph{Abstract Meaning Representation} (AMR) is a semantic representation language that encodes the meaning of a sentence as a rooted, acyclic graph \citep{banarescu2013abstract}. To this end, AMR makes use of \emph{PropBank framesets} \citep{kingsbury2002propbank, palmer2005propbank}. A PropBank frameset mainly consists of
\begin{enumerate}
\item a \emph{frameset id} (``want-01'', ``see-01'', ``develop-02'', \dots) which in turn consists of a verb and a number; the latter is used to differentiate between several meanings of the same verb and also referred to as the \emph{sense tag} of the frameset id; 
\item a list of associated \emph{semantic roles} (ARG0 -- ARG5). These roles have no intrinsic meaning but are defined on a verb-by-verb basis; for many verbs, only some semantic roles are defined. The meanings of all semantic roles specified for the frameset ids ``want-01'', ``see-01'' and ``develop-02'' can be seen in Table~\ref{tab:propbank}. 
\end{enumerate}

\begin{table}
{
\small
\begin{tabularx}{0.75\textwidth}{ l l l }
  \textbf{want-01} & \textbf{sleep-01} & \textbf{develop-02} \\
  \cmidrule(lr){1-1}\cmidrule(lr){2-2}\cmidrule(lr){3-3}
  ARG0: wanter & ARG0: sleeper & ARG0: creator \\
  ARG1: thing wanted & ARG1: cognate object & ARG1: thing created \\
  ARG2: beneficiary & & ARG2: source \\
  ARG3: in-exchange-for & & ARG3: benefactive \\
  ARG4: from & & \\
\end{tabularx}
}
\caption{PropBank framesets corresponding to the concepts \emph{want-01}, \emph{sleep-01} and \emph{develop-02}, extracted from \protect\url{propbank.github.io}. For each frameset, the specific meanings of the corresponding semantic roles are briefly described.}
\label{tab:propbank}
\end{table}

The key components of an AMR graph are \emph{concepts}, represented by the set of possible vertex labels, \emph{instances} of these concepts, represented by actual vertices, and \emph{relations} between these instances, represented by edges. For example, an edge $e = (v_0, \text{ARG0}, v_1)$ connecting two nodes $v_0$ and $v_1$ with labels ``sleep-01'' and  ``boy'', respectively, would indicate that an instance of the concept ``boy'', i.e. an actual boy, is the zeroth argument of an instance of the frameset ``sleep-01'', or in other words, he is the person who is sleeping. A simple graph consisting only of the nodes $v_0$ and $v_1$ and the edge $e$ can thus be seen as a semantic representation of the phrase ``a boy sleeps''.

The set of all AMR concepts, hereafter denoted by $L_\text{C}$,\label{los:l_c} consists of English words, numbers, names, PropBank framesets and so-called \emph{special keywords}. The latter include logical conjunctions (``and'', ``or'', \mydots), grammatical mood indicators (``interrogative'', ``imperative'', \mydots), polarity (``$-$'', ``$+$''), quantities (``monetary-quantity'', ``distance-quantity'', \mydots) and special entity types (``rate-entity'', ``date-entity'', \mydots). For further details on the meaning of these keywords and a complete list thereof, we refer to AMR Specification~1.2.2.\footnote{AMR Specification~1.2.2 can be found at \url{amr.isi.edu/language.html}.} 

Following \citet{banarescu2013abstract}, we can roughly divide the set of possible relation labels, hereafter denoted by $L_\text{R}$,\label{los:l_r} into five categories:
\begin{enumerate}
\item PropBank semantic roles (ARG0 -- ARG5), also referred to as \emph{core roles};
\item General semantic relations (location, cause, purpose, manner, topic, time, duration, direction, instrument, accompanier, age, frequency, name, \ldots);
\item Relations for quantities (quant, unit, scale, \dots);
\item Relations for date-entities (day, month, year, weekday, century, era, quarter, season, timezone, \dots);
\item Relations for enumerations and listings (OP$i$, $i \in \mathbb{N}$).
\end{enumerate}
For each relation $r$ from this list, the corresponding \emph{inverse relation}, denoted by $r$-of, is also included in $L_\text{R}$; it is sometimes necessary to exchange a relation by its inverse in order to make the corresponding AMR graph rooted. We define for all $r \in L_\text{R}$:
\[
r^{-1} = 
\begin{cases}
r' & \text{if } r = r'\text{-of} \text{ for some } r' \in L_\text{R} \\
r\text{-of} & \text{otherwise.}
\end{cases}
\]
To give an example, $\text{ARG0}^{-1}$ equals $\text{ARG0-of}$ and $\text{purpose-of}^{-1}$ equals $\text{purpose}$.
For a complete list of all possible relation labels, we again refer to AMR Specification~1.2.2.

\begin{definition}[AMR graph] \label{los:g_amr}
An AMR graph is a rooted, acyclic $(L_\text{R}, L_\text{C})$-graph $G = (V,E,L, \prec)$ with $\prec\, = \emptyset$.\footnote{Note that this definition differs slightly from the format introduced by \citet{banarescu2013abstract} where only leaf nodes have labels assigned.} The set of all AMR graphs is denoted by $\mathcal{G}_\text{AMR}$.
\end{definition}

\begin{figure}[t]
\centering

\scalebox{0.8} {
\subfloat[][]{
\begin{tikzpicture}
\tikzstyle{amr-node}=[shape=ellipse,draw, inner sep=0.2, minimum height=0.8cm, text height=1.5ex, text depth=.25ex]
\tikzstyle{amr-edge}=[text height=1.5ex, text depth=.25ex, fill=white]
\tikzstyle{text-node}=[text height=1.5ex, text depth=.25ex, align=center]

	\node (PAD2) at (-2.5,0){};
	\node (PAD2) at (2.5,0){};

    \node[amr-node] (amr-woman) at (0,0) {woman};
    \node[amr-node] (amr-attract) at (0,-2) {attract-01};

    \path [-latex](amr-woman) edge node[amr-edge] {ARG0-of} (amr-attract);
    
    \node[text-node] at(0,-4.5) {an attractive woman \\ the attractive women \\ there is an attractive woman \\ the woman who attracts};      
\end{tikzpicture}
\label{fig:amr1}
}
\subfloat[][]{
\begin{tikzpicture}
\tikzstyle{amr-node}=[shape=ellipse,draw, inner sep=0.2, minimum height=0.8cm, text height=1.5ex, text depth=.25ex, minimum width=1.2cm]
\tikzstyle{amr-edge}=[text height=1.5ex, text depth=.25ex, fill=white]
\tikzstyle{text-node}=[text height=1.5ex, text depth=.25ex, align=center]

    \node[amr-node] (amr-think) at (0,0) {think-01};
    \node[amr-node] (amr-i) at (-2.8,-2) {I};
    \node[amr-node] (amr-this) at (0,-2) {this};
    \node[amr-node] (amr-minus) at (2.8,-2) {$-$};
    
    \path [-latex](amr-think) edge node[amr-edge] {ARG0} (amr-i);
    \path [-latex](amr-think) edge node[pos=0.52, amr-edge] {ARG1} (amr-this);
    \path [-latex](amr-think) edge node[amr-edge] {polarity} (amr-minus);

    \node[text-node] at(0,-4.5) {this is not what I think \\ this is not my thought \\ this was not a thought of mine \\ these were not my thoughts};      
\end{tikzpicture}
\label{fig:amr2}
}
\subfloat[][]{
\begin{tikzpicture}
\tikzstyle{amr-node}=[shape=ellipse,draw, inner sep=0.2, minimum height=0.8cm, text height=1.5ex, text depth=.25ex, minimum width=1.2cm]
\tikzstyle{amr-edge}=[text height=1.5ex, text depth=.25ex, fill=white]
\tikzstyle{text-node}=[text height=1.5ex, text depth=.25ex, align=center]

	\node (PAD2) at (-2.5,0){};
	\node (PAD2) at (2.5,0){};

    \node[amr-node] (amr-live) at (0,0) {live-01};
    \node[amr-node] (amr-he) at (-1.5,-2) {he};
    \node[amr-node] (amr-city) at (1.5,-2) {city};

    \path [-latex](amr-live) edge node[amr-edge] {ARG0} (amr-he);
    \path [-latex](amr-live) edge node[amr-edge] {location} (amr-city);
    
    \node[text-node] at(0,-4.5) {he lives in a city \\ he is living in the city \\ his life in the city \\ he lived in the city};      
\end{tikzpicture}
\label{fig:amr3}
}
}
\caption{Graphical representation of three exemplary AMR graphs; each vertex is inscribed with its label. Below each AMR graph, some of its realizations are shown.}
\label{fig:amrs}
\end{figure}
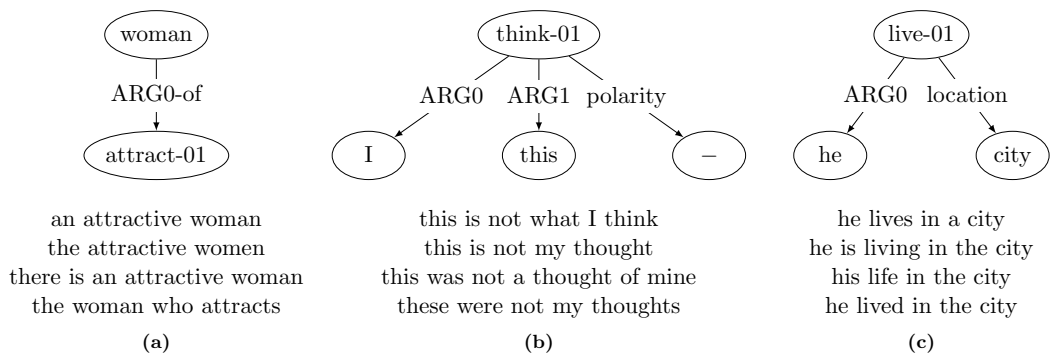

Given an AMR graph $G$, we call every sentence whose meaning is represented by $G$ a \emph{realization of $G$}.
An important goal of AMR is to assign the same graph to semantically equal sentences, even if they differ syntactically. To this end, words are mapped to PropBank framesets whenever possible; this applies not only to verbs, but also to other parts of speech (POS) such as nouns and adjectives. Examples of this are shown in the three AMR graphs depicted in Figure~\ref{fig:amrs} where the words ``attractive'', ``thought'' and ``life'' are represented by the framesets ``attract-01'', ``think-01'' and ``live-01'', respectively.

Parts of speech are by no means the only information that is not represented in AMR graphs. As can be seen in Figure~\ref{fig:amr3}, prepositions such as ``in'', ``to'' and ``for'' have no direct representation in AMR but are instead encoded through relation labels such as ``location'', ``direction'' and ``purpose''. Other limitations of AMR include that in general, neither definiteness nor grammatical number (see Figure~\ref{fig:amr1}) nor tense (Figure~\ref{fig:amr2}~and~\ref{fig:amr3}) of a sentence can directly be represented by its AMR graph. However, it is possible to explicitly include some of this information through special relations and concepts. To give an example, the grammatical number of a noun may be indicated by using the relation ``quant'' in combination with either a numerical value or an English word like ``many'', ``few'' or ``some''.

\begin{figure}[t]
\centering
\scalebox{0.8} {
\begin{tikzpicture}
\tikzstyle{amr-node}=[shape=ellipse,draw, inner sep=0.2, minimum height=0.8cm, text height=1.5ex, text depth=.25ex]
\tikzstyle{dep-node}=[shape=ellipse,draw, inner sep=0.2, minimum height=0.8cm, minimum width=1.5cm, text height=1.5ex, text depth=.25ex]
\tikzstyle{text-node}=[text height=1.5ex, text depth=.25ex]
\tikzstyle{amr-align}=[-latex,color=red]
\tikzstyle{dep-align}=[-latex,color=red]
\tikzstyle{frame}=[draw, dashed, inner sep=0.3cm]
	\node (PAD1) at (-7,0){};
	\node (PAD2) at (7,0){};
    \node[amr-node] (amr-want) at (0,0) {1\,:\,want-01};
    \node[amr-node] (amr-person) at (-3,-1.8) {2\,:\,person};
    \node[amr-node] (amr-sleep) at (4,-1.2) {3\,:\,sleep-01};
    \node[amr-node] (amr-develop) at (-1.8,-3.6) {4\,:\,develop-02};

    \path [-latex](amr-want) edge node[fill=white] {ARG0} (amr-person);
    \path [-latex](amr-want) edge node[fill=white] (amr-want-arg1) {ARG1} (amr-sleep);
    \path [-latex](amr-sleep) edge node[fill=white, pos=0.38] {ARG0} (amr-person);
    \path [-latex](amr-person) edge node[fill=white] {ARG0-of} (amr-develop);        
\end{tikzpicture}
}
\caption{Graphical representation of the AMR graph $G_1$ introduced in Example~\ref{example:amr}}
\label{fig:amr-1}
\end{figure}
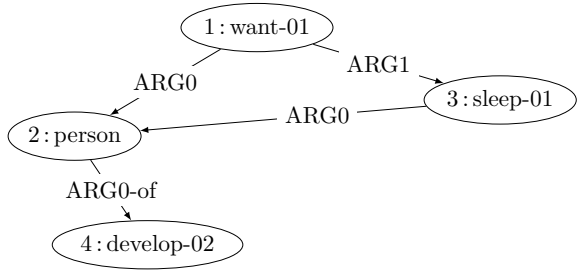

\begin{example}
\label{example:amr}
The meaning of the sentence ``The developer wants to sleep'' can be represented by the AMR graph $G_1 = (\{1, 2, 3, 4\}, E_1, L_1, \emptyset)$ with 
\begin{align*}
E_1 & = \{ (1, \text{ARG0}, 2), (1, \text{ARG1}, 3), (3, \text{ARG0}, 2), (2, \text{ARG0-of}, 4) \} \\
L_1 & = \{ (1, \text{want-01}), (2, \text{person}), (3, \text{sleep-01}), (4, \text{develop-02}) \}  \text{\,.}
\end{align*}
A graphical representation of $G_1$ can be seen in Figure~\ref{fig:amr-1}. The required PropBank framesets along with their roles are shown in Table~\ref{tab:propbank}. Note that the noun ``developer'' is represented by a combination of the English word ``person'' and the PropBank frameset ``develop-02''. Unlike the examples shown in Figure~\ref{fig:amrs}, $G_1$ is not a tree as the node labeled ``person'' is the zeroth argument to instances of both ``want-01'' and ``sleep-01``.
\end{example}

\subsubsection{Generation and Parsing}
\label{PRELIMINARIES:GenPar}

Common tasks with regard to AMR involve \emph{parsing}, the problem of finding the AMR graph corresponding to a sentence, and the inverse problem of \emph{generation}, i.e. finding a good natural-language realization of a given AMR graph.

\begin{definition}[Generator] {A function $g \colon \mathcal{G}_\text{AMR} \rightarrow \Sigma_\text{E}^*$ is called a \emph{generator}. Given a generator $g$ and an AMR graph $G \in \mathcal{G}_\text{AMR}$, we call $g(G)$ the \emph{sentence generated from $G$ by $g$} or the \emph{realization of $G$ according to $g$}.}
\end{definition}

\begin{definition}[Parser] {A function $p \colon \Sigma_\text{E}^* \rightarrow \mathcal{G}_\text{AMR}$ is called a \emph{parser}. Given a parser~$p$ and a sentence $w \in \Sigma_\text{E}^*$, we call $p(w)$ the \emph{parse of $w$ according to $p$}.}
\end{definition}

While according to the above definition, any function that maps English sentences to AMR graphs is called a parser, one would ideally like to find a parser that assigns to each English sentence $w$ the AMR graph $\hat{G}$ that best represents its meaning. As determining this unique AMR graph given an English sentence is an exceedingly difficult task, one is also interested in finding parsers that assign to each sentence $w$ an AMR graph $G$ that is at least roughly equal to $\hat{G}$. In order to be able to evaluate the quality of a parser, \citet{cai2013smatch} define the \emph{semantic match} (Smatch) metric which, given one or more pairs of graphs $(\hat{G_i}, G_i)$, $i \in [n]$ for some $n \in \mathbb{N}$, measures how similar all related graphs $\hat{G_i}$ and $G_i$ are and aggregates these similarity values to a cumulative score ranging from $0$ to $1$. Given a sequence $C = (G_1, w_1), \ldots, (G_n, w_n)$ of AMR graphs and corresponding sentences, Smatch can be used to automatically compare AMR parsers by calculating 
\[
\mathrm{score}(p) = \mathrm{Smatch}((G_1, p(w_1)), \ldots, (G_n, p(w_n)))
\]
for each parser $p$ and comparing the scores of all parsers. Details on how exactly the Smatch score can be calculated are beyond the scope of this work; we refer to~\citet{cai2013smatch} for an in-depth explanation.

Of course, the very same need for an evaluation metric arises when dealing with generation from AMR graphs: We require some way to measure the quality of generators in order to make comparisons between them. However, it is considerably more complex to evaluate a generator than a parser because given an AMR graph $G$, there is not necessarily just a single sentence $\hat{w}$ that corresponds to $G$; as the examples in Figure~\ref{fig:amrs} show, there may be several equally good realizations of $G$. 

The most common approach to the problem of evaluating generators is to make use of the \emph{bilingual evaluation understudy} (Bleu) score \citep{papineni2002bleu} that originates from the field of machine translation. Given a \emph{candidate sentence} $w$ and a \emph{reference sentence} $\hat{w}$, the basic idea of Bleu is to count the number of matching $n$-grams (i.e. contiguous phrases consisting of $n$ words) between $w$ and $\hat{w}$.\footnote{The Bleu score is actually designed to support several reference sentences $\hat{w}_1, \ldots, \hat{w}_k$. While this might sound useful to our application scenario, all currently published AMR corpora unfortunately feature only a single realization per graph (see Section~\ref{PRELIMINARIES:Corpora}).} This number is then divided by the total number of $n$-grams in the candidate sentence $w$. Typically, this computation is done not just for one but for several values of $n$ and the results are averaged subsequently; a common choice is $n = 1, \mydots, 4$. Some modifications such as clipping the count of candidate $n$-gram matches must be made in order to make the resulting score more meaningful; we will, however, not discuss these modifications here and refer to~\citet{papineni2002bleu} for further details. 

Just as Smatch, Bleu can be extended to compute a cumulative score ranging from $0$ to $1$ and measuring the pairwise similarity of each sentence pair $(\hat{w_i}, w_i)$, $i \in [n]$ contained within a sequence of $n \in \mathbb{N}$ sentence pairs. This allows us to compare a set of generators given a sequence $C = (G_1, w_1), \ldots, (G_n, w_n)$ of AMR graphs $G_i \in \mathcal{G}_\text{AMR}$ and corresponding realizations $w_i \in \Sigma_\text{E}^*$ by calculating
\[
\mathrm{score}(g) = \mathrm{Bleu}((w_1, g(G_1)), \ldots, (w_n, G(w_n)))
\]
for each generator $g$. A common modification to the above definition of Bleu is to scale the result by some factor $s \in \mathbb{N}^+$, resulting in the total score ranging from $0$ to $s$; the usual choice in the context of AMR generation is $s = 100$. Also, $w_i$ and $g(G_i)$ are often not directly used to compute the Bleu score but are converted to lower case beforehand. We refer to the so-obtained score as the \emph{case insensitive Bleu score}.

Especially in the scenario of AMR generation where given a graph $G$, there are often many -- and equally good -- realizations that may differ significantly with regards to the choice of words and syntactic structure, even scores well below the maximum do not necessarily imply that a generator performs poorly. Consider, for example, the lowercased sentence pair
\begin{align*}
\hat{w} & = \text{the boys couldn't close their eyes} \\
w & = \text{it is not possible for the boy to close his eyes}
\end{align*}
where $\hat{w}$ serves as a reference sentence and $w$ is the output of a generator. Although both sentences are equally good realizations of the AMR graph shown in Figure~\ref{fig:initial-amr-a}, they have only three common unigrams (``the'', ``close'', ``eyes'') and not a single common $n$-gram for $n \in \{2,3,4\}$, resulting in a very low score. As this example demonstrates, the Bleu score of a single generator would scarcely be meaningful. Nevertheless, it is an established baseline for relative judgments in comparison with other generators.

%
%

\subsubsection{Corpora}
\label{PRELIMINARIES:Corpora}

As we have seen in the previous section, the evaluation of parsers and generators using Smatch or Bleu requires a sequence of AMR graphs along with reference realizations; we refer to such a sequence as an \emph{AMR corpus}.

\begin{definition}[AMR corpus]
A sequence $C = ((G_1, w_1), \ldots, (G_n, w_n))$, $n \in \mathbb{N}$ where $G_i \in \mathcal{G}_\text{AMR}$ and $w_i \in \Sigma_\text{E}^*$ for all $i \in [n]$ is called an \emph{AMR corpus}. We refer to $n$ as the \emph{size of $C$} and to each tuple $(G_i, w_i)$, $i \in [n]$ as an \emph{element of $C$}.
\end{definition}

We often refer to an AMR corpus simply as \emph{corpus}. Of course, AMR corpora are not only useful for evaluation of parsers and generators, but as well for training them.
However, it is essential to not use the same data for both training and evaluation because obviously, we want a generator to perform well not only for inputs that it has already seen during training, but also for previously unknown graphs. Therefore, corpora are usually divided into several disjoint subcorpora: a sequence of \emph{training data} used to train the parser or generator, a sequence of \emph{development data} used e.g. for hyperparameter optimization, and a sequence of \emph{test data} on which the quality of the chosen approach can be evaluated.  

As AMR is a relatively new research topic, both the number of corpora and the number of graphs contained within these corpora is rather small compared to the number of available data for syntactic annotations like constituency trees and dependency trees. Importantly, all currently released AMR corpora consist only of AMR graphs with exactly one reference sentence per graph. Also, there is no information included with regards to how vertices and edges of the contained AMR graphs correspond to words of their realizations, i.e. no alignment between graphs and reference sentences is given.

An overview of some AMR corpora is given in Table~\ref{tab:corpora}. As its name suggests, the corpus \emph{The Little Prince} contains AMR graphs encoding the meaning of each sentence in the novel of the same name by Antoine de Saint-Exupéry. The Bio AMR corpus consists mostly of semantic annotations for cancer-related research papers. Both corpora released by the \emph{Linguistic Data Consortium} (LDC), LDC2014T12 and LDC2015E86, contain AMR graphs for English sentences obtained from various newswires, discussion forums and television transcripts.\footnote{Further details on the genres and contents of the listed corpora can be found at \url{amr.isi.edu/download.html}.} The latter corpus is an extension of the former, containing the same development and test data but several additional AMR graphs for training.

\begin{table}
\centering
\bgroup
\def\arraystretch{1.5}
\small
\begin{tabularx}{\textwidth}{|l|r|r|X|}
\hline
\textbf{Corpus} & \textbf{Total Size} & \textbf{Size (Train\,/\,Dev\,/\,Test)} & \textbf{Availability} \\
\hline
The Little Prince v1.6 & $1,562$ & $1,274$ / $145$ / $142$ & general release \\
Bio AMR v0.8 & $6,452$ & $5,452$ / $500$ / $500$ & general release \\
LDC2014T12 & $13,051$ & $10,313$ / $1,368$ / $1,371$ & general release \\
LDC2015E86 & $19,572$ & $16,833$ / $1,368$ / $1,371$ & not publicly available \\ 
\hline
\end{tabularx}
\caption{Overview of currently released AMR corpora. For each corpus, the total number of contained AMR graphs is listed along with the sizes of the training, development and test sets.}
\label{tab:corpora}
\egroup
\end{table}

\subsection{Dependency Trees}
\label{PRELIMINARIES:DependencyTrees}

An established way to model the syntactic structure of a sentence is through so-called \emph{dependencies} between its words \citep{tesniere1959elements,nivre2008algorithms}. A dependency consists of a \emph{head}, a \emph{dependent} and a \emph{relation} between them. While both the head and the dependent of a dependency are simply words of the analyzed sentence, their relation is usually described by a label taken from some set $L_\text{D}$\label{los:l_d} of \emph{dependency labels}.\footnote{A list of all dependency labels used throughout this work along with their meanings can be found at \url{universaldependencies.org/u/dep}.} To give an example, consider once more the sentence ``The developer wants to sleep''. The fact that ``developer'' is the nominal subject corresponding to the verb ``wants'' can be modeled through a dependency with head ``wants'', dependent ``developer'' and label ``nsubj''.

The main verb of a sentence is typically chosen to be its head, i.e. it is the only word that is not a dependent of any other word.
As dependency relations are asymmetric and every word is the dependent of at most one head, the set of all dependencies within a sentence $w$ can be viewed as a tree whose nodes correspond to the sentence's words and whose root is the main verb of $w$.

\begin{definition}[Dependency tree] \label{los:g_dep}
A $(L_\text{D}, \Sigma_\text{E})$-graph $G = (V,E,L,\prec)$ is called a \emph{dependency tree} if it is a totally ordered tree. The set of all dependency trees is denoted by $\mathcal{G}_\text{DEP}$.
\end{definition}

Let $w \in \Sigma_\text{E}^*$ be a sentence and $G = (V,E,L,\prec)$ be a dependency tree. We call $G$ a \emph{dependency tree for $w$} if there is some bijection $b \colon V \rightarrow [|w|]$ such that for all $v, v' \in V$ and $i \in [|w|]$, it holds that
$b(v) = i \Rightarrow L(v) = w(i)$ and $v \prec v' \Leftrightarrow b(v) < b(v')$.

\begin{example} \label{example:dep-tree} We consider the graph $G_2 = (\{1,2,3,4,5\},E_2,L_2,\prec_2)$ where
\begin{align*}
E_2 & = \{ (1,\text{nsubj},2), (1, \text{xcomp}, 3), (2, \text{det}, 4), (3, \text{mark}, 5) \} \\
L_2 & = \{ (1, \text{wants}), (2, \text{developer}), (3, \text{sleep}), (4, \text{The}), (5, \text{to}) \} \\
\prec_2 & = \{ (4,2), (2,1), (1,5), (5,3) \}^+ \text{\,.}
\end{align*} 
As can easily be seen, $G_2$ is a dependency tree for the sentence ``The developer wants to sleep''; the corresponding bijection is $b = \{(1,3), (2,2), (3,5), (4,1), (5,4) \}$.
A graphical representation of $G_2$ can be seen in the lower half of Figure~\ref{fig:bigraph}.
\end{example}

\subsection{Bigraphs}

\begin{definition}[Aligned bigraph]
Let $\Sigma$ be an alphabet and let $L_E, L_V$ be sets. An \emph{(aligned) bigraph over $(\Sigma,L_E,L_V)$} is a tuple $\mathcal{B} = (G_1, G_2, w, A_1, A_2)$ where
\begin{enumerate}
\item $G_1 = (V_1, E_1, L_1, \prec_1)$ and $G_2 = (V_2, E_2, L_2, \prec_2)$ are graphs with edge labels from $L_E$ and vertex labels from $L_V$;
\item $w = w_1 \ldots w_n \in \Sigma^*$ is a string over $\Sigma$ with length $n \in \mathbb{N}$;
\item $A_1 \subseteq V_1 \times [n]$ and $A_2 \subseteq V_2 \times [n]$ are \emph{alignments} that connect vertices of $G_1$ and $G_2$ with symbols of $w$. \qedhere
\end{enumerate} 
\end{definition}

If we are not interested in the particular sets $\Sigma$, $L_E$ and $L_V$, we refer to a bigraph over $(\Sigma, L_E, L_V)$ simply as \emph{bigraph}.
Let $\mathcal{B} = (G_1, G_2, w, A_1, A_2)$ be an aligned bigraph and $G_i = (V_i, E_i, L_i, \prec_i)$ for $i \in \{1,2\}$.
For $v \in V_i$, $i \in \{1, 2\}$, we denote by $A_i(v)$ the set $\{ j \in [|w|] \mid (v,j) \in A_i \}$ of all indices of symbols to which $v$ is aligned. If $v$ is only aligned to a single symbol with index $j \in [|w|]$, we sometimes identify $\{j\}$ with $j$. That is, we view $A_i(v)$ as being the actual number $j$ rather than the singleton set $\{j\}$.
We define two mappings $\pi_\mathcal{B}^1: V_1 \rightarrow \mathcal{P}( V_2  )$ and $\pi_\mathcal{B}^2: V_2 \rightarrow \mathcal{P}( V_1)$ with
\begin{align*}
\pi_\mathcal{B}^1(v_1) & = \{ v_2 \in V_2  \mid (v_1, v_2) \in A_1 A_2^{-1} \} \\
\pi_\mathcal{B}^2(v_2) & = \{ v_1 \in V_1  \mid (v_1, v_2) \in A_1 A_2^{-1} \}
\end{align*}\label{los:pi_1_2}
such that $\pi_\mathcal{B}^1$ assigns to each vertex $v$ of $G_1$ all vertices of $G_2$ that are aligned to at least one symbol of $w$ to which $v$ is also aligned; vice versa, $\pi_\mathcal{B}^2$ assigns to each vertex of $G_2$ all vertices of $G_1$ connected to it through some common alignment.

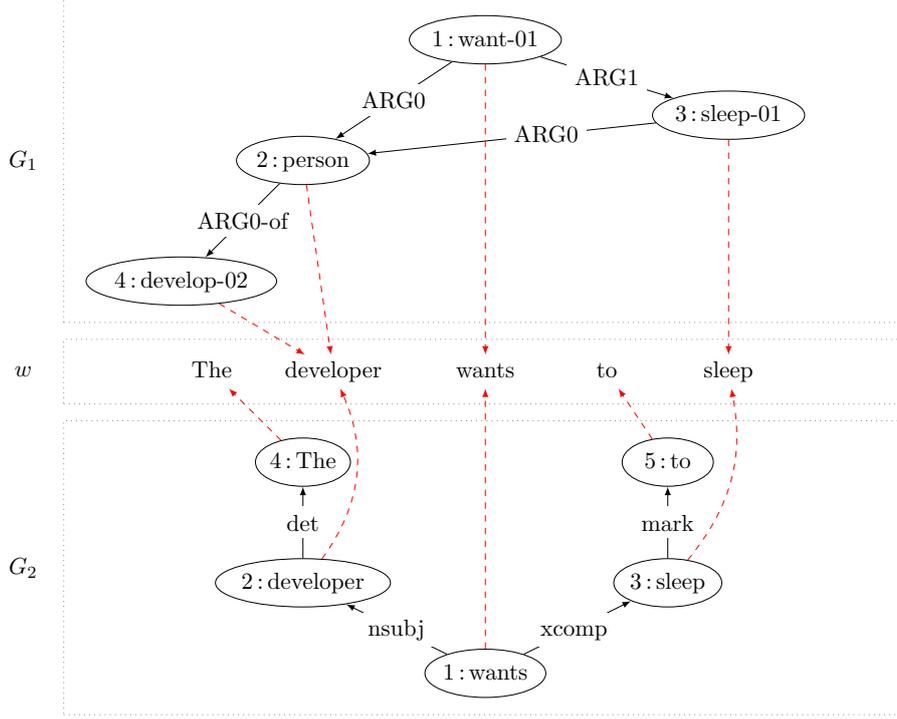
\begin{figure}[t]
\centering
\scalebox{0.8} {
\begin{tikzpicture}
\tikzstyle{amr-node}=[shape=ellipse,draw, inner sep=0.2, minimum height=0.8cm, text height=1.5ex, text depth=.25ex]
\tikzstyle{dep-node}=[shape=ellipse,draw, inner sep=0.2, minimum height=0.8cm, minimum width=1.5cm, text height=1.5ex, text depth=.25ex]
\tikzstyle{text-node}=[text height=1.5ex, text depth=.25ex]
\tikzstyle{amr-align}=[-latex,color=red,dashed]
\tikzstyle{dep-align}=[-latex,color=red,dashed]
\tikzstyle{amr-edge}=[text height=1.5ex, text depth=.25ex, fill=white]
\tikzstyle{frame}=[draw, dotted, color=gray, inner sep=0.3cm]

	\node (PAD1) at (-7,0){};
	\node (PAD2) at (7,0){};

    \node[amr-node] (amr-want) at (0,0) {1\,:\,want-01};
    \node[amr-node] (amr-person) at (-3,-2) {2\,:\,person};
    \node[amr-node] (amr-sleep) at (4,-1.25) {3\,:\,sleep-01};
    \node[amr-node] (amr-develop) at (-5,-4) {4\,:\,develop-02};

    \path [-latex](amr-want) edge node[fill=white] {ARG0} (amr-person);
    \path [-latex](amr-want) edge node[fill=white] (amr-want-arg1) {ARG1} (amr-sleep);
    \path [-latex](amr-sleep) edge node[fill=white, pos=0.38] {ARG0} (amr-person);
    \path [-latex](amr-person) edge node[fill=white] {ARG0-of} (amr-develop);
    
    \node (amr-pad1) at (-6.5,0.25){};
    \node (amr-pad2) at (6.5,-4.25){};
    \node[frame, fit=(amr-pad1)(amr-pad2)] (amr-frame) {};
    
    \node[text-node] (text-the) at (-4.5, -5.5) {The};
    \node[text-node] (text-developer) at (-2.5, -5.5) {developer};
    \node[text-node] (text-wants) at (0, -5.5) {wants};
    \node[text-node] (text-to) at (2, -5.5) {to};
    \node[text-node] (text-sleep) at (4, -5.5) {sleep};
    
    \node (text-pad1) at (-6.5,-5.4){};
    \node (text-pad2) at (6.5,-5.6){};
    \node[frame, fit=(text-pad1)(text-pad2)] (text-frame) {};
    
    \path[amr-align] (amr-person) edge node[] {} (text-developer);
    \path[amr-align] (amr-develop) edge node[] {} (text-developer);
    \path[amr-align] (amr-want) edge node[] {} (text-wants);
    \path[amr-align] (amr-sleep) edge node[] {} (text-sleep);
    
    \node[dep-node] (dep-the) at (-3, -7) {4\,:\,The};
    \node[dep-node] (dep-to) at (3, -7) {5\,:\,to};
    \node[dep-node] (dep-developer) at (-3, -9) {2\,:\,developer};
    \node[dep-node] (dep-sleep) at (3, -9) {3\,:\,sleep};
    \node[dep-node] (dep-wants) at (0, -10.5) {1\,:\,wants};

    \path [-latex](dep-wants) edge node[amr-edge] {nsubj} (dep-developer);
    \path [-latex](dep-wants) edge node[amr-edge, pos=0.47] {xcomp} (dep-sleep);
    \path [-latex](dep-developer) edge node[fill=white] {det} (dep-the);
    \path [-latex](dep-sleep) edge node[fill=white] {mark} (dep-to); 
    
    \node (dep-pad1) at (-6.5,-6.75){};
    \node (dep-pad2) at (6.5,-10.75){};
    \node[frame, fit=(dep-pad1)(dep-pad2)] (dep-frame) {};
    
    \path[dep-align] (dep-the) edge node[] {} (text-the);
    \path[dep-align, bend angle=30, bend right] (dep-developer) edge node[] {} (text-developer);
    \path[dep-align] (dep-wants) edge node[] {} (text-wants);
    \path[dep-align, bend angle=25, bend right] (dep-sleep) edge node[] {} (text-sleep);
    \path[dep-align] (dep-to) edge node[] {} (text-to);   
    
    \node[] at (-7.6, -2) {$G_1$}; 
    \node[] at (-7.6, -5.5) {$w$}; 
    \node[] at (-7.6, -8.75) {$G_2$};         
\end{tikzpicture}
}
\caption{Graphical representation of the bigraph $\mathcal{B} = (G_1, G_2, w, A_1, A_2)$ defined in Example~\ref{example:bigraph}. For $i \in \{1,2\}$, each node $v$ of $G_i$ is inscribed with $v$\,:\,$L_i(v)$; each alignment $(u,j) \in A_i$ is represented by a dashed arrow line connecting $u$ and $w(j)$.}
\label{fig:bigraph}
\end{figure}

\begin{example}
\label{example:bigraph}
Let $G_1$ and $G_2$ be defined as in Example~\ref{example:amr}~and~\ref{example:dep-tree}, respectively. We consider the bigraph  $\mathcal{B} = (G_1, G_2, w, A_1, A_2)$ over $(\Sigma_\text{E}, L_\text{R} \cup L_\text{D}, L_\text{C} \cup \Sigma_\text{E})$ where 
\begin{align*}
w & = \text{The developer wants to sleep} \\
A_1 & = \{ (1,3), (2,2), (3,5), (4,2) \} \qquad 
A_2 = \{ (1,3), (2,2), (3,5), (4,1), (5,4) \} \text{\,.}
\end{align*}
A graphical representation of $\mathcal{B}$ is shown in Figure~\ref{fig:bigraph}. The following statements are true:
\[
\pi_\mathcal{B}^1(2) = \{2\} \qquad \pi_\mathcal{B}^2(2) = \{2, 4\} \qquad \pi_\mathcal{B}^2(5) = \emptyset \text{\,.}\qedhere
\]
\end{example}

\begin{definition}[Span] \label{def:span}
Let $\mathcal{B} = (G_1, G_2, w, A_1, A_2)$ be a bigraph, $i \in \{ 1, 2 \}$ and let $G_i = (V_i, E_i, L_i, {\prec}_i)$ be an acyclic graph. The function $\mathrm{span}_\mathcal{B}^i: V_i \mapsto \mathcal{P}(\{1, \ldots, |w| \})$ is defined inductively for all $v \in V_i$ as
\[
\mathrm{span}_\mathcal{B}^i(v) = A_i(v) \cup \bigcup_{v' \in \ch[G_i]{v}} \mathrm{span}_\mathcal{B}^i(v')\,. \qedhere
\]
\end{definition}

\begin{example}
We consider once more the bigraph $\mathcal{B} = (G_1, G_2, w, A_1, A_2)$ shown in Figure~\ref{fig:bigraph}. The following holds true:
\begin{align*}
\mathrm{span}_\mathcal{B}^1(1) & = \{ 3 \} \cup \mathrm{span}_\mathcal{B}^1(2) \cup \mathrm{span}_\mathcal{B}^1(3) = \{ 2, 3, 5 \} \\
\mathrm{span}_\mathcal{B}^2(3) & = \{ 5 \} \cup \mathrm{span}_\mathcal{B}^2(5) = \{ 4, 5 \} \,. \qedhere
\end{align*}
\end{example}

\subsection{Transition Systems}

The key idea of this work is to define several actions -- such as the deletion, merging and reordering of edges and vertices -- to transform an AMR graph $G$ into a tree structure. This structure is then turned into a realization of $G$ through application of the yield function introduced in Definition~\ref{def:yield}. To embed the different kinds of required actions into a unified framework, we adapt the notion of \emph{transition systems} from \citet*{nivre2008algorithms}, but we extend the definition found therein by allowing polymorphic input and output and introducing the concept of a \emph{finalization function}. 

\begin{definition}[Transition system]
\label{def:transition-system}
Let $\mathcal{I}$ and $\mathcal{O}$ be sets (\emph{input space} and \emph{output space}). A \emph{transition system for $(\mathcal{I}, \mathcal{O})$} is a tuple $S = (C, T, C_t, c_s, c_f)$ where 
\begin{enumerate}
\item $C$ is a set of \emph{configurations} (also called \emph{states});
\item $T$ is a set of \emph{transitions}, each of which is a partial function $t \colon C \pfun C$;
\item $C_t \subseteq C$ is a set of \emph{terminal configurations};
\item $c_s: \mathcal{I} \rightarrow C$ is an \emph{initialization function} that maps each input from the set $\mathcal{I}$ to an \emph{initial configuration};
\item $c_f: C \pfun \mathcal{O}$ is a \emph{finalization function} that maps some configurations to an output from the set $\mathcal{O}$.
\end{enumerate}

Let $S = (C, T, C_t, c_s, c_f)$ be a transition system for $(\mathcal{I}, \mathcal{O})$ and let $I \in \mathcal{I}$ be some input. A \emph{partial transition sequence for $I$ in $S$} is a sequence of transitions $(t_1, \ldots, t_n) \in T^*$, $n \in \mathbb{N}^+$ where 
\[
t_{i-1} ( \ldots t_1 ( c_s(I) ) \ldots ) \in \text{dom}(t_i)
\]
for all $i \in [n]$. 
Let $\tau = (t_1, \ldots, t_n)$ be a partial transition sequence for $I$ in $S$. We denote by $\tau(I)$ the configuration obtained from applying the transitions $t_1, \ldots, t_n$ to $c_s (I)$, i.e.
\[
\tau(I) = t_n ( \ldots t_1 (c_s (I)) \ldots ) \,.
\]
If $\tau(I) \in C_t \cap \text{dom}(c_f)$, we call $(t_1, \ldots, t_n)$ a \emph{terminating transition sequence} or simply a \emph{transition sequence}. The \emph{output} of a terminating transition sequence $\tau$ with input $I$ is then defined as $out(\tau, I) = c_f(\tau(I))$. The set of all terminating transition sequences for $I$ in $S$ is denoted by $\mathcal{T}(S,I)$. \label{los:t_s_i}
\end{definition}

\subsection{Language Modeling}

A common way to improve results in natural language generation from AMR graphs is to judge each candidate realization based on two criteria: Firstly, how well does it transfer the meaning encoded by the graph? Secondly, how well does it fit into the target language? Of course, the second question can be answered regardless of the underlying graph. This is typically done using a \emph{language model} that assigns a probability to each sentence of the target language.  

\begin{definition}[Language model] 
\label{def:language-model}
Let $\Sigma$ be an alphabet. A function $p: \Sigma^* \rightarrow [0,1]$ is called a \emph{$\Sigma$-language model} if it is a probability distribution of $\Sigma^*$.
\end{definition}

Let $\Sigma$ be some alphabet, $w = (w_1, \ldots, w_m)$, $m \in \mathbb{N}$ be a string over $\Sigma$ and let $P(w_1, \ldots, w_n)$ denote the probability of observing this very string. The general product rule allows us to write
\[
P(w_1, \ldots, w_m) = P(w_1) \cdot P(w_2 \mid w_1) \cdot \ldots \cdot P(w_m \mid w_1, \ldots, w_{m-1})\,.
\]
A simplifying assumption often made is that the probability of a symbol $w_i$, $i \in [m]$ occurring in $w$ does not depend on \emph{all} previously occurring symbols $w_1$ to $w_{i-1}$, but only on a fixed number $n \in \mathbb{N}$ of previous symbols. As the first $n-1$ symbols in a sequence $w$ do not have $n$ previous symbols, we simply insert $n-1$ \emph{start symbols} (denoted by $\langle s \rangle$) at the very left of the sequence. Under this assumption, we can rewrite
\[
P(w_1, \ldots, w_m) = \prod_{i=1}^m P(w_i \mid w_{i-n}, \ldots, w_{i-1})
\]
where $w_i = \langle s \rangle$ for $i \leq 0$. 
A language model implementing this assumption is called an \emph{$n$-gram language model}. The conditional probability $P(w_i \mid w_{i-n}, \ldots, w_{i-1})$ is often approximated by a conditional probability distribution $p$ of $\Sigma$ given $\Sigma^n$ estimated from a natural language corpus $C = (w^1, \ldots, w^k) \in (\Sigma^*)^k$, $k \in \mathbb{N}$ as 
\[
p(w_i \mid w_{i-n}, \ldots, w_{i-1}) = \frac{ \text{count}_C ((w_{i-n}, \ldots, w_{i-1}, w_i)) }{ \text{count}_C ((w_{i-n}, \ldots, w_{i-1}))}
\]
where for all $w \in \Sigma^*$, $\text{count}_C(w)$ denotes the number of occurrences of $w$ as a substring within all strings in $C$. However, this simple approach suffers from the fact that whenever some sequence $(w_{i-n}, \ldots, w_{i-1}, w_i)$ does not occur at all in $C$, the corresponding estimated value of $p(w_i \mid w_{i-n}, \ldots, w_{i-1})$ and the probability assigned to all strings containing this sequence is equal to zero; thus, a language model trained this way is not able to handle previously unseen symbols or sequences thereof. To overcome this problem, several \emph{smoothing} methods can be applied; the underlying idea is to subtract a small amount~$\delta$ from all observed $n$-gram counts and to distribute it among unobserved sequences. 

\begin{example}
Let 
$C = (\text{the man sleeps, the man and the boy, a man}) \in (\Sigma_\text{E}^*)^3$ be an English corpus. The conditional probability $p(\text{man} \mid \text{the})$ estimated from $C$ is
\[
p(\text{man} \mid \text{the}) = \frac{ \text{count}_C (\text{the man}) }{ \text{count}_C (\text{the})} = \frac{2}{3}\,. \qedhere
\]
\end{example}

A natural language corpus commonly used to train $n$-gram models for the English language is \emph{Gigaword}, which consists of several million sentences obtained from various English newswire sources. As of now, five versions of Gigaword have been released, the first one being Gigaword~v1 (LDC2003T05) and the newest one being Gigaword~v5 (LDC2011T07).\footnote{The general releases of Gigaword~v1 (LDC2003T05) and Gigaword~v5 (LDC2011T07) are available at \url{catalog.ldc.upenn.edu/ldc2003t05} and \url{catalog.ldc.upenn.edu/ldc2011t07}, respectively.} 

The language model used in Section~\ref{EXPERIMENTS} of this work is a $3$-gram language model trained on Gigaword~v1. For smoothing, we make use of a method commonly known as \emph{Kneser-Ney smoothing}. The details of this method are beyond the scope of this work; we refer to \citet{kneser1995improved}. 

\subsection{Maximum Entropy Modeling}
\label{PRELIMINARIES:LinearClassification}

Maximum entropy modeling is a concept that can be used to estimate conditional probabilities given a set of training data \citep{berger1996maximum}. We will make frequent use of maximum entropy models when defining our transition system in Section~\ref{GENERATION}; for example, given a configuration $c$ and a transition $t$, we will use maximum entropy models to estimate $P(t \mid c)$, the probability that $t$ is the correct transition to be applied next. 

For the remainder of this section, let $\mathcal{Y}$ be a finite set of possible \emph{outputs} and let $\mathcal{X}$ be a set of \emph{contexts}. We will show how for all $y \in \mathcal{Y}$ and $x \in \mathcal{X}$, a maximum entropy model estimates the conditional probability of $y$ being the correct output given context $x$.  To this end, we use the definitions of features and maximum entropy models introduced in \citet{berger1996maximum} with some slight adjustments to our special use case.

\begin{definition}[Feature function] A function $f \colon \mathcal{X} \times \mathcal{Y} \rightarrow \mathbb{R}$ is called a \emph{feature function} or, in short, a \emph{feature}. \end{definition}

Let $\mathbf{f} = (f_1, \ldots, f_n )$ be a finite sequence of features $f_i \colon \mathcal{X} \times \mathcal{Y} \rightarrow \mathbb{R}$. 
The reason for introducing the concept of features is that we would like to reduce each pair $(x,y) \in \mathcal{X} \times \mathcal{Y}$ of arbitrary complexity to a real-valued vector $\mathbf{f}(x,y) = (f_1(x,y), \ldots, f_n(x,y)) \in \mathbb{R}^n$. A maximum entropy model then estimates the probability of $y$ given $x$ only from $\mathbf{f}(x,y)$; all information contained within $x$ and $y$ but not represented in $\mathbf{f}(x,y)$ is discarded.

\begin{example}
\label{example:features}
Let $\mathcal{X} = \mathcal{G}_\text{AMR}$ and $\mathcal{Y} = \{ q, s \}$ where given an AMR graph $G$, the output $q$ indicates that $G$ represents a question and $s$ indicates that $G$ represents a statement. A reasonable choice of feature functions could be $\mathbf{f} = (f_1^q, f_1^s, f_2^q, f_2^s)$ where
\begin{align*}
f_1^y((V, E, L, \prec), y') & = \begin{cases} 1 & \text{if } y = y' \wedge \exists v \in V \colon L(v) = \text{interrogative} \\
0 & \text{otherwise} \end{cases} \\
f_2^y((V, E, L, \prec), y') & = \begin{cases} |V| & \text{if } y = y' \\
0 & \text{otherwise} \end{cases}
\end{align*}
for all $y, y' \in \mathcal{Y}$ and $(V, E, L, \prec) \in \mathcal{G}_\text{AMR}$. That is, we try to decide upon whether $G$ represents a question or a statement by considering only whether it contains a vertex with label ``interrogative'' and how many vertices it contains in total. 
\end{example}

\begin{definition}[Maximum entropy model]
A \emph{maximum entropy model for $\mathcal{Y}$ and $\mathcal{X}$} is a conditional probability distribution $p$ of $\mathcal{Y}$ given $\mathcal{X}$ where 
\[
p(y \mid x) = \frac{1}{Z_\lambda(x)} \exp \left( \sum_{i=1}^n \lambda_i f_i (x,y) \right)
\]
with $\mathbf{f} = (f_1, \ldots f_n)$ being a finite sequence of features, $\lambda = (\lambda_1, \ldots, \lambda_n)$ being a sequence of real-valued parameters $\lambda_i \in \mathbb{R}$ for $i \in [n]$ and  
\[
 Z_\lambda(x) = \sum_{y \in \mathcal{Y}} \exp \left( \sum_{i=1}^{n} \lambda_i f_i (x,y)\right)
\]
being a normalizing factor to ensure that $p$ is indeed a probability distribution.
\end{definition}

For a detailed derivation of the above definition and a discussion of the assumptions required so that $P(y \mid x)$ can be estimated by $p(y \mid x)$, we refer to~\citet{berger1996maximum}. When the sets $\mathcal{Y}$ and $\mathcal{X}$ are clear from the context, we refer to a maximum entropy model for $\mathcal{Y}$ and $\mathcal{X}$ simply as a \emph{maximum entropy model}. While the sequence of features $\mathbf{f}$ to be used by a maximum entropy model must be specified by hand, the optimal parameter vector $\lambda$ can automatically be determined given a sequence of training data for which the true output is known, i.e. a sequence $C = (x_1, y_1) , \ldots, (x_m, y_m) \in (\mathcal{X} \times \mathcal{Y})^*$. The log likelihood of parameter $\lambda$ given $C$ can be calculated as
\begin{align*}
L(\lambda \mid C) = \log \prod_{j=1}^m p (y_j \mid x_j) = \sum_{j=1}^m \sum_{i=1}^n \lambda_i f_i(x_j, y_j) - \sum_{j=1}^m \log Z_\lambda(x_j)
\end{align*}
and the optimal parameter vector 
\[
\hat{\lambda} = \argmax_{\lambda \in \mathbb{R}^n} L(\lambda \mid C)
\]
can be obtained through several numerical methods such as the \emph{Improved Iterative Scaling} (IIS) algorithm \citep{della1997inducing}. As the details of this process -- which is also referred to as \emph{training} of the model -- are not relevant for the design of our generator, we again refer to~\citet{berger1996maximum} for further details.

For the rest of this section, we discuss some convenient methods to turn various functions into features or feature vectors. While none of the following definitions is required for maximum entropy modeling, they simplify the notation of features used throughout this work considerably.

It is often useful to construct features by combining some information extracted only from $\mathcal{X}$ with just a single output $y \in \mathcal{Y}$. We therefore introduce a concise notation for features constructed in such a way. To this end, let $f \colon \mathcal{X} \mapsto \mathbb{R}$ and let $Y = (y_1, \ldots, y_n)$ be some enumeration of $\mathcal{Y}$. We denote by $f^Y$ the sequence $(f^{y_1}, \ldots f^{y_n})$ where each $f^{y_i}$, $i \in [n]$ is a feature function with
\[
f^{y_i}(x, y) =
 \begin{cases}
 f(x) & \text{if } y = y_i \\
 0 & \text{otherwise.} 
 \end{cases}
\]
As the actual order within $f^Y$ is irrelevant as long as it is used consistently, we denote by $f^\mathcal{Y}$ the sequence of features obtained in the above way from some arbitrary but fixed enumeration of $\mathcal{Y}$.  

\begin{example}
We consider once again the features $f_2^q$ and $f_2^s$ introduced in Example~\ref{example:features}. For $f \colon \mathcal{G}_\text{AMR} \rightarrow \mathbb{R}$, defined for each $G = (V,E,L,\prec) \in \mathcal{G}_\text{AMR}$ by $f(G) = |V|$, it holds that $f^{(q,s)} = (f_2^q, f_2^s)$.
\end{example}

\begin{definition}[Indicator feature function]
Let $S$ be an arbitrary set. We refer to a function $s \colon \mathcal{X} \rightarrow \mathcal{P}(S)$ where $s(x)$ is finite for all $x \in \mathcal{X}$ as an \emph{indicator feature function} or, in short, an \emph{indicator feature}. 
\end{definition}

Given a sequence $(x_1, \ldots, x_n) \in \mathcal{X}^n$ of training data, each indicator feature $s \colon \mathcal{X} \rightarrow \mathcal{P}(S)$ can be turned into a sequence of features as follows: Let $\{s_1, \ldots, s_m \} = \bigcup_{i=1}^n s(x_i)$. We first construct the ancillary sequence $f_{s_1}, \ldots, f_{s_m}$ where
\[
f_{s_i}(x) = 
	\begin{cases}
	1 & \text{if } s_i \in s(x) \\
	0 & \text{otherwise} 
	\end{cases}
\]
for all $i \in [m]$. On this basis, we construct the sequence of features $\mathbf{f} = f_{s_1}^\mathcal{Y} \cdot \ldots \cdot f_{s_m}^\mathcal{Y}$.

\begin{definition}[Indicator feature composition] Let $S_1$ and $S_2$ be sets and let $s_1 \colon \mathcal{X} \rightarrow \mathcal{P}(S_1)$ and $s_2 \colon \mathcal{X} \rightarrow \mathcal{P}(S_2)$ be indicator feature functions. The \emph{composition of $s_1$ and $s_2$} is the indicator feature function $s_1 \circ s_2 \colon \mathcal{X} \rightarrow \mathcal{P}(S_1 \times S_2)$ with
\[
(s_1 \circ s_2) (x) = \{ (a, b) \in S_1 \times S_2 \mid a \in s_1(x) \wedge b \in s_2(x) \}\,. \qedhere
\] \end{definition}

\begin{example}
Let $G = (V,E,L,\prec)$ be an AMR graph. For a maximum entropy model to predict transitions, a reasonable set of contexts could be $\mathcal{X} = \mathcal{G}_\text{AMR} \times V$ where for each tuple $(G', v) \in \mathcal{X}$, $G'$ is the graph obtained from $G$ so far through previously applied transitions and $v$ is the vertex to which we want to apply the next transition. Two interesting indicator features might be $s_1 \colon \mathcal{X} \rightarrow \mathcal{P}(L_\text{C})$ and $s_2 \colon \mathcal{X} \rightarrow \mathcal{P}(L_\text{C})$ where given $G' = (V, E', L', \prec')$ and $v \in V$,
\[
s_1((G', v)) = \{ L'(c) \mid c \in \ch[G']{v} \} \qquad
s_2((G', v)) = \{ L'(p) \mid p \in \pa[G']{v} \} \,.
\] In other words, $s_1$ and $s_2$ assign to a context $(G',v)$ the set of all labels assigned to children and parents of $v$ in $G'$, respectively. The composition of $s_1$ and $s_2$ is the new indicator feature function  $s_1 \circ s_2 \colon \mathcal{X} \rightarrow \mathcal{P}({L_\text{C}}^2)$ where
\[
(s_1 \circ s_2)((G', v)) = \{ (L'(c), L'(p)) \mid c \in \ch[G']{v}  \wedge p \in \pa[G']{v} \}\,. \qedhere
\]
\end{example}

\clearpage


\section{Transition-based Generation from AMR}
\label{GENERATION}

We now define a transition system $S_\text{AMR}$ for $(\mathcal{G}_\text{AMR}, \Sigma_\text{E}^*)$ which we then extend to an actual generator by assigning probabilities to its transitions. For this purpose, we proceed as follows: After introducing the concept of \emph{syntactic annotations} in Section~\ref{GENERATION:SyntacticAnnotation}, we define the actual transition system $S_\text{AMR}$ in Section~\ref{GENERATION:TransitionSystem} and derive how given a probability distribution of its transitions, a generator $g \colon \mathcal{G}_\text{AMR} \rightarrow \Sigma_\text{E}^*$ can be built from it. To this end, we first theoretically derive the optimal output $\hat{w}$ of $g$ given an AMR graph $G$. As computing this optimal output is not feasible for large graphs, we then devise an efficient algorithm to approximate $\hat{w}$. In Section~\ref{GENERATION:Training}, it is described how given a corpus of AMR graphs and reference realizations, the required probability distribution can be learned using several maximum entropy models. We discuss how postprocessing steps can be applied to the generated sentence for further improvement of our results in Section~\ref{GENERATION:Postprocessing}.  Finally, we investigate in Section~\ref{GENERATION:HyperparameterOptimization} how hyperparameters used throughout the generation process can be optimized using a set of development data.


\subsection{Syntactic Annotations}
\label{GENERATION:SyntacticAnnotation}

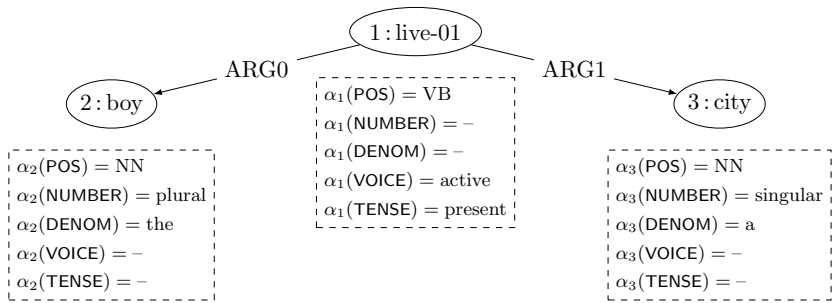
\begin{figure}[t]
\centering
\scalebox{0.8} {
\begin{tikzpicture}[baseline=(current bounding box.center)]
\tikzstyle{amr-node}=[shape=ellipse,draw, inner sep=0.2, minimum height=0.8cm, minimum width=1cm]

    \node[amr-node] (amr-live) at (0,0) {1\,:\,live-01};
    \node[amr-node] (amr-boy) at (-5,-1.2) {2\,:\,boy};
    \node[amr-node] (amr-city) at(5, -1.2) {3\,:\,city};
    
        \path [-latex](amr-live) edge node[fill=white] {ARG0} (amr-boy);
        \path [-latex](amr-live) edge node[fill=white] {ARG1} (amr-city);
    
        \node[below= 0.35cm of amr-city, fill=white, draw, dashed, align=left] (rho-to) {
	        \footnotesize $\alpha_3(\anno{POS}) = \text{NN}$ \\ 
	        \footnotesize $\alpha_3(\anno{NUMBER}) = \text{singular}$ \\ 
	        \footnotesize $\alpha_3(\anno{DENOM}) = \text{a}$ \\ 
	        \footnotesize $\alpha_3(\anno{VOICE}) = \text{--}$ \\ 
	        \footnotesize $\alpha_3(\anno{TENSE}) = \text{--}$
        };
        \node[below= 0.35cm of amr-boy, fill=white, draw, dashed, align=left] (rho-he) {
	        \footnotesize $\alpha_2(\anno{POS}) = \text{NN}$ \\ 
	        \footnotesize $\alpha_2(\anno{NUMBER}) = \text{plural}$ \\ 
	        \footnotesize $\alpha_2(\anno{DENOM}) = \text{the}$ \\ 
	        \footnotesize $\alpha_2(\anno{VOICE}) = \text{--}$ \\ 
	        \footnotesize $\alpha_2(\anno{TENSE}) = \text{--}$
        };
        \node[below=0.35cm of amr-live, fill=white, draw, dashed, align=left] (rho-want) {
	        \footnotesize $\alpha_1(\anno{POS}) = \text{VB}$ \\ 
	        \footnotesize $\alpha_1(\anno{NUMBER}) = \text{--}$ \\ 
	        \footnotesize $\alpha_1(\anno{DENOM}) = \text{--}$ \\ 
	        \footnotesize $\alpha_1(\anno{VOICE}) = \text{active}$ \\ 
	        \footnotesize $\alpha_1(\anno{TENSE}) = \text{present}$
        };
\end{tikzpicture}
}
\caption{Exemplary syntactic annotations for an AMR graph; the annotations for each vertex are written below it and surrounded by dashed lines. A reasonable realization of the graph would be ``the boys live in a city'' whereas, for example, neither ``the boy lives in a city'' nor ``the boys' life in the city'' would be consistent with the given syntactic annotation.}
\label{fig:syntactic-annotation}
\end{figure}

As we have seen in Section~\ref{PRELIMINARIES:AMR}, a lot of -- mostly syntactic -- information like parts of speech, number and tense gets lost in the text-to-AMR parsing process. As this information would be useful for the generation of an English sentence from an AMR graph, a key idea of this work is to annotate AMR graphs with reconstructed versions thereof. Although the desired information  is arguably not purely syntactic, we refer to its reconstruction as a \emph{syntactic annotation}. To represent syntactic annotations in a uniform way, we define a set of \emph{syntactic annotation keys} and, for each key, a set of possible \emph{syntactic annotation values}. A complete list of all syntactic annotation keys along with possible annotation values can be found in Table~\ref{tab:annotations}; exemplary syntactic annotations for vertices of an AMR graph are shown in Figure~\ref{fig:syntactic-annotation}.\footnote{For the annotation key \anno{POS}, only some exemplary values are shown in Table~\ref{tab:annotations}. A list of common POS tags can be found at \url{www.ling.upenn.edu/courses/Fall_2003/ling001/penn_treebank_pos.html}. We use, however, only a small subset of these POS tags (see Section~\ref{TRAINING:SyntacticAnnotations}).} We denote the set of all syntactic annotation keys by $\mathcal{K}_\text{syn} = \{ \anno{POS}, \anno{DENOM}, \anno{TENSE}, \anno{NUMBER}, \anno{VOICE} \}$\label{los:k_syn} and for each syntactic annotation key $k \in \mathcal{K}_\text{syn}$, we refer to the set of possible annotation values as $\mathcal{V}_k$.\label{los:v_1} The set of all syntactic annotation values is denoted by $\mathcal{V}_\text{syn} = \bigcup_{k \in \mathcal{K}_\text{syn}} \mathcal{V}_k$.\label{los:v_syn}

\begin{definition}[Syntactic annotation]
Let $G = (V,E,L, \prec)$ be a graph and let $v \in V$.
A \emph{syntactic annotation (for $v$)} is a mapping $\alpha \colon \mathcal{K}_\text{syn} \rightarrow \mathcal{V}_\text{syn}$ where for each $k \in \mathcal{K}_\text{syn}$, it holds that $\alpha(k) \in \mathcal{V}_k$. The set of all syntactic annotations is denoted by $\mathcal{A}_\text{syn}$. \label{los:a_syn}
\end{definition}

It is important to note that syntactic annotations as introduced here are strongly biased towards the English language. However, the underlying principle can easily be transfered to many other natural languages by revising the sets $\mathcal{K}_\text{syn}$ and $\mathcal{V}_\text{syn}$ of syntactic annotation keys and values. For example, adapting syntactic annotations to the German language may require the introduction of an additional key $\anno{CASE}$ to reflect the German case system and the redefinition of $\mathcal{V}_\anno{DENOM}$ to represent the set of German denominators.

\begin{table}
\centering
\bgroup
\def\arraystretch{1.5}
\small
\begin{tabular}{|l|l|l|}
\hline
\textbf{Key} & \textbf{Values} & \textbf{Meaning} \\ 
\hline
\anno{POS} & $\{\text{VB, NN, JJ, CC,} \ldots, \text{--} \}$ & The POS tag assigned to $v$ \\
\anno{DENOM} & $\{ \text{the, a, --} \}$ & The denominator assigned to $v$ \\
\anno{TENSE} & $\{ \text{past, present, future, --} \}$ & The tense assigned to $v$ \\
\anno{NUMBER} & $\{ \text{singular, plural, --} \}$ & The number assigned to $v$ \\
\anno{VOICE} & $\{ \text{passive, active, --} \}$ & The voice assigned to $v$ \\
\hline
\end{tabular}
\caption{Syntactic annotations used by our transition-based generator. For each syntactic annotation key $k \in \mathcal{K}_\text{syn}$, the set of possible values $\mathcal{V}_k$ is given and the meaning of $\alpha(k)$ for some vertex $v$ is briefly explained.}
\label{tab:annotations}
\egroup
\end{table}

As discussed in Section~\ref{PRELIMINARIES:AMR}, there is often not just one reasonable syntactic annotation for the nodes of an AMR graph. To account for this in our generator, we simply consider multiple syntactic annotations per node and assign probabilities to them. For this purpose, let $G = (V, E, L, \prec)$ be a graph and let $\alpha \colon \mathcal{K}_\text{syn} \rightarrow \mathcal{V}_\text{syn}$ be a syntactic annotation for some node $v \in V$. Furthermore, let $k_1, \ldots, k_n$ be some enumeration of $\mathcal{K}_\text{syn}$. We denote by $P( \alpha \mid G, v)$ the probability of $\alpha$ being the correct annotation for $v$ given $G$ and $v$. As a syntactic annotation, like any other function, is fully defined by the values it assigns to each element of its domain, we may write
\begin{equation}
P(\alpha \mid G, v) = P( \alpha(k_1), \ldots, \alpha(k_n) \mid G, v) \,, \label{equ:syn1}
\end{equation}
i.e. the probability of $\alpha$ being the correct syntactic annotation for $v$ is equal to the joint probability of $\alpha(k_i)$ being the correct annotation value for key $k_i$ at vertex $v$ for all $i \in [n]$. 
We note that it might be useful not to look at the syntactic annotations of all nodes in $V$ independently; for example, the tense assigned to a node depends to a large extent on the tense assigned to its predecessors. However, ignoring these dependencies allows us to handle syntactic annotations much more efficiently as we can store the $m$-best syntactic annotations $\alpha_1, \ldots, \alpha_m$ for each node $v \in V$ independently.

Using the general product rule, we can transform Eq.~(\ref{equ:syn1}) into
\begin{equation}
\begin{split}
& P(\alpha(k_1), \ldots, \alpha(k_n) \mid G, v ) \\
& = P(\alpha(k_1) \mid G, v ) \cdot P(\alpha(k_2) \mid G, v, \alpha(k_1)) \cdot \ldots \cdot P(\alpha (k_n) \mid G, v, \alpha(k_1), \ldots, \alpha(k_{n-1}))
\end{split}
\label{equ:syn2}
\end{equation}
and as the above holds for any enumeration $k_1, \ldots, k_m$ of $\mathcal{K}_\text{syn}$, we are free to choose 
\[
k_1 = \anno{POS}\quad k_2 = \anno{NUMBER}\quad k_3 = \anno{DENOM}\quad k_4 = \anno{VOICE}\quad k_5 = \anno{TENSE}\,.
\] Importantly, there are several strong dependencies between the values assigned to different syntactic annotation keys $k_i \in \mathcal{K}_\text{syn}$ by $\alpha$. For instance, a word that is not a verb should have no tense or voice assigned to it (i.e. $\alpha(\anno{TENSE}) = \alpha(\anno{VOICE}) = \text{--}$) and a plural noun can not have the article ``a'' as a denominator. 
On the other hand, it seems reasonable to assume that, for example, the tense of a verb is independent of its voice. In other words, $\alpha(\anno{TENSE})$ is conditionally independent of $\alpha(\anno{VOICE})$ given $\alpha(\anno{POS})$. We formulate several such conditional independence assumptions, allowing us to rewrite Eq.~(\ref{equ:syn2}) as follows:
\begin{equation}
\begin{split}
P(\alpha(k_1), \ldots, \alpha(k_n) \mid G,& v ) = P(\alpha(\anno{POS}) \mid G,v ) \cdot P(\alpha(\anno{NUMBER}) \mid G, v, \alpha(\anno{POS})) \\
& \cdot P(\alpha(\anno{DENOM}) \mid G, v, \alpha(\anno{POS}), \alpha(\anno{NUMBER})) \\
& \cdot P(\alpha(\anno{VOICE}) \mid G, v, \alpha(\anno{POS}))
\cdot P(\alpha(\anno{TENSE}) \mid G, v, \alpha(\anno{POS}))\,.
\end{split} \label{equ:syn3}
\end{equation}
Finally, we estimate the above conditional probabilities using maximum entropy models $p_k$ for each $k \in \mathcal{K}_\text{syn}$ and arrive at
\begin{equation}
\begin{split}
P(\alpha \mid G, v ) =\ & p_\anno{POS}(\alpha(\anno{POS}) \mid G, v ) \cdot p_\anno{NUMBER}(\alpha(\anno{NUMBER}) \mid G, v, \alpha(\anno{POS})) \\
& \cdot p_\anno{DENOM}(\alpha(\anno{DENOM}) \mid G, v, \alpha(\anno{POS}), \alpha(\anno{NUMBER})) \\
& \cdot p_\anno{VOICE}(\alpha(\anno{VOICE}) \mid G, v, \alpha(\anno{POS}))
\cdot p_\anno{TENSE}(\alpha(\anno{TENSE}) \mid G, v, \alpha(\anno{POS}))\,.
\end{split} \label{equ:syn4}
\end{equation}
Both the features extracted from $G$, $v$ and $\alpha$ to obtain the maximum entropy models $p_k$ and the training of these models is discussed in Section~\ref{GENERATION:Training}.
As a final modification to the above equation, we introduce weights $w_k \in \mathbb{R}$ for each $k \in \mathcal{K}_\text{syn}$ and we raise each conditional probability $p_k$ to the $w_k$-th power; for example, we replace $p_\anno{POS}(\alpha(\anno{POS}) \mid G, v )$ by $p_\anno{POS}(\alpha(\anno{POS}) \mid G, v )^{w_\anno{POS}}$. We denote the value obtained from $P(\alpha \mid G, v)$ through introducing these weights by $P^\text{w}(\alpha \mid G, v)$. While this modification is not mathematically justified, it allows our generator to decide how important it is that an applied transition actually complies with the values predicted by each of the above models.
We view the weights $w_k$ as hyperparameters; how they are obtained is described in Section~\ref{GENERATION:HyperparameterOptimization}. 

\subsection{Transition System}
\label{GENERATION:TransitionSystem}

We now define the core part of our generator, the transition system $S_\text{AMR}$. The two main tasks to be performed by this transition system are the restructuring of the input AMR graph -- for example by inserting and removing vertices or edges, merging multiple vertices into a single one or changing the order among them -- and the determination of some additional information. The latter includes, among others, each node's syntactic annotation and its realization, i.e. a continuous sequence of words by which the node is represented in the final output of our generator. To store all additional information obtained for each node in a unified manner, we introduce the notion of an \emph{annotation function} that generalizes the concept of syntactic annotations. We denote by 
\[ 
\mathcal{K} = \mathcal{K}_\text{syn} \cup \{ \anno{REAL}, \anno{DEL}, \anno{INS-DONE}, \anno{LINK}, \anno{SWAPS}, \anno{INIT-CONCEPT} \} 
\] \label{los:k}the set of all \emph{annotation keys}. For each annotation key $k \in \mathcal{K} \setminus \mathcal{K}_\text{syn}$, the set of corresponding \emph{annotation values} $\mathcal{V}_k$ is shown in Table~\ref{tab:trans-annotations};\label{los:v_2} for syntactic annotations, we refer to Table~\ref{tab:annotations}. While the meaning of some annotation keys might be unclear at this moment, it will become clear during the discussion of $S_\text{AMR}$. We denote by $\mathcal{V} = \bigcup_{k \in \mathcal{K}} \mathcal{V}_k$\label{los:v} the set of all possible annotation values.

\begin{table}
\centering
\bgroup
\def\arraystretch{1.5}
\small
\begin{tabularx}{\textwidth}{|l|l|X|}
\hline
\textbf{Key} & \textbf{Values} & \textbf{Meaning} \\ 
\hline
\anno{REAL} & $\Sigma_\text{E}^*$ & The realization of $v$, i.e. the sequence of words that represents it in the generated sentence \\
\anno{DEL} & $\{0, 1\}$ & A flag indicating whether $v$ needs to be deleted \\
\anno{INS-DONE} & $\{0, 1\}$ & A flag indicating whether child insertion for $v$ is complete\\
\anno{LINK} & $V$ & The original vertex, if $v$ is a copy \\
\anno{SWAPS} & $\mathbb{Z}$ & The number of times $v$ has been swapped up ($ \rho(\anno{SWAPS})(v) > 0$) or down ($ \rho(\anno{SWAPS})(v) < 0$) \\
\anno{INIT-CONCEPT} & $L_\text{C}$ & The concept initially assigned to $v$, if it is overwritten through a \textsc{Merge} transition \\
\hline
\end{tabularx}
\caption{Additional annotations used in the generation pipeline, assuming an AMR graph $G = (V, E, L , \prec)$. For each annotation key $k \in \mathcal{K} \setminus \mathcal{K}_\text{syn}$, the set of possible values $\mathcal{V}_k$ is given and the meaning of $\rho(k)(v)$ for $v \in V$ is briefly explained.}
\label{tab:trans-annotations}
\egroup
\end{table}

\begin{definition}[Annotation function] Let $V$ be a set of vertices. An \emph{annotation function for $V$} is a function $\rho \colon \mathcal{K} \rightarrow ( V \pfun \mathcal{V})$ such that for all $k \in \mathcal{K}$ and for all $v \in \text{dom}(\rho(k))$, it holds that $\rho(k)(v) \in \mathcal{V}_k$.
\end{definition}
To give an example, an annotation function $\rho$ where
\[
\rho(\anno{POS})(v_1) = \text{NN} \qquad \rho(\anno{REAL})(v_2) = \text{at least}
\] 
would indicate that the POS tag assigned to node $v_1$ is \text{NN} and that the realization of $v_2$ is the sequence ``at least''. As values are assigned to annotation keys incrementally during the generation process through application of transitions, we allow $\rho(k)$ to be partial for all $k \in \mathcal{K}$. Building up on the concept of annotation functions, we may now define the set of configurations used by our generator.

\begin{definition}[Configuration for AMR generation] \label{los:c_amr}
A \emph{configuration for AMR generation} is a tuple $c = (G, \sigma, \beta, \rho)$ where
\begin{enumerate}
\item $G = (V,E,L,\prec)$ is a rooted, acyclic $(L_\text{R} \cup \{ \star \}, L_\text{C} \cup \Sigma_\text{E}^* )$-graph with $\star \notin L_\text{R}$ being a special \emph{placeholder edge label};
\item $\sigma = (\sigma_1, \ldots, \sigma_n) \in V^*$ is a finite sequence of nodes (\emph{node buffer}) such that for all $v \in V$, there is at most one $i \in [n]$ with $\sigma_i = v$;
\item $\beta = (\beta_1, \ldots, \beta_m) \in \ch{\sigma_1}^*$ is a finite sequence of nodes (\emph{child buffer}) such that for all $v \in \ch{\sigma_1}$, there is at most one $i \in [m]$ with $\beta_i = v$;
\item $\rho\colon \mathcal{K} \rightarrow ( V' \pfun \mathcal{V})$ is an annotation function for some $V' \supseteq V$.
\end{enumerate}
The set of all configurations for AMR generation is denoted by $C_\text{AMR}$. 
\end{definition}

This definition is inspired by \citet{wang2015transition} where configurations are defined as triples consisting of a node buffer, an edge buffer and a graph. The underlying idea is as follows: Given a configuration $c \in C_\text{AMR}$, the transition to be applied next is to modify primarily the top element of the node buffer, $\sigma_1$, and, if $\beta \neq \varepsilon$, its child $\beta_1$. If this application completes the required modifications at node $\sigma_1$ (or $\beta_1$), the latter is removed from $\sigma$ (or $\beta$). That way, each node contained within $\sigma$ and $\beta$ gets processed one at a time until they are both empty.

\begin{definition}[$S_\text{AMR}$] The tuple \label{los:s_amr}
$
S_\text{AMR} = (C_\text{AMR}, T_\text{AMR}, \ct, \cs, \cf)
$
is a transition system for $(\mathcal{G}_\text{AMR}, \Sigma_\text{E}^*)$ where

\begin{enumerate}

\item $ \begin{aligned}[t]
        T_\text{AMR} = & \ \{ \textsc{Delete-Reentrance-}(v,l) \mid v \in V, l \in L_\text{R} \} \\
        \cup & \ \{ \textsc{Merge-}(l,p) \mid l \in \Sigma_\text{E}^*, p \in \mathcal{V}_\anno{POS} \} \\
        \cup & \ \{\textsc{Swap, Delete, Keep, No-Insertion} \} \\
        \cup & \ \{ \textsc{Realize-}(w, \alpha) \mid w \in \Sigma_\text{E}^*, \alpha \in \mathcal{A}_\text{syn} \} \\
        \cup & \ \{ \textsc{Insert-\textasteriskcentered-}(w,p) \mid \text{\textasteriskcentered} \in \{ \textsc{Child}, \textsc{Between} \}, w \in \Sigma_\text{E}, p \in \{ \textsf{left}, \textsf{right} \} \} \\
        \cup & \ \{  \textsc{Reorder-}(v_1, \ldots, v_n) \mid v_i \in V, i \in [n], n \in \mathbb{N} \} \text{ for any set } V \text{;}
         \end{aligned}$
\item $\ct = \{ (G, \varepsilon, \varepsilon, \rho) \in C_\text{AMR} \}$ is the set of all configurations with both an empty node buffer and an empty child buffer;
\item $\cs(G) = (G, \sigma_G, \varepsilon, \rho)$ for all $G \in \mathcal{G}_\text{AMR}$ where $\sigma_G$ is some bottom-up traversal of all nodes in $G$ and $\rho = \{ (k, \emptyset) \mid k \in \mathcal{K} \}$; 
\item $\cf(c) = \text{yield}_{\rho(\anno{REAL})}(G)$ for all $c = (G, \sigma, \beta, \rho) \in C_\text{AMR}$ if $G = (V, E, L, \prec)$ is totally ordered and $V \subseteq \text{dom}(\rho(\anno{REAL}))$; otherwise, $\cf(c)$ is undefined. \qedhere
\end{enumerate}
\end{definition}

Before looking into the transitions contained within $T_\text{AMR}$, it is worth nothing that there is a strong connection between some of the transitions used by our generator and the transitions used by the CAMR parser of \citet{wang2015transition}. For example, \textsc{Delete-Reentrance} can be seen as a counterpart of the \textsc{Insert-Reentrance} transition used in CAMR and \textsc{Merge}, \textsc{Swap} and \textsc{Delete} transitions are used in both systems. However, other transitions such as \textsc{Reorder} have no direct counterpart in CAMR.

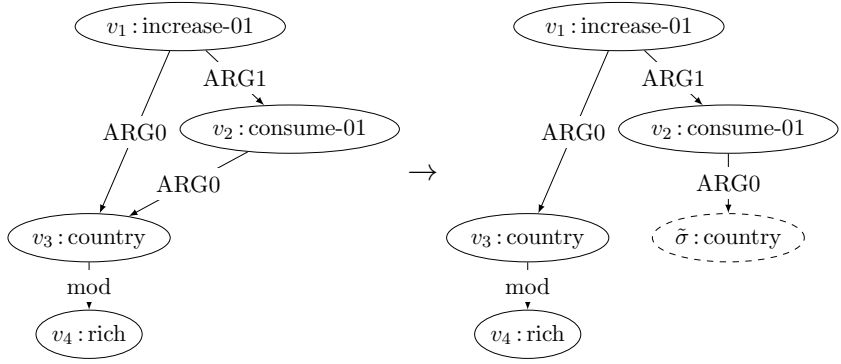
\begin{figure}[t]
\centering
\scalebox{0.8} {
\begin{tikzpicture}[baseline=(current bounding box.center)]
\tikzstyle{amr-node}=[shape=ellipse,draw, inner sep=0.2, minimum height=0.8cm, text height=1.5ex, text depth=.25ex, minimum width=1cm]
\tikzstyle{text-node}=[text height=1.5ex, text depth=.25ex, align=center]

	\node (PAD2) at (-2.5,-1){};
	\node (PAD2) at (2.5,-1){};

    \node[amr-node] (amr-increase) at (0,-0.3) {$v_1$\,:\,increase-01};
    \node[amr-node] (amr-country) at (-1.5,-3.8) {$v_3$\,:\,country};
    \node[amr-node] (amr-consume) at (1.8,-2) {$v_2$\,:\,consume-01};
    \node[amr-node] (amr-rich) at (-1.5,-5.4) {$v_4$\,:\,rich};

    \path [-latex](amr-increase) edge node[fill=white] {ARG0} (amr-country);
    \path [-latex](amr-increase) edge node[fill=white] {ARG1} (amr-consume);
    \path [-latex](amr-consume) edge node[fill=white] {ARG0} (amr-country);
    \path [-latex](amr-country) edge node[fill=white] {mod} (amr-rich);
       
\end{tikzpicture}
{\Large $\rightarrow$}
\begin{tikzpicture}[baseline=(current bounding box.center)]
\tikzstyle{amr-node}=[shape=ellipse,draw, inner sep=0.2, minimum height=0.8cm, text height=1.5ex, text depth=.25ex, minimum width=1cm]
\tikzstyle{text-node}=[text height=1.5ex, text depth=.25ex, align=center]

	\node (PAD2) at (-2.5,-1){};
	\node (PAD2) at (2.5,-1){};

    \node[amr-node] (amr-increase) at (0,-0.3) {$v_1$\,:\,increase-01};
    \node[amr-node] (amr-country) at (-1.5,-3.8) {$v_3$\,:\,country};
    \node[amr-node, dashed] (amr-country-link) at (1.8,-3.8) {$\tilde{\sigma}$\,:\,country};
    \node[amr-node] (amr-consume) at (1.8,-2) {$v_2$\,:\,consume-01};
    \node[amr-node] (amr-rich) at (-1.5,-5.4) {$v_4$\,:\,rich};

    \path [-latex](amr-increase) edge node[fill=white] {ARG0} (amr-country);
    \path [-latex](amr-increase) edge node[fill=white] {ARG1} (amr-consume);
    \path [-latex](amr-consume) edge node[fill=white] {ARG0} (amr-country-link);
    \path [-latex](amr-country) edge node[fill=white] {mod} (amr-rich);
       
\end{tikzpicture}
}
\caption{\textsc{Delete-Reentrance}-$(v_2,\text{ARG0})$ transition applied to the node with label ``country''; the new node $\tilde{\sigma}$ is indicated by a dashed border. The reference realization of this partial AMR graph is ``rich countries increase their consumption''.}
\label{fig:delete-reentrance}
\end{figure}

For the remainder of this section, let $G = (V, E, L, \prec)$ be an arbitrary rooted acyclic graph. If a node $v \in V$ has exactly one parent, we denote the latter by $p_v$. As it may be necessary to insert new nodes during the generation process, we make use of a set $V_\text{ins} = \{ \tilde{\sigma}_i \mid i \in \mathbb{N} \}$ of \emph{insertable nodes} for which we demand that $V \cap V_\text{ins} = \emptyset$. \label{los:v_ins}
For each transition $t \in T_\text{AMR}$, we formally define both the actual mapping $t: C_\text{AMR} \pfun C_\text{AMR}$ and $\text{dom}(t)$, the set of configurations for which $t$ is defined. In addition, we provide a textual description and briefly justify the necessity of each class of transitions. For the more complex transitions, exemplary applications are shown in Figures~\ref{fig:delete-reentrance}~to~\ref{fig:insert-between}. All AMR graphs and realizations shown in these examples are taken directly from the LDC2014T12 corpus (see Section~\ref{PRELIMINARIES:Corpora}) to demonstrate the actual need for the corresponding transitions. 

The transitions used by our generator are defined as follows:
\bgroup

\begin{itemize}
\renewcommand{\arraystretch}{\tbarraystretch}
\item
\textsc{Delete-Reentrance-}$(v,l)$ ($v \in V$, $l \in L_\text{R}$)

\begin{mdframed}
\begin{tabularx}{\linewidth}{@{}l X@{}}
Mapping: & 
 $(G, \sigma_1{:}\sigma, \varepsilon, \rho) \mapsto (G', \sigma_1{:}\tilde{\sigma}{:}\sigma, \varepsilon, \rho[\anno{LINK}(\tilde{\sigma}) = \sigma_1])$ where $\tilde{\sigma} \in V_\text{ins} \setminus V$ is some new node and
{
\setlength{\belowdisplayskip}{-1em}
\setlength{\belowdisplayshortskip}{0pt} 
\begin{align*}
G' & = (V \cup \{ \tilde{\sigma} \}, E', L \cup \{ (\tilde{\sigma}, L(\sigma_1)) \}, \prec) \\[\tblinepad]
E' & = E \setminus \{ (v, l, \sigma_1) \} \cup \{ (v, l, \tilde{\sigma}) \}\,.
\end{align*}} \\[\tbcolumnpad]

Domain: & $\{(G, \sigma_1{:}\sigma, \varepsilon, \rho) \in C_\text{AMR} \mid (v, l, \sigma_1) \in \text{in}_G(\sigma_1) \wedge  |\text{in}_G(\sigma_1)| \geq 2 \}$ \\
\end{tabularx}
\end{mdframed}

This transition removes the edge $(v, l, \sigma_1)$; it is thus only applicable if such an edge exists and $\sigma_1$ has at least one more incoming edge. As the deleted edge may contain useful information for the generation process, a new node $\tilde{\sigma}$ is added as a copy of $\sigma_1$ and connected to $v$. Further handling of this copy must be decided in separate transitions; therefore, $\tilde{\sigma}$ is inserted into the node buffer directly after $\sigma_1$.   

Through application of \textsc{Delete-Reentrance}, the input is stepwise converted into a tree: Whenever a node $\sigma_1$ has multiple incoming edges, all but one of these edges are successively removed using this transition. An example can be seen in Figure~\ref{fig:delete-reentrance}, where 
one of the incoming edges for the node labeled ``country'' gets removed and a copy of said node is added to $G$; the information that $\tilde{\sigma}$ is a copy of $v_3$ is stored in $\rho$ by setting $\rho(\anno{LINK})(\tilde{\sigma}) = v_3$. To obtain the desired realization, $\tilde{\sigma}$'s realization must then be set to ``their'' in a subsequent transition step.
\end{itemize}

\noindent\begin{minipage}{\textwidth}
\renewcommand{\arraystretch}{\tbarraystretch}
\begin{itemize}
\item \textsc{Merge}-$(l,p)$ ($l \in \Sigma_\text{E}^*, p \in \mathcal{V}_\anno{POS}$)
\begin{mdframed}
\begin{tabularx}{\linewidth}{@{}l X@{}}
Mapping: & $(G, \sigma_1{:}\sigma, \varepsilon, \rho) \mapsto (G', \sigma, \varepsilon, \rho')$ where $G' = (V \setminus \{\sigma_1\}, E', L', \prec)$ and 
{
\setlength{\belowdisplayskip}{-1em}
\setlength{\belowdisplayshortskip}{0pt} 
\begin{align*}
E' =\ & E \setminus \{ (v_1, l, v_2) \mid \sigma_1 \in \{v_1, v_2 \}, l \in L_\text{R} \} \\
& \cup \{ (p_{\sigma_1}, l, v) \mid (\sigma_1, l, v) \in E \} \\[\tblinepad]
L' =\ & L \setminus \{ (\sigma_1, L(\sigma_1)), (p_{\sigma_1}, L(p_{\sigma_1})) \} \cup \{ (p_{\sigma_1}, l) \} \\[\tblinepad]
\rho' =\ & \rho[\anno{POS}(p_{\sigma_1}) \mapsto p, \anno{INIT-CONCEPT}(p_{\sigma_1}) \mapsto L(p_{\sigma_1})]
\end{align*}
}
\\[\tbcolumnpad]
Domain: & $\{(G, \sigma_1{:}\sigma, \varepsilon, \rho) \in C_\text{AMR} \mid |\text{in}(\sigma_1)| = 1 \wedge \sigma_1 \notin \text{dom}(\rho(\anno{DEL})) \}$ \\
\end{tabularx}
\end{mdframed}

This transition merges the top element of the node buffer, $\sigma_1$, and its parent $p_{\sigma_1}$ into a single node with a new vertex label $l \in \Sigma_\text{E}^*$ and POS tag $p \in \mathcal{V}_\anno{POS}$; it is only applicable if $\sigma_1$ has exactly one incoming edge. All outgoing edges previously connected to $\sigma_1$ get reconnected to $p_{\sigma_1}$; the initial concept of $p_{\sigma_1}$ is preserved in $\rho(\anno{INIT-CONCEPT})(p_{\sigma_1})$.

Whenever two nodes are realized by a mutual word or their realizations share at least one common word, a \textsc{Merge} transition must be applied to fuse both nodes. An example can be seen in Figure~\ref{fig:merge} where the nodes labeled ``early'' and ``more'' are realized by the single word ``earlier'' in the reference realization.
\end{itemize}
\end{minipage}

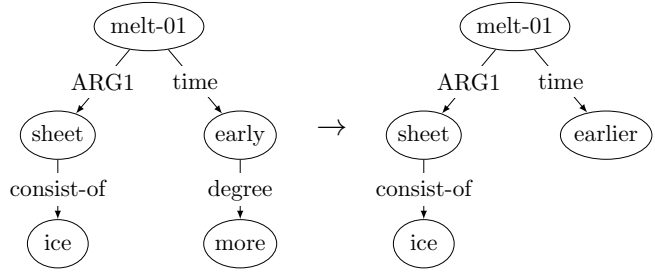
\begin{figure}[]
\centering
\scalebox{0.8} {
\begin{tikzpicture}[baseline=(current bounding box.center)]
\tikzstyle{amr-node}=[shape=ellipse,draw, inner sep=0.2, minimum height=0.8cm, text height=1.5ex, minimum width=1cm, text depth=.25ex]
\tikzstyle{amr-edge}=[text height=1.5ex, text depth=.25ex, fill=white]
\tikzstyle{text-node}=[text height=1.5ex, text depth=.25ex, align=center]

	\node (PAD2) at (-2.5,0){};
	\node (PAD2) at (2.5,0){};

    \node[amr-node] (amr-melt) at (0,0) {melt-01};
    \node[amr-node] (amr-sheet) at (-1.5,-1.8) {sheet};
    \node[amr-node] (amr-ice) at (-1.5,-3.6) {ice};
    \node[amr-node] (amr-early) at (1.5,-1.8) {early};
    \node[amr-node] (amr-more) at (1.5,-3.6) {more};

    \path [-latex](amr-melt) edge node[fill=white] {ARG1} (amr-sheet);
    \path [-latex](amr-melt) edge node[fill=white] {time} (amr-early);
    \path [-latex](amr-early) edge node[amr-edge] {degree} (amr-more);
    \path [-latex](amr-sheet) edge node[amr-edge] {consist-of} (amr-ice);
       
\end{tikzpicture}
{\Large $\rightarrow$}
\begin{tikzpicture}[baseline=(current bounding box.center)]
\tikzstyle{amr-node}=[shape=ellipse,draw, inner sep=0.2, minimum height=0.8cm, text height=1.5ex, text depth=.25ex, minimum width=1cm]
\tikzstyle{amr-edge}=[text height=1.5ex, text depth=.25ex, fill=white]
\tikzstyle{text-node}=[text height=1.5ex, text depth=.25ex, align=center]

	\node (PAD2) at (-2.5,0){};
	\node (PAD2) at (2.5,0){};

    \node[amr-node] (amr-melt) at (0,0) {melt-01};
    \node[amr-node] (amr-sheet) at (-1.5,-1.8) {sheet};
    \node[amr-node] (amr-ice) at (-1.5,-3.6) {ice};
    \node[amr-node] (amr-earlier) at (1.5,-1.8) {earlier};

    \path [-latex](amr-melt) edge node[fill=white] {ARG1} (amr-sheet);
    \path [-latex](amr-melt) edge node[fill=white] {time} (amr-earlier);
    \path [-latex](amr-sheet) edge node[amr-edge] {consist-of} (amr-ice);
       
\end{tikzpicture}
}
\caption{\textsc{Merge}-(earlier,JJ) transition applied to the node with label ``more''. The reference realization of this partial AMR graph is ``the ice sheet has melted earlier''.}
\label{fig:merge}
\end{figure}

\noindent\begin{minipage}{\textwidth}
\renewcommand{\arraystretch}{\tbarraystretch}
\begin{itemize}
\item
\textsc{Swap}
\begin{mdframed}
\begin{tabularx}{\linewidth}{@{}l X@{}}
Mapping: & $(G, \sigma_1{:}\sigma, \varepsilon, \rho) \mapsto ((V, E', L, \prec), p_{\sigma_1}{:}\sigma_1{:}(\sigma \setminus \{ p_{\sigma_1} \}), \varepsilon, \rho')$ where 
{\begin{align*}
\rho' =\ & \rho[\anno{SWAPS}(\sigma_1) \mapsto \anno{S}(\sigma_1)+1,
\anno{SWAPS}(p_{\sigma_1}) \mapsto \anno{S}(p_{\sigma_1})-1 ] \\[\tblinepad]
\anno{S}(v) =\ & \begin{cases} \rho(\anno{SWAPS})(v) & \text{ if } v \in \text{dom}(\rho(\anno{SWAPS})) \\ 0 & \text{ otherwise} \end{cases} \\[\tblinepad]
E' =\ & E \setminus ( \{ (p_{\sigma_1}, l_{\sigma_1}, \sigma_1) \} \cup \{ (v, l, p_{\sigma_1}) \mid v \in V, l \in L_\text{R} \} ) \\
& \cup \{ ({\sigma_1}, l_{\sigma_1}^{-1}, p_{\sigma_1})\} \cup \{ (v, l, \sigma_1) \mid (v, l, p_{\sigma_1}) \in E \}
\end{align*}}
and $l_{\sigma_1}$ denotes the label of the edge connecting $p_{\sigma_1}$ and $\sigma_1$. \\[\tbcolumnpad]
Domain: & $\{(G, \sigma_1{:}\sigma, \varepsilon, \rho) \in C_\text{AMR} \mid |\text{in}(\sigma_1)| = 1 \wedge \sigma_1 \notin \text{dom}(\rho(\anno{DEL})) \}$ \\
\end{tabularx}
\end{mdframed}

This transition swaps the top node of the node buffer, $\sigma_1$, with its parent node. It is therefore only applicable if $\sigma_1$ has exactly one parent node $p_{\sigma_1}$ and there is only one edge connecting $\sigma_1$ and $p_{\sigma_1}$. Both the direction and the label of this single incoming edge get inverted; all parents of $p_{\sigma_1}$ get disconnected from $p_{\sigma_1}$ and reconnected to $\sigma_1$. The information that $\sigma_1$ and $p_{\sigma_1}$ were swapped is stored in $\rho$ by incrementing $\rho(\anno{SWAPS})(\sigma_1)$ and decrementing $\rho(\anno{SWAPS})(p_{\sigma_1})$.

\textsc{Swap} transitions are required due to the projectivity of $\text{yield}_{\rho(\anno{REAL})}$ (see Definition~\ref{def:yield}). For instance, consider the AMR graph shown in Figure~\ref{fig:swap}. If we assume that the vertices labeled ``possible'', ``make-05'', ``Hallmark'' and ``fortune'' are realized by ``could'', ``make'', ``Hallmark'' and ``a fortune'', respectively, then for the graph on the left, there is no order $\prec$ such that $\text{yield}_{\rho(\anno{REAL})}$ produces the desired phrase ``Hallmark could make a fortune''. This is the case because $\rho(\anno{REAL})(v_1)$ cannot occur between $\rho(\anno{REAL})(v_3)$ and $\rho(\anno{REAL})(v_2)$ as $v_1$ is not a successor of $v_2$. After swapping the node labeled ``possible'' with the node labeled ``make-05'', such an order can easily be found, namely ${\prec} = \{ (v_3, v_1), (v_1, v_2), (v_2, v_4) \}^+$.
\end{itemize}
\end{minipage}

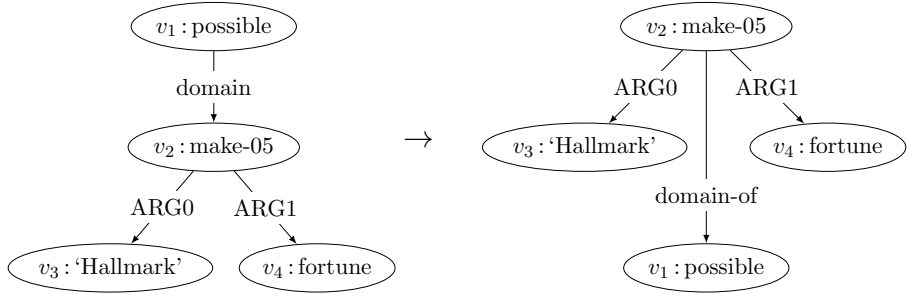
\begin{figure}[]
\centering
\scalebox{0.8} {
\begin{tikzpicture}[baseline=(current bounding box.center)]
\tikzstyle{amr-node}=[anchor=base, shape=ellipse,draw, inner sep=0.2, minimum height=0.8cm, text height=1.5ex, text depth=.25ex]
\tikzstyle{text-node}=[text height=1.5ex, text depth=.25ex, align=center]

	\node (PAD2) at (-2.5,0){};
	\node (PAD2) at (2.5,0){};

    \node[amr-node] (amr-possible) at (0,0) {$v_1$\,:\,possible};
    \node[amr-node] (amr-make) at (0,-2) {$v_2$\,:\,make-05};
    \node[amr-node] (amr-hallmark) at (-1.7,-4) {$v_3$\,:\,`Hallmark'};
    \node[amr-node] (amr-fortune) at (1.7,-4) {$v_4$\,:\,fortune};

    \path [-latex](amr-possible) edge node[fill=white] {domain} (amr-make);
    \path [-latex](amr-make) edge node[fill=white] {ARG0} (amr-hallmark);
    \path [-latex](amr-make) edge node[fill=white] {ARG1} (amr-fortune);
       
\end{tikzpicture}
{\Large $\rightarrow$ \quad}
\begin{tikzpicture}[baseline=(current bounding box.center)]
\tikzstyle{amr-node}=[shape=ellipse,draw, inner sep=0.2, minimum height=0.8cm, text height=1.5ex, text depth=.25ex]
\tikzstyle{text-node}=[text height=1.5ex, text depth=.25ex, align=center]

	\node (PAD2) at (-2.5,0){};
	\node (PAD2) at (2.5,0){};

    \node[amr-node] (amr-possible) at (0,-4) {$v_1$\,:\,possible};
    \node[amr-node] (amr-make) at (0,0) {$v_2$\,:\,make-05};
    \node[amr-node] (amr-hallmark) at (-2,-2) {$v_3$\,:\,`Hallmark'};
    \node[amr-node] (amr-fortune) at (2,-2) {$v_4$\,:\,fortune};

    \path [-latex](amr-make) edge node[fill=white, near end] {domain-of} (amr-possible);
    \path [-latex](amr-make) edge node[fill=white] {ARG0} (amr-hallmark);
    \path [-latex](amr-make) edge node[fill=white] {ARG1} (amr-fortune);
       
\end{tikzpicture}
}
\caption{\textsc{Swap} transition applied to the node labeled ``make-05''; the edge label ``domain'' is converted into its inverse, ``domain-of''. The reference realization of this partial AMR graph is ``Hallmark could make a fortune''.}
\label{fig:swap}
\end{figure}

\noindent\begin{minipage}{\textwidth}
\renewcommand{\arraystretch}{\tbarraystretch}
\begin{itemize}
\item
\textsc{Delete}

\begin{mdframed}
\begin{tabularx}{\linewidth}{@{}l X@{}}
Mapping: & $(G, \sigma_1{:}\sigma, \varepsilon, \rho) \mapsto (G, \sigma_1{:}\sigma, \varepsilon, \rho[\anno{DEL}(\sigma_1) \mapsto 1, \anno{REAL}(\sigma_1) \mapsto \varepsilon])$ \\[\tbcolumnpad]
Domain: & $\{(G, \sigma_1{:}\sigma, \varepsilon, \rho) \in C_\text{AMR} \mid |\text{in}(\sigma_1)| = 1 \wedge \sigma_1 \notin \text{dom}(\rho(\anno{DEL})) \}$ \\
\end{tabularx}
\end{mdframed}

Although the name may suggest otherwise, this transition does not directly remove node $\sigma_1$ from $G$. Instead, an application of \textsc{Delete} merely indicates that node $\sigma_1$ is not represented in the generated sentence by setting the \anno{DEL} flag to $1$ and the realization to $\varepsilon$. The reason for not directly deleting $\sigma_1$ is that although it is not represented in the generated sentence, it may still provide useful information with regard to the realization and ordering of its child nodes. 

An exemplary application of \textsc{Delete} is shown in Figure~\ref{fig:delete} where it is applied to the node with label ``mass-quantity'' as the latter has no representation in the reference realization.

\end{itemize}
\end{minipage} \\[\baselineskip]

\begin{figure}[]
\centering
\scalebox{0.8} {
\begin{tikzpicture}[baseline=(current bounding box.center)]
\tikzstyle{amr-node}=[shape=ellipse,draw, inner sep=0.2, minimum height=0.8cm, text height=1.5ex, text depth=.25ex, minimum width=1cm]
\tikzstyle{amr-edge}=[text height=1.5ex, text depth=.25ex, fill=white]
\tikzstyle{text-node}=[text height=1.5ex, text depth=.25ex, align=center]

	\node (PAD2) at (-2.5,0){};
	\node (PAD2) at (2.5,0){};
	
    \node[amr-node] (amr-weigh) at (0,0) {weigh-01};
    \node[amr-node] (amr-mass-quantity) at (0,-2) {mass-quantity};
    \node[amr-node] (amr-number) at (-1.5,-4) {$1.1$};
    \node[amr-node] (amr-kilogram) at (1.5,-4) {kilogram};

    \path [-latex](amr-weigh) edge node[amr-edge] {ARG3} (amr-mass-quantity);
    \path [-latex](amr-mass-quantity) edge node[amr-edge] {mod} (amr-number);
    \path [-latex](amr-mass-quantity) edge node[amr-edge] {poss} (amr-kilogram);
       
\end{tikzpicture}
{\Large $\rightarrow$}
\begin{tikzpicture}[baseline=(current bounding box.center)]
\tikzstyle{amr-node}=[shape=ellipse,draw, inner sep=0.2, minimum height=0.8cm, text height=1.5ex, text depth=.25ex, minimum width=1cm]
\tikzstyle{amr-edge}=[text height=1.5ex, text depth=.25ex, fill=white]
\tikzstyle{text-node}=[text height=1.5ex, text depth=.25ex, align=center]

	\node (PAD2) at (-2.5,0){};
	\node (PAD2) at (2.5,0){};
	
    \node[amr-node] (amr-weigh) at (0,0) {weigh-01};
    \node[amr-node, dotted] (amr-mass-quantity) at (0,-2) {mass-quantity};
    \node[amr-node] (amr-number) at (-1.5,-4) {$1.1$};
    \node[amr-node] (amr-kilogram) at (1.5,-4) {kilogram};

    \path [-latex](amr-weigh) edge node[amr-edge] {ARG3} (amr-mass-quantity);
    \path [-latex](amr-mass-quantity) edge node[amr-edge] {mod} (amr-number);
    \path [-latex](amr-mass-quantity) edge node[amr-edge] {poss} (amr-kilogram);
       
\end{tikzpicture}
}
\caption{\textsc{Delete} transition applied to the node with label ``mass-quantity''; deletion is indicated by a dotted border. The reference realization of this partial AMR graph is ``weighs 1.1 kilogram''.}
\label{fig:delete}
\end{figure}
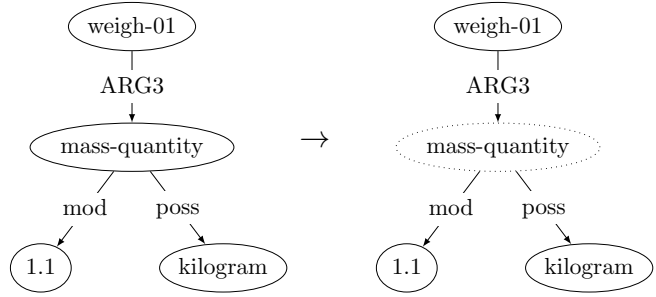

\noindent\begin{minipage}{\textwidth}
\renewcommand{\arraystretch}{\tbarraystretch}
\begin{itemize}
\item \textsc{Keep}

\begin{mdframed}
\begin{tabularx}{\linewidth}{@{}l X@{}}
Mapping: & $(G, \sigma_1{:}\sigma, \varepsilon, \rho) \mapsto (G, \sigma_1{:}\sigma, \varepsilon, \rho[\anno{DEL}(\sigma_1) \mapsto 0])$ \\[\tbcolumnpad]
Domain: & $\{(G, \sigma_1{:}\sigma, \varepsilon, \rho) \in C_\text{AMR} \mid |\text{in}(\sigma_1)| = 1 \wedge \sigma_1 \notin \text{dom}(\rho(\anno{DEL})) \}$ \\
\end{tabularx}
\end{mdframed}

This transition serves as a counterpart to \textsc{Delete} as its application indicates that the realization of node $\sigma_1$ is a part of the generated sentence. The \textsc{Keep} transition also fixes the position of $\sigma_1$ with respect to its predecessors, i.e. no more \textsc{Merge} or \textsc{Swap} transitions can be applied to it afterwards. 

While \textsc{Keep} is not an absolutely necessary transition for our transition system to work, including it allows us to make the generation process more efficient (see Section~\ref{GENERATION:Decoding}).
\end{itemize}
\end{minipage}

\noindent\begin{minipage}{\textwidth}
\renewcommand{\arraystretch}{\tbarraystretch}
\begin{itemize}
\item 
\textsc{Realize}-$(w,\alpha)$ ($w \in \Sigma_\text{E}^*, \alpha \in \mathcal{A}_\text{syn}$)

\begin{mdframed}
\begin{tabularx}{\linewidth}{@{}l X@{}}
Mapping: & $(G, \sigma_1{:}\sigma, \varepsilon, \rho) \mapsto (G, \sigma_1{:}\sigma, \varepsilon, \rho'[\anno{REAL}(\sigma_1) \mapsto w])$ where $\rho'$ is obtained from $\rho$ by setting $\rho'(k)(\sigma_1) = \alpha(k)$ for all $k \in \mathcal{K}_\text{syn}$. \\[\tbcolumnpad]
Domain: & $\{(G, \sigma_1{:}\sigma, \varepsilon, \rho) \in C_\text{AMR} \mid \rho(\anno{DEL})(\sigma_1) = 0 \wedge \sigma_1 \notin \text{dom}(\rho(\anno{REAL})) \wedge ( \sigma_1 \notin \text{dom}(\rho(\anno{POS})) \vee \rho(\anno{POS})(\sigma_1) = \alpha(\anno{POS}) ) \}$ \\
\end{tabularx}
\end{mdframed}

\textsc{Realize}-$(w,\alpha)$ specifies both the syntactic annotation and the realization of node $\sigma_1$, i.e. a consecutive sequence of words $w$ by which $\sigma_1$ is represented in the generated sentence. To give an example, reasonable transitions for a node labeled ``possible'' include \textsc{Realize}-$(\text{can},\alpha_1)$, \textsc{Realize}-$(\text{could},\alpha_1)$, \textsc{Realize}-$(\text{possible},\alpha_2)$ and \textsc{Realize}-$(\text{possibility},\alpha_3)$ where 
\begin{align*}
\alpha_1 & = \{(k, \text{--}) \mid k \in \mathcal{K}_\text{syn}\}[\anno{POS} \mapsto \text{MD}]
\qquad
\alpha_2 = \{(k, \text{--}) \mid k \in \mathcal{K}_\text{syn}\}[\anno{POS} \mapsto \text{JJ}] \\
\alpha_3 & = \{ (\anno{POS}, \text{NN}), (\anno{DENOM}, \text{a}), (\anno{TENSE}, \text{--}), (\anno{NUMBER}, \text{singular}), (\anno{VOICE}, \text{--}) \}\,.
\end{align*}
\end{itemize}
\end{minipage}

\begin{figure}[]
\centering
\scalebox{0.8} {
\begin{tikzpicture}[baseline=(current bounding box.center)]
\tikzstyle{amr-node}=[shape=ellipse,draw, inner sep=0.2, minimum height=0.8cm, text height=1.5ex, minimum width=1cm, text depth=.25ex]
\tikzstyle{amr-edge}=[text height=1.5ex, text depth=.25ex, fill=white]
\tikzstyle{text-node}=[text height=1.5ex, text depth=.25ex, align=center]

	\node (PAD2) at (-2.5,0){};
	\node (PAD2) at (2.5,0){};

    \node[amr-node] (amr-follow) at (0,0) {follow-02};
    \node[amr-node] (amr-i) at (-1.8,-1.7) {I};
    \node[amr-node] (amr-polarity) at (1.8,-1.7) {$-$};

    \path [-latex](amr-follow) edge node[amr-edge] {ARG0} (amr-i);
    \path [-latex](amr-follow) edge node[amr-edge] {polarity} (amr-polarity);
       
\end{tikzpicture}
{\Large $\rightarrow$}
\begin{tikzpicture}[baseline=(current bounding box.center)]
\tikzstyle{amr-node}=[shape=ellipse,draw, inner sep=0.2, minimum height=0.8cm, text height=1.5ex, text depth=.25ex, minimum width=1cm]
\tikzstyle{amr-edge}=[text height=1.5ex, text depth=.25ex, fill=white]
\tikzstyle{text-node}=[text height=1.5ex, text depth=.25ex, align=center]

	\node (PAD2) at (-2.5,0){};
	\node (PAD2) at (2.5,0){};

    \node[amr-node] (amr-follow) at (0,0) {follow-02};
    \node[amr-node] (amr-do) at (0, -1.7) {do};
    \node[amr-node] (amr-i) at (-2.2,-1.7) {I};
    \node[amr-node] (amr-polarity) at (2.2,-1.7) {$-$};

    \path [-latex](amr-follow) edge node[amr-edge] {ARG0} (amr-i);
    \path [-latex](amr-follow) edge node[amr-edge] {polarity} (amr-polarity);
    \path [-latex](amr-follow) edge node[fill=white] {$\star$} (amr-do);
       
\end{tikzpicture}
}
\caption{\textsc{Insert-Child}-(do,$\mathsf{left}$) transition applied to the node with label ``follow-02''. The reference realization of this partial AMR graph is ``I do not follow''.}
\label{fig:insert-child}
\end{figure}
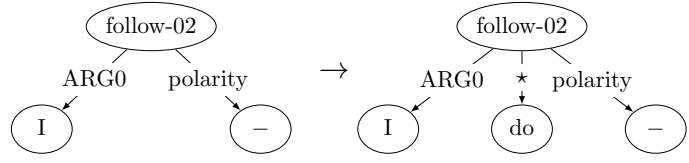

\renewcommand{\arraystretch}{\tbarraystretch}
\begin{itemize}
\item 
\textsc{Insert-Child}-$(w,p)$ ($w \in \Sigma_\text{E}$, $p \in \{ \mathsf{left, right} \}$)

\begin{mdframed}
\begin{tabularx}{\linewidth}{@{}l X@{}}
Mapping: & $(G, \sigma_1{:}\sigma, \varepsilon, \rho) \mapsto (G', \tilde{\sigma}{:}\sigma_1{:}\sigma, \varepsilon, \rho[\anno{DEL}(\tilde{\sigma})\mapsto 0, \anno{INS-DONE}(\tilde{\sigma})\mapsto 1])$ where $\tilde{\sigma} \in V_\text{ins} \setminus V$ is some new node and
{\setlength{\belowdisplayskip}{0pt}
\setlength{\belowdisplayshortskip}{0pt} 
\begin{align*}
G' & = (V \cup \{ \tilde{\sigma} \}, E \cup \{ (\sigma_1, \star, \tilde{\sigma}) \}, L \cup \{ (\tilde{\sigma}, w) \}, \prec' ) \\[\tblinepad]
{\prec'} & = {
	\begin{cases} 
	{\prec} \cup \{ (\tilde{\sigma}, \sigma_1) \} & \text{ if } p = \mathsf{left} \,, \\ 
	{\prec} \cup \{ (\sigma_1, \tilde{\sigma}) \} & \text { if } p = \mathsf{right} \,. 
	\end{cases}
}
\end{align*}} \\[\tbcolumnpad]
Domain: & $\{(G, \sigma_1{:}\sigma, \varepsilon, \rho) \in C_\text{AMR} \mid \rho(\anno{DEL})(\sigma_1) = 0 \wedge \sigma_1 \in \text{dom}(\rho(\anno{REAL})) \wedge \sigma_1 \notin \text{dom}(\rho(\anno{INS-DONE})) \cup  \text{dom}(\rho(\anno{LINK})) \}$ \\
\end{tabularx}
\end{mdframed}

This transition inserts a new node $\tilde{\sigma}$ with label $w$ as a child of $\sigma_1$; it also specifies whether the realization of the new node is to be left or right of $\sigma_1$ in the generated sentence. A placeholder label $\star$ is assigned to the edge connecting $\sigma_1$ and $\tilde{\sigma}$; the latter is put on top of the node buffer. To assure that the inserted node can not have children on its own, $\rho(\anno{INS-DONE})(\tilde{\sigma})$ is set to $1$. 

Commonly inserted child nodes include prepositions, articles and auxiliary verbs; an exemplary application of \textsc{Insert-Child}-(do,$\mathsf{left}$) is shown in Figure~\ref{fig:insert-child}.
\end{itemize}

\noindent\begin{minipage}{\textwidth}
\renewcommand{\arraystretch}{\tbarraystretch}
\begin{itemize}
\item 
\textsc{Reorder}-$(v_1, \ldots, v_n)$ ($v_i \in V$, $i \in [n]$, $n \in \mathbb{N}$)

\begin{mdframed}
\begin{tabularx}{\linewidth}{@{}l X@{}}
Mapping: & $(G, \sigma_1{:}\sigma, \varepsilon, \rho) \mapsto (G', \sigma', (v_1, \ldots, v_n)\setminus \{ \sigma_1 \}, \rho)$ where
{
\setlength{\belowdisplayskip}{0pt}
\setlength{\belowdisplayshortskip}{0pt} 
\begin{align*}
G' &= (V,E,L,\prec') \\[\tblinepad]
{\prec'} & = ({\prec} \cup \{ (v_i, v_{i+1}) \mid i \in [n-1] \})^+ \\[\tblinepad]
\sigma' & = 
\begin{cases}
\sigma_1{:}\sigma & \text{if } n \geq 2 \\
\sigma & \text{otherwise.}
\end{cases}
\end{align*}
}
\\[\tbcolumnpad]
Domain: &
 $ \begin{aligned}[t]
        & \{(G, \sigma_1{:}\sigma, \varepsilon, \rho) \in C_\text{AMR} \mid \{\sigma_1\} \cup \ch[G]{\sigma_1} = \{v_1, \ldots, v_n \} \\ & \wedge (\sigma_1 \in \text{dom}(\rho(\anno{INS-DONE})) \cap \text{dom}(\rho(\anno{REAL})) \vee \rho(\anno{DEL})(\sigma_1) = 1 ) \\
        & \wedge {({\prec} \cup \{ (v_i, v_{i+1}) \mid i \in [n-1] \})^+} \text{ is a strict order} \}
         \end{aligned}$ \\
\end{tabularx}
\end{mdframed}

With this transition, the order among $\ch[G]{\sigma_1} \cup \{\sigma_1\}$ in the realization of $G$ is specified. After the application of \textsc{Reorder}, the $\sigma_1$-subgraph $\restr{G}{\sigma_1}$ is guaranteed to be a totally ordered graph because $G$ is processed bottom-up, i.e. for each node $v \in \text{succ}(\sigma_1)$, some instance of \textsc{Reorder} has already been applied.
\end{itemize}
\end{minipage}

\begin{itemize}

\item {
\textsc{Insert-Between}-$(w,p)$ ($w \in \Sigma_\text{E}$, $p \in \{\mathsf{left, right} \}$)

\begin{mdframed}
\begin{tabularx}{\linewidth}{@{}l X@{}}
Mapping: & $(G, \sigma_1{:}\sigma, \beta_1{:}\beta, \rho) \mapsto (G', \sigma', \beta, \rho[\anno{REAL}(\tilde{\sigma}) \mapsto w])$  where $\tilde{\sigma} \in V_\text{ins} \setminus V$ is some new node, $l_{\beta_1}$ denotes the label of the edge connecting $\sigma_1$ with $\beta_1$ and
{\setlength{\belowdisplayskip}{0pt}
\setlength{\belowdisplayshortskip}{0pt} 
\begin{align*}
G' & = (V \cup \{ \tilde{\sigma} \}, E', L \cup \{ (\tilde{\sigma}, w) \}, \prec' ) \\[\tblinepad]
E' & =  E \setminus \{ (\sigma_1, l_{\beta_1}, \beta_1) \} \cup \{ (\sigma_1, l_{\beta_1}, \tilde{\sigma}), (\tilde{\sigma}, \star, \beta_1) \} \\[\tblinepad]
{\prec'} & = ( {\prec} \cup {\prec''} \cup \{ (v, \tilde{\sigma}) \mid (v, \beta_1) \in {\prec} \} \cup \{ (\tilde{\sigma},v) \mid (\beta_1,v) \in {\prec} \} )^+ \\[\tblinepad]
{\prec''} & = {
	\begin{cases} 
	{\prec} \cup \{ (\tilde{\sigma}, \beta_1) \} & \text{ if } p = \mathsf{left} \\ 
	{\prec} \cup \{ (\beta_1, \tilde{\sigma}) \} & \text { if } p = \mathsf{right}
	\end{cases}
}
\qquad
 \sigma' = 
 \begin{cases}
 \sigma_1{:}\sigma & \text{if } \beta \neq \varepsilon \\
 \sigma & \text{otherwise.}
 \end{cases}
\end{align*}} \\[\tbcolumnpad]
Domain: & $\{(G, \sigma_1{:}\sigma, \beta_1{:}\beta, \rho) \in C_\text{AMR} \mid \rho(\anno{DEL})(\sigma_1) = 0 \}$ \\
\end{tabularx}
\end{mdframed}

This transition inserts a new node $\tilde{\sigma}$ with label $w$ and realization $w$ between $\sigma_1$, the top element of the node buffer, and $\beta_1$, the top element of the child buffer; it also specifies whether the realization of $\tilde{\sigma}$ should be left or right of $\beta_1$ in the generated sentence. As \textsc{Insert-Between}-$(w,p)$ specifies both the realization and the position of the inserted node, the latter is already completely processed right after its insertion and therefore does not need to be put onto the node buffer. The placeholder edge label $\star$ is assigned to the new edge connecting $\tilde{\sigma}$ and $\beta_1$.

\textsc{Insert-Between} transitions are mostly used to insert adpositions (e.g. ``of'', ``to'', ``in'', ``for'', ``on'') between two nodes; an example can be seen in Figure~\ref{fig:insert-between}.

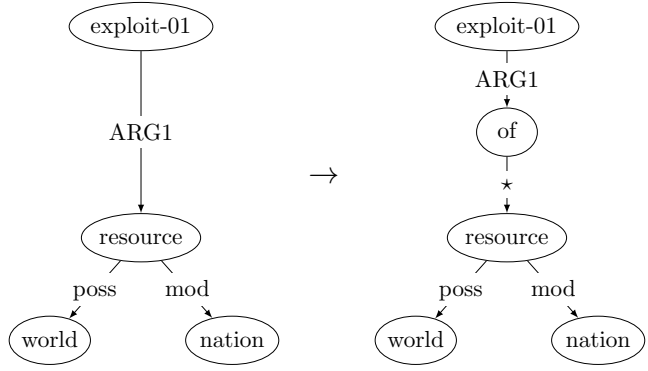
\begin{figure}[]
\centering
\scalebox{0.8} {
\begin{tikzpicture}[baseline=(current bounding box.center)]
\tikzstyle{amr-node}=[shape=ellipse,draw, inner sep=0.2, minimum height=0.8cm, text height=1.5ex, text depth=.25ex, minimum width=1cm]
\tikzstyle{amr-edge}=[text height=1.5ex, text depth=.25ex, fill=white]
\tikzstyle{text-node}=[text height=1.5ex, text depth=.25ex, align=center]

	\node (PAD2) at (-2.5,0){};
	\node (PAD2) at (2.5,0){};
	
    \node[amr-node] (amr-exploit) at (0,0) {exploit-01};
    \node[amr-node] (amr-resource) at (0,-3.5) {resource};
    \node[amr-node] (amr-nation) at (1.5,-5.2) {nation};
    \node[amr-node] (amr-world) at (-1.5,-5.2) {world};

    \path [-latex](amr-exploit) edge node[fill=white] {ARG1} (amr-resource);
    \path [-latex](amr-resource) edge node[amr-edge, pos=0.51] {mod} (amr-nation);
    \path [-latex](amr-resource) edge node[amr-edge] {poss} (amr-world);
       
\end{tikzpicture}
{\Large $\rightarrow$}
\begin{tikzpicture}[baseline=(current bounding box.center)]
\tikzstyle{amr-node}=[shape=ellipse,draw, inner sep=0.2, minimum height=0.8cm, text height=1.5ex, text depth=.25ex, minimum width=1cm]
\tikzstyle{amr-edge}=[text height=1.5ex, text depth=.25ex, fill=white]
\tikzstyle{text-node}=[text height=1.5ex, text depth=.25ex, align=center]

	\node (PAD2) at (-2.5,0){};
	\node (PAD2) at (2.5,0){};

    \node[amr-node] (amr-exploit) at (0,0) {exploit-01};
    \node[amr-node] (amr-of) at (0,-1.75) {of};
    \node[amr-node] (amr-resource) at (0,-3.5) {resource};
    \node[amr-node] (amr-nation) at (1.5,-5.2) {nation};
    \node[amr-node] (amr-world) at (-1.5,-5.2) {world};

    \path [-latex](amr-exploit) edge node[fill=white] {ARG1} (amr-of);
    \path [-latex](amr-of) edge node[fill=white] {$\star$} (amr-resource);
    \path [-latex](amr-resource) edge node[amr-edge, pos=0.51] {mod} (amr-nation);
    \path [-latex](amr-resource) edge node[amr-edge] {poss} (amr-world);
       
\end{tikzpicture}
}
\caption{\textsc{Insert-Between}-(of,$\textsf{left}$) transition applied to the nodes with labels ``exploit-01'' and ``resource''. The reference realization of this partial AMR graph is ``the exploitation of the world's national resources''.}
\label{fig:insert-between}
\end{figure}

}

\item {
\textsc{No-Insertion}

\begin{mdframed}
\begin{tabularx}{\linewidth}{@{}l X@{}}
Mapping: & $(G, \sigma_1{:}\sigma, \varepsilon, \rho) \mapsto (G, \sigma_1{:}\sigma, \varepsilon, \rho[\anno{INS-DONE}(\sigma_1) \mapsto 1])$ \\[\tblinepad]
 & $(G, \sigma_1{:}\sigma, \beta_1{:}\beta, \rho) \mapsto (G, \sigma', \beta, \rho)$ where 
 \setlength{\belowdisplayskip}{0pt}
 \setlength{\belowdisplayshortskip}{0pt} 
 \[
 \sigma' = 
 \begin{cases}
 \sigma_1{:}\sigma & \text{if } \beta \neq \varepsilon \\
 \sigma & \text{otherwise.}
 \end{cases}
 \]\\[\tbcolumnpad]
Domain: & $\{(G, \sigma_1{:}\sigma, \varepsilon, \rho) \in C_\text{AMR} \mid \rho(\anno{DEL})(\sigma_1) = 0 \wedge \sigma_1 \in \text{dom}(\rho(\anno{REAL})) \wedge \sigma_1 \notin \text{dom}(\rho(\anno{INS-DONE}))  \} \cup \{ (G, \sigma_1{:}\sigma, \beta_1{:}\beta, \rho) \in C_\text{AMR} \}$ \\
\end{tabularx}
\end{mdframed}

\textsc{No-Insertion} serves as counterpart to both \textsc{Insert-Between} and \textsc{Insert-Child} and indicates that no node needs to be inserted. In case the edge buffer is not empty, this transition removes the top element $\beta_1$; otherwise, it leaves the graph and both buffers unchanged, but sets the \anno{INS-DONE} flag of $\sigma_1$ to $1$.
}

\end{itemize}
\egroup

This concludes our discussion of $T_\text{AMR}$. For each transition $t \in T_\text{AMR}$, we denote by $\mathcal{C}(t)$ the \emph{class} to which it belongs; this class is obtained by simply removing all parameters from $t$.\label{los:c} To give a few examples, $\mathcal{C}(\textsc{Insert-Between-}(\text{of},\textsf{left})) = \textsc{Insert-Between}$ and $\mathcal{C}(\textsc{Merge-}(\text{earlier,JJ})) = \textsc{Merge}$. We extend this definition to subsets $T$ of $T_\text{AMR}$ and denote by $\mathcal{C}(T)$ the set $\{ \mathcal{C}(t) \mid t \in T \}$; in particular, $\mathcal{C}(T_\text{AMR})$ denotes the set of all classes of transitions used in our transition system $S_\text{AMR}$.

\subsubsection{Modeling}
\label{GENERATION:Modeling}

We now turn the transition system $S_\text{AMR}$ into an actual generator; in other words, we derive from it a function $g \colon \mathcal{G}_\text{AMR} \rightarrow \Sigma_\text{E}^*$ that assigns to each AMR graph $G$ some realization $\hat{w} = g(G)$. Given an AMR graph $G$ as input, our key idea is to rank all possible transition sequences according to some score. We then take the sentence generated by the highest scoring transition sequence to be the output of our generator:
\begin{equation}
\hat{w} = \text{out}(\hat{t}, G)\ \text{ where } \hat{t} = \argmax_{t \in \mathcal{T}({S_\text{AMR}},G)}{\text{score}(t, G)}\,. \label{equ:generator}
\end{equation}
We define the score of a transition sequence $t = (t_1, \ldots, t_n) \in \mathcal{T}({S_\text{AMR}},G)$ to be a linear combination of a score assigned to its output by a language model, denoted by $\text{score}_\text{LM}$, and a score assigned to the individual transitions $t_i$, $i \in [n]$, denoted by $\text{score}_\text{TS}$:
\begin{equation}
\text{score}(t, G) = \theta_\text{LM} \cdot \text{score}_\text{LM}( \text{out}(t, G) ) + \sum_{i=1}^n \theta_{\mathcal{C}(t_i)} \cdot \text{score}_\text{TS} (t_i, t, G)\,. \label{equ:score} 
\end{equation}
In the above equation, $\theta_\text{LM} \in \mathbb{R}^+$ and $\theta_\tau \in \mathbb{R}^+$, $\tau \in \mathcal{C}(T_\text{AMR})$ are hyperparameters; how they are obtained is described in Section~\ref{GENERATION:HyperparameterOptimization}. We may theoretically define $\text{score}_\text{LM}$ using an arbitrary language model $p_\text{LM}$ (see Definition~\ref{def:language-model}) but we explicitly assume here an $n$-gram model and set 
\begin{equation}
\text{score}_\text{LM}(w) = \log{ p_\text{LM}(w)} \cdot |w|^{-1} \label{equ:score_lm}
\end{equation}
where the additional factor of $|w|^{-1}$ is used to compensate for the fact that $n$-gram language models tend to favor sentences with only few words. We finally set
\begin{equation}
\text{score}_\text{TS}(t_i, t, G) = \log P(t_i \mid t_1, \ldots, t_{i-1}, G ) \label{equ:score_ts}
\end{equation}
where $P(t_i \mid t_1, \ldots, t_{i-1}, G )$ denotes the probability of $t_i$ being the correct transition to be applied next when the input to the transition system is $G$ and the previously applied transitions are $t_1$ to $t_{i-1}$. We assume that this probability depends only on the current configuration and not on all previously applied transitions, allowing us to simplify
\begin{equation}
P(t_i \mid t_1, \ldots, t_{i-1}, G ) = P(t_i \mid c) 
\end{equation}
where $c = (t_1, \ldots, t_{i-1})(G)$ denotes the configuration obtained from applying $t_1, \ldots, t_{i-1}$ to $\cs(G)$ (see Definition~\ref{def:transition-system}).
If $t_i$ does not belong to one of the classes \textsc{Realize} and \textsc{Reorder}, we simply estimate the above conditional probabilities $P(t_i \mid c )$ using a maximum entropy model, i.e. we assume  
\begin{equation}
P(t_i \mid c) = p_\text{TS}(t_i \mid c) \label{equ:p_ts}
\end{equation}
where $p_\text{TS}$ is a maximum entropy model for $T_\text{AMR}$ and $C_\text{AMR}$; the features used by $p_\text{TS}$ will be described in Section~\ref{GENERATION:Training} where we will also discuss the training procedure.

We now consider the two special cases of \textsc{Realize} and \textsc{Reorder} transitions. For this purpose, let $c = (G, \sigma_1{:}\sigma, \beta, \rho) \in C_\text{AMR}$ be a configuration for AMR generation where ${G = (V,E,L,\prec)}$. Furthermore, let $w \in \Sigma_\text{E}^*$ and $\alpha \in \mathcal{A}_\text{syn}$. Using the law of total probabilities, we derive
\begin{equation}
P(\textsc{Realize-}(w,\alpha) \mid c) = \sum_{\alpha' \in \mathcal{A}_\text{syn}} P(\alpha', \textsc{Realize-}(w,\alpha) \mid c) \label{equ:rea1}
\end{equation}
where $P(\alpha', t \mid c)$ denotes the joint probability of $\alpha'$ being the right annotation for $\sigma_1$ and $t$ being the correct transition to be applied next given $c$. As this transition must assign the right syntactic annotation to $\sigma_1$, we argue that $P(\alpha', \textsc{Realize-}(w,\alpha) \mid c) = 0$ for all $\alpha' \neq \alpha$, allowing us to simplify Eq.~(\ref{equ:rea1}) to
\begin{align}
P(\textsc{Realize-}(w,\alpha) \mid c) & = P(\alpha, \textsc{Realize-}(w,\alpha) \mid c) \label{equ:rea2} \\ 
& = P(\alpha \mid c) \cdot P(\textsc{Realize-}(w,\alpha) \mid c, \alpha) \label{equ:rea3} 
\end{align}
where Eq.~(\ref{equ:rea3}) is obtained from Eq.~(\ref{equ:rea2}) using the general product rule.

We make the simplifying assumption that $P(\alpha \mid c)$ depends only on $G$ and $\sigma_1$, but we replace $P(\alpha \mid G, \sigma_1)$ with its weighted version $P^\text{w}(\alpha \mid G, \sigma_1)$ as introduced in Section~\ref{GENERATION:SyntacticAnnotation}. Furthermore, we use a maximum entropy model $p_\textsc{Real}$ for $T_\text{AMR}$ and $C_\text{AMR} \times \mathcal{A}_\text{syn}$ to estimate $P(t \mid c, \alpha)$ and obtain

\begin{equation}
P(\textsc{Realize-}(w,\alpha) \mid c) = P^\text{w}(\alpha \mid G, \sigma_1) \cdot p_\textsc{Real}(\textsc{Realize-}(w,\alpha) \mid c, \alpha)\,. \label{equ:rea4}
\end{equation}

For \textsc{Reorder} transitions, we use an approach similar to the one of \citet{pourdamghani2016generating}. Let $c$ and $G$ be defined as above. Furthermore, let $s = (v_1, \ldots, v_n)$, $n \in \mathbb{N}$ be a sequence of vertices from $V$ such that $c \in \text{dom}(\textsc{Reorder-}(v_1, \ldots, v_n))$. Then there is some $k \in [n]$ such that $s = (v_1, \ldots, v_{k-1}, \sigma_1, v_{k+1}, \ldots, v_n)$. Let
\[
{\lessdot} = \{ (v_i, v_j) \mid 1 \leq i < j \leq n \}
\]
denote the total order such that $s$ is the $(\ch{\sigma_1} \cup \{ \sigma_1 \})$-sequence induced by $\lessdot$. As applying $\textsc{Reorder-}(v_1, \ldots, v_n)$ has the effect of adding $\lessdot$ to $\prec$, we rewrite
\begin{equation}
P(\textsc{Reorder-}(v_1,\ldots,v_n) \mid c) = P( {\lessdot} \mid c) 
\end{equation}
where $P({\lessdot} \mid c)$ denotes the probability of $\lessdot$ being the correct order among $\ch{\sigma_1} \cup \{ \sigma_1 \}$ given $c$. We extract from $\lessdot$ three disjoint sets
\begin{align*}
{\lessdot_*} & = \{ (v_1, v_2) \in {\lessdot} \mid v_1 = \sigma_1 \vee v_2 = \sigma_1 \} \\
{\lessdot_l} & = \{ (v_i, v_j) \in {\lessdot} \mid 1 \leq i < j \leq k-1 \} \\
{\lessdot_r} & = \{ (v_i, v_j) \in {\lessdot} \mid k +1 \leq i < j \leq n \}
\end{align*}
such that $\lessdot_{*}$ contains all tuples from $\lessdot$ involving $\sigma_1$, $\lessdot_l$ contains all tuples for which both vertices are left of $\sigma_1$ and $\lessdot_r$ contains all tuples for which both vertices are right of $\sigma_1$. We note that ${\lessdot} = ( {\lessdot_*} \cup {\lessdot_r} \cup {\lessdot_l} )^+$ and assume
\begin{equation}
P( {\lessdot} \mid c) = P({\lessdot_*} , {\lessdot_r} , {\lessdot_l} \mid c) \,.
\end{equation}
Under the further assumption that the order among the vertices left of $\sigma_1$ is independent of the order among those right of $\sigma_1$, we can use the general product rule to obtain
\begin{equation}
P( {\lessdot} \mid c) = P({\lessdot_*} \mid c) \cdot P({\lessdot_r} \mid c, {\lessdot_*}) \cdot P({\lessdot_l} \mid c, {\lessdot_*}) \,. \label{equ:reo1}
\end{equation}
We finally assume that firstly, the elements contained within $\lessdot_*$ are conditionally independent of one another given $c$ and that secondly, for all $1 \leq i < j \leq n$ with $k \notin \{i, j\}$, the probability of $v_i$ occurring before $v_j$ depends only on $c$ and the relative position of both $v_1$ and $v_2$ with respect to $\sigma_1$. This allows us to transform Eq.~(\ref{equ:reo1}) into
\begin{equation}
\begin{split}
P({\lessdot} \mid c) & = \prod_{i=1}^{k-1} P(v_i \lessdot \sigma_1 \mid c) \cdot \prod_{i=k+1}^n  P(\sigma_1 \lessdot v_i \mid c)\\ 
 & \cdot \prod_{\substack{i=1\\\hphantom{i=k+2}}}^{k-2} \prod_{j=i+1}^{k-1} P(v_i \lessdot v_j \mid c, v_i \lessdot \sigma_1, v_j \lessdot \sigma_1) \\ 
 & \cdot \prod_{i=k+1}^{n-1} \prod_{j = i+1}^n P(v_i \lessdot v_j \mid c, \sigma_1 \lessdot v_i, \sigma_1 \lessdot v_j)\,.
\end{split} \label{equ:reo2}
\end{equation}
We note that as $\lessdot$ is a total order, for all $v, v' \in \ch{\sigma_1} \cup \{ \sigma_1 \}$ we must either have $v \lessdot v'$ or $v' \lessdot v$. We can thus rewrite 
\[P(v \lessdot v' \mid c) = 1 - P(v' \lessdot v \mid c)\,.\]
Using this identity, slightly reordering the terms from Eq.~(\ref{equ:reo2}) and estimating all required probabilities through maximum entropy models $p_*$, $p_l$ and $p_r$, respectively, we arrive at our final equation 
\begin{equation}
\begin{split}
P&(\textsc{Reorder-}(v_1,\ldots,v_n) \mid c) \\
&= \prod_{i=1}^{k-1} \left( p_*(v_i \lessdot \sigma_1 \mid c) \cdot  \prod_{j=i+1}^{k-1} p_l(v_i \lessdot v_j \mid c, v_i \lessdot \sigma_1, v_j \lessdot \sigma_1)\right) \\
&\hphantom{=} \cdot \prod_{i=k+1}^n \left( ( 1 - p_*(v_i \lessdot \sigma_1 \mid c) ) \cdot \prod_{j = i+1}^n p_r(v_i \lessdot v_j \mid  c, \sigma_1 \lessdot v_i, \sigma_1 \lessdot v_j) \right)\,.
\end{split} \label{equ:reo3}
\end{equation}
Like for the other classes of transitions, the details of training the maximum entropy models from the above equation are described in Section~\ref{GENERATION:Training}.

\subsubsection{Decoding}
\label{GENERATION:Decoding}

Unfortunately, finding the solution to Eq.~(\ref{equ:generator}) by simply trying all possible transition sequences $t \in \mathcal{T}(S_\text{AMR}, G)$ is far from being feasible for large AMR graphs $G$. Therefore, the aim of this section is to find a good approximation $\tilde{w}$ of $g(G)$ that can efficiently be computed. We then use this approximation $\tilde{w}$ as the output of our generator.

An obvious first approach to approximate $g(G)$ would be to start with the initial configuration $\cs(G)$ and then continuously apply the most likely transition until a terminal configuration $c_t \in \ct$ is reached. This idea is implemented in Algorithm~\ref{alg:greedy-generation}, which is the equivalent of the parsing algorithm used by \citet{wang2015transition}; we will refer to it as the \emph{greedy generation algorithm} and denote the obtained terminal configuration $c_t$ by $\text{generateGreedy}(G)$.

\begin{algorithm}
\DontPrintSemicolon
\KwIn{AMR graph $G = (V, E, L , \prec)$}
\KwOut{terminal configuration $c \in \ct$}
\SetKwBlock{Begin}{function}{end function}
\Begin($\text{generateGreedy} {(} G {)}$)
{

	$c \gets \cs (G)$\;
    \While{$c \notin \ct$}
    {
        $T^* \gets \{ t \in T_\text{AMR} \mid c \in \text{dom}(t) \}$\;
        $t^* \gets \argmax_{t \in T^*} P(t \mid c)$\;
        $c \gets t^*(c)$\;
    }
    \Return{$c$}
}
\caption{Greedy generation algorithm}\label{alg:greedy-generation}
\end{algorithm}

While this first algorithm is both extremely simple and efficient, it suffers from the obvious problem that it does not in any way integrate the language model into the generation process and thus approximates the best solution to Eq.~(\ref{equ:generator}) rather poorly. A simple fix for this problem might be to consider for each configuration not just one, but the $n$-best applicable transitions $t_1, \ldots, t_n$, $n \in \mathbb{N}$ and to rerank all so-obtained transition sequences using the language model. However, even for small values of $n$ this approach is unfeasible as for $n > 1$, the number of transition sequences to consider grows exponentially with the number of vertices.
 
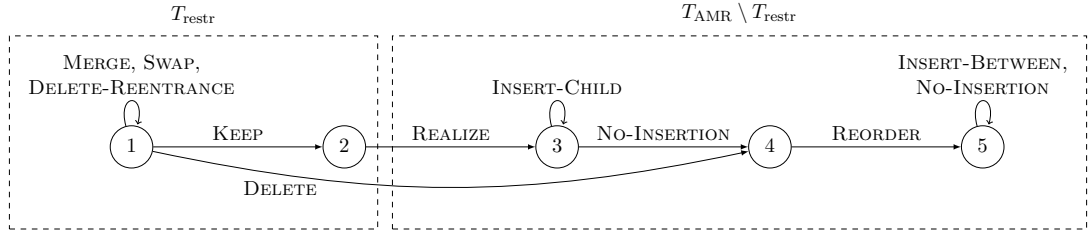
\begin{figure}[t]
\centering
\scalebox{0.7} {
\begin{tikzpicture}
\tikzstyle{p-node}=[shape=ellipse,draw, inner sep=0.2, minimum size=0.8cm, text height=1.5ex, text depth=.25ex]

    \node[p-node] (1) at (0,0) { 1 };
    \node[p-node] (2) at (4,0) { 2 };
    \node[p-node] (3) at (8,0) { 3 };
    \node[p-node] (4) at (12,0) { 4 };
    \node[p-node] (5) at (16,0) { 5 };

    \path [-latex](1) edge node[above] {\textsc{Keep}} (2);
    \path [-latex](1) edge[ bend angle=12, bend right] node[below, pos=0.2] (d) {\textsc{Delete}} (4);
    \path [-latex](1) edge[loop above] node[above, align=center] (msd) {\textsc{Merge}, \textsc{Swap},\\ \textsc{Delete-Reentrance}} (1);
    \path [-latex](2) edge node[above] (r) {\textsc{Realize}} (3);
    \path [-latex](3) edge[loop above] node[above, align=center] {\textsc{Insert-Child}} (3);
    \path [-latex](3) edge node[above, align=center] {\textsc{No-Insertion}} (4);
    \path [-latex](4) edge node[above, align=center] {\textsc{Reorder}} (5);
    \path [-latex](5) edge[loop above] node[above, align=center] (in) {\textsc{Insert-Between}, \\ \textsc{No-Insertion}} (5);
    
    \node (pad1) at (1, -1.2) {};
    \node (pad2) at (15, -1.2) {};
        
    \node[draw, dashed, fit=(1)(2)(d)(msd)(pad1), inner sep=0.22cm] (firststage) {};
    \node[draw, dashed, fit=(3)(4)(5)(in)(pad2)(r), inner sep=0.22cm] (secondstage) {};
    
    \node[above=0.1cm of firststage, text centered]{$T_\text{restr}$};
    \node[above=0.1cm of secondstage, text centered]{$T_\text{AMR} \setminus T_\text{restr}$};

\end{tikzpicture}
}
\caption{Graphical representation of the order in which transitions can be applied to a node}
\label{fig:transition-sequences}
\end{figure}

Another approach would be to directly take the language model into account at each transition step. It is, however, not clear how a partial transition sequence or a single transition might be scored by our language model; even more so if said transition does not directly effect the realization of a node.
Our solution to this problem stems from an observation shown in Figure~\ref{fig:transition-sequences}: The transitions in $T_\text{AMR}$ are applied to each node $v$ of our input graph $G$ in a very specific order; this order can roughly be divided into five stages (numbered $1$ to $5$ in Figure~\ref{fig:transition-sequences}). First, $\textsc{Merge}$, $\textsc{Swap}$ and $\textsc{Delete-Reentrance}$ transitions modify the relation between $v$ and its predecessors~($1$). Afterwards, it is decided whether $v$ is deleted or kept; in the latter case, a realization must be determined and child nodes may be inserted ($2$, $3$). Irrespective of whether $v$ was deleted, an order among its children must be determined in the next stage ($4$) before finally, insertions between $v$ and its children are applied ($5$).

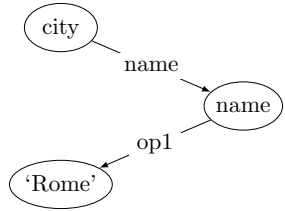
\begin{figure}[]
\centering
\scalebox{0.8} {
\begin{tikzpicture}
\tikzstyle{amr-node}=[shape=ellipse,draw, inner sep=0.2, minimum height=0.8cm, minimum width=1.2cm, text height=1.5ex, text depth=.25ex]
    \node[amr-node] (amr-city) at (0,0) {city};
    \node[amr-node] (amr-name) at (3,-1.3) {name};
    \node[amr-node] (amr-rome) at (0,-2.6) {`Rome'};
    \path [-latex](amr-city) edge node[fill=white] {name} (amr-name);
    \path [-latex](amr-name) edge node[fill=white] {op1} (amr-rome);
\end{tikzpicture}
}
\caption{AMR representation of Rome}
\label{fig:amr-graph-city}
\end{figure}

In accordance with these five stages, we partition the set $T_\text{AMR}$ into two disjoint sets of consecutive transitions (denoted by $T_\text{restr}$ and $T_\text{AMR} \setminus T_\text{restr}$, respectively). We choose this partition in such a way that the first set is restricted to transitions for which we believe that a language model is not helpful in rating them; the second one contains all remaining transitions. Each set can then be processed separately: In a first processing phase, we modify the input AMR graph using only transitions from $T_\text{restr}$ and completely ignoring the language model. In a second phase, we run a modified version of our generation algorithm on the output of the previous run, this time using only transitions from $T_\text{AMR} \setminus T_\text{restr}$, considering multiple possible transition sequences for each vertex and scoring them using the language model.  
As indicated in Figure~\ref{fig:transition-sequences}, we set \begin{align*}
T_\text{restr} = \{  t \in T_\text{AMR} \mid \mathcal{C}(t) \in \{ \textsc{Delete-Reentrance}, \textsc{Merge}, \textsc{Swap}, \textsc{Delete}, \textsc{Keep} \} \}\,.
\end{align*}\label{los:t_restr}
The reason for this specific choice is that all these transitions are applied to a node \emph{before} its realization is determined. Therefore, it often takes several subsequent transition steps until their effects on the generated sentence become clear; this makes it difficult to assign language model scores to them. While this is not entirely true for the \textsc{Delete} transition -- which does have a direct impact on the realizations of nodes -- a language model would still hardly be useful in rating it. For an example, consider the concepts ``city'' and ``name'' as used in Figure~\ref{fig:amr-graph-city}. Possible realizations of the corresponding AMR graph include ``the city with name Rome''  and simply ``Rome''. In most cases, we would prefer the latter realization over the first; thus, \textsc{Delete} transitions should be applied to the vertices labeled ``name'' and ``city''. However, as both ``city'' and ``name'' are frequent English words, it is likely that 
\[
\text{score}_\text{LM}(\text{the city with name Rome}) > \text{score}_\text{LM}(\text{Rome})
\]
and thus, the language model strongly favors applying \textsc{Keep} to both vertices. 

For the first phase of our generation algorithm -- in which only transitions from $T_\text{restr}$ are applied --, we slightly modify the definition of \textsc{Delete} and \textsc{Keep} transitions such that the top element $\sigma_1$ is removed from the node buffer whenever one of them is applied. We denote the result of applying this modified version of the greedy generation algorithm to some input graph $G$ by $\text{generateGreedy}_\text{restr}(G)$.

For the second phase of our two-phase approach, we must define how a partial transition sequence with transitions only from $T_\text{AMR} \setminus T_\text{restr}$ can be scored by a language model. As a starting point towards this goal, we first introduce the concept of \emph{partial transition functions}. 

\begin{definition}[Partial transition function]
Let $G = (V, E, L \prec)$ be a rooted acyclic graph. A \emph{partial transition function (for $G$)} is a function $b\colon V \cup V_\text{ins} \pfun (T_\text{AMR} \times [0,1])^*$ that assigns to some nodes $v \in V \cup V_\text{ins}$ a sequence of transitions to be applied when $v$ is the top element of the node buffer along with their probabilities. The set of all partial transition functions is denoted by $\mathcal{T}_\text{AMR}^\text{par}$. \label{los:t_amr_par}
\end{definition} 

Using this notion of a partial transition function $b$, we derive Algorithm~\ref{alg:partial-generation} that, given some configuration $c = (G, \varepsilon, \varepsilon, \rho) \in C_\text{AMR}$, applies to each node $v$ of $G$ exactly those transitions specified by $b$; we refer to this algorithm as the \emph{partial generation algorithm} and denote the result of its application by $\text{generatePartial}(c,b)$.

\begin{algorithm}
\DontPrintSemicolon
\KwIn{configuration $c = (G, \varepsilon, \varepsilon, \rho) \in C_\text{AMR}$ where $G = (V, E, L , \prec)$ is rooted and acyclic, partial transition function $b \in \mathcal{T}_\text{AMR}^\text{par}$}
\KwOut{configuration $c_r \in C_\text{AMR}$, the result of partially processing $c$ with $b$}
\SetKwBlock{Begin}{function}{end function}
\Begin($\text{generatePartial} {(} c, b {)}$)
{
		let $\sigma$ be a bottom-up traversal of all nodes in $G$\;
        $c \gets (G, \sigma, \varepsilon, \rho)$\;
        \While{$c \notin \ct$}
        {
        	let $c = (G', \sigma_1{:}\sigma', \beta, \rho')$\;
        	\uIf{$\sigma_1 \in \normalfont\text{dom}(b) \wedge b(\sigma_1) \neq \varepsilon$}
        	{
        		let $b(\sigma_1) = (t_1, s_1) \cdot \ldots \cdot (t_n, s_n)$\;
        		$i \gets 1$\;
        		\While{$i \leq n \wedge c \in \normalfont\text{dom}(t_i)$} {
        			$c \gets t_i(c)$\;
        			$i \gets i +1$\;
        		}
        		$b(\sigma_1) \gets (t_i, s_i) \cdot \ldots \cdot (t_n, s_n)$\;
	        }
	        \Else
	        {
	        	$c \gets (G', \sigma', \varepsilon, \rho')$\;
            }
        }
        \Return{$c$}
}
\caption{Partial generation algorithm}\label{alg:partial-generation}
\end{algorithm}

The partial generation algorithm allows us to process a graph even if the required transitions for some vertices are still unknown; it does so by simply ignoring these vertices. However, we are still unable to actually assign language model scores to partial transition functions. This is because we must apply $\cf$ to obtain a sentence from a configuration, but $\cf$ can only be applied to states whose first component is a totally ordered graph $G$ and whose annotation function $\rho$ assigns a realization to each node contained within said graph; otherwise, $\text{yield}_{\rho( \anno{REAL} )}(G)$ would not be defined. We therefore generalize $\text{yield}$ to a \emph{partial yield function} which allows for arbitrary acyclic graphs and partial realization functions.

\begin{definition}[Partial yield]
\label{def:partial-yield}
Let $G = (V,E,L,\prec)$ be an acyclic graph. Furthermore, let $\Sigma$ be an alphabet, $V \subseteq V'$ and $\rho: V' \pfun \Sigma^*$. The function $\text{yield}^\text{par}_{(G,\rho)}: V \rightarrow \Sigma^*$ is defined for each $v \in V$ as
\begin{align*}
\text{yield}^\text{par}_{(G,\rho)}(v) = 
\begin{cases}
* & \text{ if $\prec$ is a total order on } \ch{v} \cup \{v\} \text{ and } v \in \text{dom}(\rho) \\
\varepsilon & \text{ otherwise.}
\end{cases}
\end{align*}
where
\[ * := \text{yield}^\text{par}_{(G,\rho)}(c_1) \cdot \ldots \cdot \text{yield}^\text{par}_{(G,\rho)}(c_k) \cdot \rho(v) \cdot \text{yield}^\text{par}_{(G,\rho)}(c_{k+1}) \cdot \ldots \cdot \text{yield}^\text{par}_{(G,\rho)}(c_{|\ch{v}|}) \]
and $(c_1, \ldots, c_k, v, c_{k+1}, \ldots, c_{|\ch{v}|})$, $k \in [|\ch{v}|]_0$ is the $(\ch{v} \cup \{v\})$-sequence induced by $\prec$. If $G$ is rooted, we write $\text{yield}^\text{par}_\rho(G)$ as a shorthand for $\text{yield}^\text{par}_{(G,\rho)}(\text{root}(G))$.
\end{definition}

From the above definition it is easy to see that $\text{yield}^\text{par}_{(G,\rho)}(v)$ behaves almost like $\text{yield}_{(G,\rho)}(v)$, the only difference being that the partial yield function sets the realization of all unprocessed nodes to $\varepsilon$ and ignores all $v'$-subtrees of $\restr{G}{v}$ for which no total order among $\ch{v'} \cup \{v'\}$ is specified. 

We are now able to make the desired generalization of our $\text{score}$ function so that it is not only applicable to terminating transition sequences, but also to partial transition functions given an initial configuration. For this purpose, let $c$ be a configuration and $b$ be a partial transition function. Furthermore, let $\text{generatePartial}(c,b) = (G, \sigma, \beta, \rho)$ and $v \in V$. We define the \emph{partial score of $b$ at $v$ given $c$} to be 
\begin{equation}
\text{score}^\text{par}(c,b,v) = \theta_\text{LM} \cdot \text{score}_\text{LM}(\text{yield}_{(G,\rho(\anno{REAL}))}^\text{par} (v)) +  \sum_{v' \in \text{dom}(b)}\text{score}_\text{TS}^\text{par}(b(v'))
\end{equation}
where
\[
\text{score}_\text{TS}^\text{par}(s) = \sum_{i = 1}^{n} \theta_{\mathcal{C}(t_i)} \cdot \log{p_i}
\]
for all $s = (t_1, p_1) \cdot \ldots \cdot (t_n, p_n) \in (T_\text{AMR} \times [0,1])^*$ and for all $\tau \in \mathcal{C}(T_\text{AMR})$, $\theta_\tau$ denotes the hyperparameter by the same name introduced in Eq.~(\ref{equ:score}).

\begin{figure}[t!]
\centering
\scalebox{0.8} {
\begin{tikzpicture}[baseline=(current bounding box.center)]
\tikzstyle{amr-node}=[shape=ellipse,draw, inner sep=0.2, minimum height=0.8cm, minimum width=1cm]

    \node[amr-node] (amr-want) at (0,0) {1\,:\,want-01};
    \node[amr-node] (amr-he) at (-1.5,-4) {2\,:\,he};
    \node[amr-node] (amr-go) at (1.5,-6.2) {3\,:\,go-01};
    \node[amr-node] (amr-he-link) at (1.5,-8.5) {4\,:\,he};
  
        \path [-latex](amr-want) edge node[fill=white, pos=0.6] {ARG0} (amr-he);
        \path [-latex](amr-want) edge node[fill=white, pos=0.6] (amr-want-arg1) {ARG1} (amr-go);
        \path [-latex](amr-go) edge node[fill=white, pos=0.7] {ARG0} (amr-he-link);
    
        \node[below= 0.1cm of amr-he, fill=white, draw, dashed, align=left] (rho-he) {\footnotesize $\anno{DEL} \mapsto 0$};
        \node[below=0.1cm of amr-want, fill=white, draw, dashed, align=left] (rho-want) {\footnotesize $\anno{DEL} \mapsto 0$};
        \node[below=0.1cm of amr-go, fill=white, draw, dashed] (rho-go) {\footnotesize $\anno{DEL} \mapsto 0$};
        \node[below=0.1cm of amr-he-link, align=left, fill=white, draw, dashed] (rho-he-link) 
        		{\footnotesize $\anno{DEL} \mapsto 1$ \\ \footnotesize $\anno{LINK} \mapsto 2$};
        \node[align=center] (buffers) at (0,-11) {$\sigma = \varepsilon \quad \beta = \varepsilon$ \\ ${\prec} = \emptyset$};
\end{tikzpicture}
{\Large $\quad\rightarrow\quad$}
\begin{tikzpicture}[baseline=(current bounding box.center)]
\tikzstyle{amr-node}=[shape=ellipse,draw, inner sep=0.2, minimum height=0.8cm, minimum width=1cm]

    \node[amr-node] (amr-want) at (0,0) {1\,:\,want-01};
    \node[amr-node] (amr-he) at (-1.5,-4) {2\,:\,he};
    \node[amr-node] (amr-to) at(1.5, -4) {5\,:\,to};
    \node[amr-node] (amr-go) at (1.5,-6.2) {3\,:\,go-01};
    \node[amr-node] (amr-he-link) at (1.5,-8.5) {4\,:\,he};
    
        \path [-latex](amr-want) edge node[fill=white, pos=0.85] {ARG0} (amr-he);
        \path [-latex](amr-want) edge node[fill=white, pos=0.85] {ARG1} (amr-to);
        \path [-latex](amr-to) edge node[fill=white, pos=0.7] {$\star$} (amr-go);
        \path [-latex](amr-go) edge node[fill=white, pos=0.7] {ARG0} (amr-he-link);
    
        \node[below= 0.1cm of amr-to, fill=white, draw, dashed, align=left] (rho-to) {\footnotesize $\anno{REAL} \mapsto \text{to}$};
        \node[below= 0.1cm of amr-he, fill=white, draw, dashed, align=left] (rho-he) {
        \footnotesize $\anno{DEL} \mapsto 0$ \\ 
        \footnotesize $\anno{REAL} \mapsto \text{he}$ \\ 
        \footnotesize $\anno{INS-DONE} \mapsto 1$ \\
        \footnotesize $\anno{POS} \mapsto \text{PRP}$ \\
        \footnotesize $\anno{DENOM} \mapsto \text{--}$ \\
        \footnotesize $\anno{TENSE} \mapsto \text{--}$ \\
        \footnotesize $\anno{NUMBER} \mapsto \text{--}$ \\
        \footnotesize $\anno{VOICE} \mapsto \text{--}$ 
        };
        \node[below=0.1cm of amr-want, fill=white, draw, dashed, align=left] (rho-want) {
           \begin{tabular}{@{}ll@{}}
            \footnotesize $\anno{DEL} \mapsto 0$ & \footnotesize $\anno{DENOM} \mapsto \text{--}$ \\
            \footnotesize $\anno{REAL} \mapsto \text{wants}$ & \footnotesize $\anno{TENSE} \mapsto \text{present}$\\
            \footnotesize $\anno{INS-DONE} \mapsto 1$ & \footnotesize $\anno{NUMBER} \mapsto \text{--}$ \\
            \footnotesize $\anno{POS} \mapsto \text{VB}$ & \footnotesize $\anno{VOICE} \mapsto \text{active}$  \\
            \end{tabular} };
        \node[below=0.1cm of amr-go, fill=white, draw, dashed] (rho-go) {\footnotesize $\anno{DEL} \mapsto 0$};
        \node[below=0.1cm of amr-he-link, align=left, fill=white, draw, dashed] (rho-he-link) 
        		{\footnotesize $\anno{DEL} \mapsto 1$ \\ \footnotesize $\anno{LINK} \mapsto 2$};
        \node[align=center] (buffers) at (0,-11) {$\sigma = \varepsilon \quad \beta = \varepsilon$ \\ ${\prec'} = \{ (2,1), (1,5), (5,3) \}^+$};
\end{tikzpicture}
}
\caption{Application of Algorithm~\ref{alg:partial-generation} where $b$ is the partial transition function described in Example~\ref{example:partial-generation}, $c$ is shown on the left and the resulting configuration $\text{generatePartial}(c,b)$ is shown on the right.}
\label{fig:partial-generation}
\end{figure}
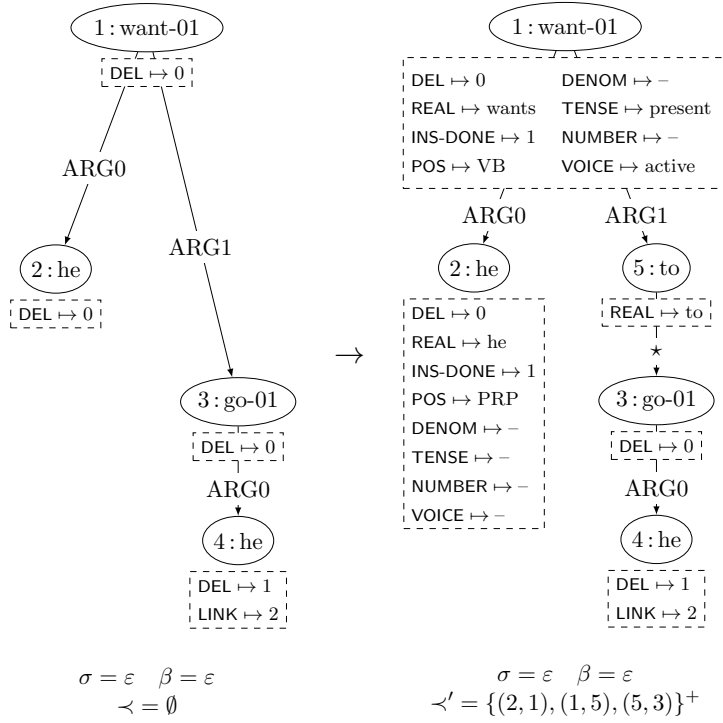

\begin{example}
\label{example:partial-generation}

We consider the partial transition function $b_1 \colon V \pfun (T_\text{AMR} \times [0,1])^*$ where $\text{dom}(b_1) = \{1, 2\}$ and
\begin{align*}
b_1(1) = &\ ( \textsc{Realize-}(\text{wants}, a_1), 0.75 ) \cdot ( \textsc{No-Insertion}, 0.8 ) \cdot (\textsc{Reorder-}(2, 1, 3), 0.01) \\
&\ \cdot (\textsc{No-Insertion}, 0.9) \cdot (\textsc{Insert-Between-}(\text{to},\textsf{left}), 0.4) \\
b_1(2) = &\ ( \textsc{Realize-}(\text{he}, a_2), 0.9 ) \cdot ( \textsc{No-Insertion}, 0.95 ) \cdot (\textsc{Reorder-}(2), 1) \\
a_1 = &\ \{ (\anno{POS}, \text{VB}), (\anno{DENOM}, \text{--}), (\anno{TENSE}, \text{present}), (\anno{NUMBER}, \text{--}), (\anno{VOICE}, \text{active}) \} \\ 
a_2 = &\ \{ (\anno{POS}, \text{PRP}), (\anno{DENOM}, \text{--}), (\anno{TENSE}, \text{--}), (\anno{NUMBER}, \text{--}), (\anno{VOICE}, \text{--}) \}\,.
\end{align*}
Additionally, we consider the state $c = (G, \sigma, \beta, \rho)$ shown in Figure~\ref{fig:partial-generation} where $G = (V,E,L,\prec)$ and $\rho$ is represented as follows: For each $k \in \mathcal{K}$ and each $v \in \text{dom}(\rho(k))$, the box directly below the graphical representation of $v$ is inscribed with $k \mapsto \rho(k)(v)$. The result of applying the partial generation algorithm, $\text{generatePartial}(c,b) = (G', \sigma', \beta', \rho')$ with $G' = (V', E', L', \prec')$ is shown in the right half of Figure~\ref{fig:partial-generation}. It holds that
\begin{align*}
\text{yield}_{{\rho'}(\anno{REAL})}^\text{par} (G') & = \text{yield}_{(G',{\rho'}(\anno{REAL}))}^\text{par}(2) \cdot {\rho'}(\anno{REAL})(1) \cdot \text{yield}_{(G',{\rho'}(\anno{REAL}))}^\text{par}(5) \\
& = {\rho'}(\anno{REAL})(2) \cdot {\rho'}(\anno{REAL})(1) \cdot {\rho'}(\anno{REAL})(5) \cdot \text{yield}_{(G',{\rho'}(\anno{REAL}))}^\text{par}(3) \\
& = {\rho'}(\anno{REAL})(2) \cdot {\rho'}(\anno{REAL})(1) \cdot {\rho'}(\anno{REAL})(5) \cdot \varepsilon = \text{he wants to} \,.
\end{align*}
Let $\theta_{\tau} = 1$ for all $\tau \in \mathcal{C}(T_\text{AMR})$. Then 
\[
\text{score}^\text{par}(c,b,1) = \theta_\text{LM} \cdot \text{score}_\text{LM}(\text{he wants to}) + \text{score}_\text{TS}^\text{par}(b(1)) + \text{score}_\text{TS}^\text{par}(b(2)) 
\]
where
\begin{align*}
\text{score}_\text{TS}^\text{par}(b(1)) & = \log 0.75 + \log 0.8 + \log 0.01 + \log 0.9 + \log 0.4 \\
\text{score}_\text{TS}^\text{par}(b(2)) & = \log 0.9 + \log 0.95 + \log 1\,. \qedhere
\end{align*}
\end{example}

While we are now able to compute scores for partial transition sequences, it is still unclear how a good such sequence for a given input $G = (V, E, L, \prec)$ can efficiently be found. Our approach is to create a set of candidate partial transition functions for each $v$-subgraph of $G$ bottom-up, factoring in the language model at each step. More formally, we successively construct a function $\text{best} \colon V \rightarrow \mathcal{P}(\mathcal{T}_\text{AMR}^\text{par} \times \mathbb{R})$ such that for each ${v \in V}$, $\text{best}(v) = \{(b_1, s_1), \ldots, (b_n, s_n) \}$ contains partial transition functions $b_1, \ldots b_n$ that specify transitions for exactly the nodes of $\restr{G}{v}$, i.e. ${b_i \colon \text{succ}(v) \cup \{ v \} \rightarrow (T_\text{AMR} \times [0,1])^*}$ for all $i \in [n]$; each number $s_i$ is the partial score of the corresponding partial transition function $b_i$. 
Before we give an actual algorithm to calculate $\text{best}(v)$, we define two important functions of which we will make use in said algorithm.

\begin{definition}[All]
The mapping $\text{all} \colon C_\text{AMR} \rightarrow \mathcal{P}(T_\text{AMR} \times \mathbb{R})$, defined by
\[
\text{all}(c) = \{ (t, p) \in T_\text{AMR} \times \mathbb{R} \mid c \in \text{dom}(t) \wedge p = P(t \mid c) \}
\]
for all $c \in C_\text{AMR}$, assigns to each configuration $c$ the set of all applicable transitions along with their probabilities. 
\end{definition}

\begin{definition}[Prune]
\label{def:prune}
Let $A$ be a set, $S = \{ (a_1, p_1), \ldots, (a_m, p_m) \} \in \mathcal{P}(A \times \mathbb{R})$ be a set, $n \in \mathbb{N}$ and $r \in \mathbb{R}^+_0$. The set $\text{prune}_{n}(S)$ is defined recursively by
\[
\text{prune}_n(S) = 
 \begin{cases}
 \emptyset & \text{if } S = \emptyset \vee n = 0 \\
 \{ \hat{s} \} \cup \text{prune}_{n-1}(S \setminus \{ \hat{s} \} ) & \text{otherwise}
 \end{cases}
\]
where $\hat{s} = \argmax_{(a,p) \in S} p$.
In other words, $\text{prune}_n (S)$ is the set obtained from $S$ by including only the $k = \min(n,m)$ pairs $(a_i, p_i)$ with the highest scores $p_i$. We define
\[
\text{prune}_{(n,r)}(S) = \{ (a, p) \in \text{prune}_n(S) \mid p \geq p_\text{max} - r \}
\]
where $p_\text{max} = \max_{(a, p) \in S} p$. That is, $\text{prune}_{(n,r)}(S)$ is obtained from $\text{prune}_n (S)$ by retaining only pairs for which the score is lower than $p_\text{max}$ by at most  $r$.
\end{definition}

\begin{example}
Let $A = \{ \alpha, \beta, \gamma, \delta \}$ and $S = \{ (\alpha, 0.9), (\beta, 0.3), (\gamma, 0.8), (\delta, 0.45) \}$. The following holds true:
\begin{align*}
\text{prune}_n(S) & = S \text{ for } n \geq 4 \\
\text{prune}_3(S) & = \{ (\alpha, 0.9), (\gamma, 0.8), (\delta, 0.45) \} \\
\text{prune}_{(3,\, 0.15)}(S) & = \{ (\alpha, 0.9), (\gamma, 0.8) \}\,. \qedhere
\end{align*}
\end{example}

With the help of the above definitions, we can now formulate Algorithm~\ref{alg:lm-generation} that, given an initial state $c \in C_\text{AMR}$, a node $v \in V$ and a partial function $\text{best} \colon V \pfun \mathcal{P}(\mathcal{T}_\text{AMR}^\text{par} \times \mathbb{R})$ with $\text{succ}(v) \subseteq \text{dom}(\text{best})$, computes the set $\text{best}(v)$ containing an approximation of the best transition sequences for $\text{succ}(v) \cup \{v\}$. We call this algorithm the \emph{best transition sequence algorithm} and refer to its output given the above input by $\text{getBest}(v, c, \text{best})$. Note that this algorithm makes use of hyperparameters $h_i = (n_i, r_i) \in \mathbb{N}^+ \times \mathbb{R}^+_0$, $i \in [5]$. These tuples are used in several places for pruning the number of transitions to be considered; the maximum size of $\text{best}(v)$ is determined by $n_5$. 

\begin{algorithm}[]
\DontPrintSemicolon
\SetArgSty{textrm}
\KwIn{configuration $c = (G, \varepsilon, \varepsilon, \rho) \in C_\text{AMR}$ with $G = (V, E, L , \prec)$, vertex~$v \in V$~with $\rho(\anno{DEL})(v) = 0$ and $v \notin \text{dom}(\rho(\anno{REAL}))$, function~$\text{best} \colon V \pfun \mathcal{P}(\mathcal{T}_\text{AMR}^\text{par} \times \mathbb{R})$ such that $\text{succ}(v) \subseteq \text{dom}(\text{best})$}
\KwOut{$n_5$-best transition sequences for $\text{succ}(v) \cup \{v\}$}
\SetKwBlock{Begin}{function}{end function}
\SetKwRepeat{Do}{do}{while}
\Begin($\text{getBest} {(} v, c, \text{best} {)}$)
{

 $c \gets (G, v, \varepsilon, \rho)$\; \label{line:new_c}
 $\text{best}(v) \gets \emptyset$\; \label{line:init_best}
        \For{$(t_{\textsf{real}}, s_{\textsf{real}}) \in \text{prune}_{h_1}(\text{all}(c))$}
        { \label{line:real_iter}
        	$\text{hist} \gets (t_\textsf{real}, s_\textsf{real})$\; \label{line:real_hist}
           	$c_\textsf{real} \gets  t_\textsf{real}(c)$\; \label{line:real_apply}
         
           	\Repeat{$t^* = {\textsc{No-Insertion}}$ \label{line:c_ins_end} } { \label{line:c_ins_start}
        		$T^* \gets \{ t \in T_\text{AMR} \mid c_\textsf{real} \in \text{dom}(t) \}$\;
	            $t^* \gets \argmax_{t \in T^*} P(t \mid c_\textsf{real})$\;
	            $\text{hist} \gets \text{hist} \cdot (t^*, P(t^* \mid c_\textsf{real}))$\;
        		$c_\textsf{real} \gets t^*(c_\textsf{real})$\;
        		\If{$t^* \neq \textsc{No-Insertion}$}
        		{
	        		let $c_\textsf{real} = (G', (\tilde{\sigma}, v), \varepsilon, \rho')$\;
	        		$\text{best}(\tilde{\sigma}) \gets \text{getBest}(\tilde{\sigma}, c_\textsf{real}, \text{best})$\;
	        		$c_\textsf{real} \gets (G', v, \varepsilon, \rho')$\;
        		}
           	}
           	
            \For{$(t_\textsf{reor}, s_\textsf{reor}) \in \text{prune}_{h_2}(\text{all}(c_\textsf{real}))$}
            { \label{line:reor_iter}
            	$\text{hist} \gets \text{hist} \cdot (t_\textsf{reor}, s_\textsf{reor})$\; \label{line:reor_hist}
              	$c_\text{reor} \gets  t_\textsf{reor}(c_\textsf{real})$\; \label{line:reor_apply}
              	let $c_\text{reor} = (G', \sigma, (\beta_1, \ldots, \beta_n), \rho')$\;
              	$b_0 \gets \{ (v, \text{hist}) \}$\;
              	$\text{best}_{\leq 0}(v) \gets \{ (v, \{ (b_0, 1) \} ) \}$\; \label{line:best_0}
              	\For{$i \gets 1, \ldots, n$}
              	{
              		$c_i \gets (G', \sigma, \beta_i, \rho')$\;
              		$\text{best}_{\leq i}(v) \gets \emptyset$\;
              		\For{$b \in \text{best}_{\leq i-1}(v)$}
              		{
              			\For{$b_i \in \text{best}(\beta_i)$}
              			{
              				\For{$(t_\textsf{insb}, s_\textsf{insb}) \in \text{prune}_{h_3}(\text{all}(c_\textsf{reor}))$}
              				{
              					$b_\text{new} \gets b[v \mapsto b(v) \cdot  (t_\textsf{insb}, s_\textsf{insb}) ] \cup b_i $\;
              					$s_\text{new} \gets \text{score}^\text{par}(c, b_\text{new}, v)$\; \label{line:score_new}
              					$\text{best}_{\leq i}(v) \gets \text{prune}_{h_4}(\text{best}_{\leq i}(v) \cup \{ (b_\text{new}, s_\text{new}) \})$\; \label{line:best_leq_i}
              				}
              			}
              		}	
              	}
              	
              	$\text{best}(v) \gets \text{prune}_{h_5}(\text{best}(v) \cup \text{best}_{\leq n}(v))$\; \label{line:best_v}
              	
           }
        }
        \Return{$\text{best}(v)$}
}
\caption{Best transition sequence algorithm}\label{alg:lm-generation}
\end{algorithm}

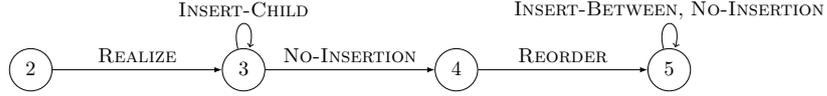
\begin{figure}[t]
\centering
\scalebox{0.7} {
\begin{tikzpicture}
\tikzstyle{p-node}=[shape=ellipse,draw, inner sep=0.2, minimum size=0.8cm, text height=1.5ex, text depth=.25ex]

    \node[p-node] (2) at (-6,0) { 2 };
    \node[p-node] (3) at (-2,0) { 3 };
    \node[p-node] (4) at (2,0) { 4 };
    \node[p-node] (5) at (6,0) { 5 };

    \path [-latex](2) edge node[above] (r) {\textsc{Realize}} (3);
    \path [-latex](3) edge[loop above] node[above, align=center] {\textsc{Insert-Child}} (3);
    \path [-latex](3) edge node[above, align=center] {\textsc{No-Insertion}} (4);
    \path [-latex](4) edge node[above, align=center] {\textsc{Reorder}} (5);
    \path [-latex](5) edge[loop above] node[above, align=center, inner sep=0.07cm] {\textsc{Insert-Between}, \textsc{No-Insertion}} (5);
    
	\node (pad1) at (-10, 0) {};
	\node (pad2) at (10, 0) {}; 
\end{tikzpicture}
}
\caption{Representation of the order in which transitions from $T_\text{AMR} \setminus T_\text{restr}$ can be applied}
\label{fig:transition-sequences2}
\end{figure}

As the best transition sequence algorithm is far more complex than the ones previously shown, we give a more detailed explanation. For this purpose, we again consider the five stages of processing a node shown in Figure~\ref{fig:transition-sequences}; the stages relevant for Algorithm~\ref{alg:lm-generation} are recapped in Figure~\ref{fig:transition-sequences2}. Algorithm~\ref{alg:lm-generation} processes the input node $v$ from stage~2 to stage~5, each time considering multiple possible transitions:

\begin{itemize}
\item Line \ref{line:new_c} -- \ref{line:init_best}: Configuration $c$ is slightly modified as we are interested in the sequence of transitions to apply when $v$ is on top of the node buffer; $\text{best}(v)$ is set to $\emptyset$.

\item Line \ref{line:real_iter}: Given $c = (G, v, \varepsilon, \rho)$, all applicable transitions belong to the class \textsc{Realize}; this follows directly from the fact that $\rho(\anno{DEL})(v) = 0$ and there is no realization assigned to $v$. The $n_1$-best \textsc{Realize-}$(w,\alpha)$ transitions are obtained through $\text{all}(c)$.

\item Line \ref{line:real_hist} -- \ref{line:real_apply}: The currently chosen \textsc{Realize}-$(w,\alpha)$ transition $t_\textsf{real}$ is stored in a sequence $\text{hist}$ and applied to $c$; we thereby move from stage $2$ to stage $3$.

\item Line \ref{line:c_ins_start} -- \ref{line:c_ins_end}: The most likely \textsc{Insert-Child} transitions are greedily applied until the best transition is \textsc{No-Insertion}. For each newly inserted vertex $\tilde{\sigma}$, the set of best transition sequences $\text{best}(\tilde{\sigma})$ is determined. Through application of \textsc{No-Insertion}, we move from stage~3 to stage~4. 

\item Line \ref{line:reor_iter}: Given configuration $c_\text{real}$, only \textsc{Reorder} transitions can be applied; we obtain the $n_2$-best \textsc{Reorder}-$(v_1, \ldots, v_n)$ transitions from $\text{all}(c_\text{real})$.

\item Line \ref{line:reor_hist} -- \ref{line:reor_apply}: The current \textsc{Reorder}-$(v_1, \ldots, v_n)$ transition $t_\textsf{reor}$ is stored in $\text{hist}$ and applied to $c_\text{real}$; the final stage of processing $v$ is reached.

\item Line \ref{line:best_0} -- \ref{line:best_leq_i}: We successively construct sets $\text{best}_{\leq i}(v) \subseteq \mathcal{T}_\text{AMR}^\text{par} \times \mathbb{R}$, $i \in [n]$ that, given state $c_\text{reor}$, store the best partial transition sequences for $v$, its children $\beta_1, \ldots, \beta_i$ and their successors. Accordingly, $\text{best}_{\leq 0}(v)$ contains only transitions previously applied to $v$; these transitions are inferred from $\text{hist}$. The set $\text{best}_{\leq i}(v)$ is obtained by iterating over all partial transition functions in both $\text{best}_{\leq i-1}(v)$ and $\text{best}(\beta_i)$ as well as the $n_3$-best \textsc{Insert-Between} (or \textsc{No-Insertion}) transitions for $v$ and $\beta_i$, computing the corresponding partial transition function $b_\text{new}$ along with its score and collecting the $n_4$-best so-obtained functions. In other words, we combine the best partial transition functions for $\{v\} \cup \bigcup_{j = 1}^{i-1} ( \{\beta_j\} \cup \text{succ}(\beta_j) )$ with the best partial transition functions for $\{\beta_i\} \cup \text{succ}(\beta_i)$ and the best applicable transitions when $v$ is on top of the node buffer and $\beta_i$ is on top of the child buffer. 

\item Line \ref{line:best_v}: For each considered \textsc{Realize}-$(w,\alpha)$ and \textsc{Reorder}-$(v_1, \ldots, v_n)$ transition, the set $\text{best}_{\leq n}(v)$ is added to $\text{best}(v)$ which is then pruned to obtain only the $n_5$-best partial transition functions.
\end{itemize}

This concludes our discussion of the best transition sequence algorithm. We note that this algorithm is currently only defined for vertices $v$ where $\rho(\anno{DEL})(v) = 0$. However, it can easily be extended to support also vertices with $\rho(\anno{DEL})(v) = 1$. We do not explicitly write down this extension, but it can be derived from Algorithm~\ref{alg:lm-generation} by simply skipping both the realization of $v$ and all possible insertions, i.e. only considering possible reorderings. Whenever we refer to $\text{getBest}(v, c, \text{best})$ in the future, we explicitly mean this modified version that works for each vertex $v$ regardless of $\rho(\anno{DEL})(v)$.

In a last step, we combine Algorithms~\ref{alg:greedy-generation}~to~\ref{alg:lm-generation} and construct Algorithm~\ref{alg:generation}, our final generation algorithm that takes as input an AMR graph $G$ and outputs $\tilde{w}$, the desired approximation of $\hat{w}$ as defined in Eq.~(\ref{equ:generator}): We first apply the restricted version of Algorithm~\ref{alg:greedy-generation} to $G$, resulting in a state of the form $c = (G', \varepsilon, \varepsilon, \rho)$. Subsequently, we compute the sets $\text{best}(v)$ for each node $v$ in $G'$ bottom-up using Algorithm~\ref{alg:lm-generation}. Finally, Algorithm~\ref{alg:partial-generation} is applied to $c$ using $\hat{b}$, the best partial transition function found for the root of $G'$. Note that $\hat{b}$ is guaranteed to assign a \textsc{Realize} and \textsc{Reorder} transition to every node of $G'$, so we can apply $\cf$ to the resulting configuration.

\begin{algorithm}
\DontPrintSemicolon
\KwIn{AMR graph $G = (V, E, L , \prec)$}
\KwOut{generated sentence $\tilde{w} \in \Sigma_\text{E}^*$}
\SetKwBlock{Begin}{function}{end function}
\Begin($\text{generate} {(} G {)}$)
{
 $c = (G', \varepsilon, \varepsilon, \rho) \gets \text{generateGreedy}_\text{restr}(G)$\; \label{alg:gen-line:gengreedy}
 let $\sigma = (\sigma_1, \ldots, \sigma_n)$ be a bottom-up traversal of all nodes in $G'$\; \label{alg:gen-line:butraversal}
 	$\text{best} \gets \emptyset$\;
	\For{$i \gets 1, \ldots, n$}
  	{ \label{alg:gen-line:bestiter}
  		$\text{best} \gets \text{best} \cup \{ (\sigma_i , \text{getBest}(\sigma_i, c, \text{best})) \}$\; \label{alg:gen-line:best}	
  	}
    $(\hat{b},\hat{s}) \gets \argmax_{(b,s) \in \text{best}(\text{root}(G'))}{ s }$\; \label{alg:gen-line:argmax}
    $\hat{c} \gets \text{generatePartial}(c, \hat{b})$\; \label{alg:gen-line:genpartial}
    $\tilde{w} \gets \cf(\hat{c})$\; \label{alg:gen-line:famr}       	
              	      
    \Return{$\tilde{w}$}
}
\caption{Generation algorithm}\label{alg:generation}
\end{algorithm}

\subsubsection{Complexity Analysis}

We derive a theoretical upper bound for the number $N(G)$ of operations required to compute $\tilde{w} = \text{generate}(G)$ for an AMR graph $G$ using Algorithm~\ref{alg:generation}.
Before we derive this upper bound, we add several constraints to our transition system, limiting the number of possible transitions. For example, the number of $\textsc{Insert-Child}$ transitions that can be applied to a vertex is currently unlimited, resulting in $N(G)$ being unbounded; we therefore set the maximum number of $\textsc{Insert-Child}$ transitions per vertex to some constant $C_\text{ins} \in \mathbb{N}$.
We additionally demand that \textsc{Swap} is never applied to vertices added through \textsc{Delete-Reentrance} transitions and, as is done in \citet{wang2015transition}, that \textsc{Swap} can not be reversed; that is, if a \textsc{Swap} transition was applied to some vertex $v$ with parent $p_v$, it may not be applied to $p_v$ with parent $v$ in a subsequent step. 
For our study of Algorithm~\ref{alg:generation}, let $G = (V,E,L,\prec)$ be the input AMR graph. Furthermore, let $G' = (V', E', L', \prec')$ be the graph constructed in line~\ref{alg:gen-line:gengreedy} and $\hat{c} = (\hat{G}, \varepsilon, \varepsilon, \hat{\rho})$ with $\hat{G} = (\hat{V}, \hat{E}, \hat{L}, \hat{\prec})$ be the configuration obtained in line~\ref{alg:gen-line:genpartial}. 

Finding a bottom-up traversal of all vertices in $G'$ (line~\ref{alg:gen-line:butraversal}) requires us to completely process all nodes therein once; it therefore takes $\mathcal{O}(|V'|)$ steps. Similarly, computing $\cf(\hat{c})$ (line~\ref{alg:gen-line:famr}) requires $\mathcal{O}(|\hat{V}|)$ steps. As for each $v \in \text{dom}(\text{best})$, $|\text{best}(v)| \leq n_5$ where $n_5$ is the hyperparameter introduced in Algorithm~\ref{alg:lm-generation}, finding the $\argmax$ (line~\ref{alg:gen-line:argmax}) requires $\mathcal{O}(n_5)$ steps. We will see below that all these operations are negligible compared to the number of steps required by the subroutines called in lines~\ref{alg:gen-line:gengreedy},~\ref{alg:gen-line:best}~and~\ref{alg:gen-line:genpartial}. For each of these three subroutines, we assume all operations performed therein to require only a constant number of atomic steps and we denote the number of executed such operations by $N_1$, $N_2$ and $N_3$, respectively. 

We first discuss the complexity of $\text{generateGreedy}_\text{restr}(G)$ as called in line~\ref{alg:gen-line:gengreedy} of the generation algorithm. As the restricted version of the greedy generation algorithm only considers transitions from the set $T_\text{restr}$, we can derive
\[
N_1 \in \mathcal{O}(\sum_{\tau \in \mathcal{C}(T_\text{restr})} N_1'(\tau))
\]
where for each $\tau \in \mathcal{C}(T_\text{restr})$, $N_1'(\tau)$ is an upper bound for the number of transitions from $\tau$ applied during the processing of $G$. As each \textsc{Delete-Reentrance} transition removes an edge and no other transition from $T_\text{restr}$ increases the number of edges, we can easily derive the upper bound $N_1'(\textsc{Delete-Reentrance}) = |E|$. Similarly, each \textsc{Merge} transition removes a vertex and as $\textsc{Delete-Reentrance}$ may add up to $|E|$ new vertices, we obtain the upper bound $N_1'(\textsc{Merge}) = |V| + |E|$. For each pair of vertices, at most one \textsc{Swap} transition can be applied and vertices inserted by $\textsc{Delete-Reentrance}$ can not be swapped; therefore, $N_1'(\textsc{Swap}) = |V|^2$ is an upper bound for the number of \textsc{Swap} transitions. Finally, we derive $N_1'(\textsc{Delete}) + N_1'(\textsc{Keep}) = |V| + |E|$ from the fact that each vertex is either kept or deleted and this is decided exactly once.
From these considerations, we can conclude that $N_1 \in \mathcal{O}(|E| + |V|^2)$. Furthermore, we can easily derive~$|V'| \leq |V| + |E|$.

We now consider the subroutine $\text{getBest}(\sigma_i, c, \text{best})$ called in line~\ref{alg:gen-line:best}. For this purpose, let $C_\text{max} = \max_{v \in V}|\ch[G]{v}|$ be the maximum number of children for all nodes in $G$. A straightforward analysis of the for-loops in Algorithm~\ref{alg:lm-generation} gives 
\[
N_2 \in \mathcal{O}( n_1 \cdot ( C_\text{ins} \cdot N_\text{ins} + n_2 \cdot (C_\text{max} + C_\text{ins}) \cdot n_4 \cdot n_5 \cdot n_3) )
\]
where the term $C_\text{ins} \cdot N_\text{ins}$ comes from the fact that up to $C_\text{ins}$ \textsc{Insert-Child} transitions may be applied and for each inserted child $\tilde{\sigma}$, routine $\text{getBest}$ is called recursively, requiring up to $N_\text{ins}$ additional operations. However, as inserted vertices have no children of their own and \textsc{Insert-Child} transitions are not applicable to them, $N_\text{ins}$ is in $\mathcal{O}(n_1)$. Due to our assumption of $C_\text{ins}$ being a constant, we can further simplify
\[
N_2 \in \mathcal{O}(n_1^2 + C_\text{max} \cdot \prod_{i=1}^5 n_i)\,.
\]
We must take into account that $\text{getBest}(\sigma_i, c, \text{best})$ is computed once for each node $v \in V'$ and, as shown before, $|V'| \leq |V| + |E|$.  However, for vertices $\tilde{\sigma}$ added through $\textsc{Delete-Reentrance}$ transitions, only $\mathcal{O}(n_1)$ operations are required to compute the set $\text{best}(\tilde{\sigma})$; the reasoning is the same as above in the case of vertices added through $\textsc{Insert-Child}$ transitions. Therefore, the number of operations required for executing lines~\ref{alg:gen-line:bestiter}~to~\ref{alg:gen-line:best} of the generation algorithm is
\[
N_2' \in \mathcal{O}(|V| \cdot N_2 + |E| \cdot n_1)\,.
\]

To compute $\text{generatePartial}(c, \hat{b})$ as called in line~\ref{alg:gen-line:genpartial}, a constant number of transitions needs to be applied to each vertex; the number of vertices is bounded by $|V| + |E|$. Additionally, up to $C_\text{max}$ \textsc{Insert-Between} or \textsc{No-Insertion} transitions are applied for each vertex with at least one child; in total, however, the number of such transitions is also bounded by $|V| + |E|$ as each node is at most once the top element of the child buffer $\beta$. The resulting number of operations for the partial generation algorithm is therefore
\[
N_3 \in \mathcal{O}(|V| + |E|)\,.
\]
As the number of transitions applied is constant in the number of vertices and so is the number of added vertices per transition, it follows directly that $|\hat{V}| \in \mathcal{O}(|V| + |E|)$. 

Combining all of the above considerations, we arrive at the sought-after upper bound
\[
N(G) \in \mathcal{O}(N_1 + N_2' + N_3) = \mathcal{O}(|E| + |V| \cdot (|V| + n_1^2 + C_\text{max} \cdot \prod_{i=1}^5 n_i ))
\]
for the number of operations required by the generation algorithm with input $G$.
As can be seen from the above equation, this number depends tremendously on the values chosen for hyperparameters $n_1$ to $n_5$. However, it is worth noting that in practice, the actual number of required operations is often well below this upper bound. For example, the number of \textsc{Swap} transitions required to process an AMR graph from one of the corpora discussed in Section~\ref{PRELIMINARIES:Corpora} is rarely higher than $3$, whereas our upper bound is quadratic in the number of vertices. We will further discuss the performance of Algorithm~\ref{alg:generation} from a practical point of view in Section~\ref{EXPERIMENTS}.


\subsection{Training}
\label{GENERATION:Training}

The aim of this section is to describe how the maximum entropy models introduced in Sections~\ref{GENERATION:SyntacticAnnotation} and \ref{GENERATION:Modeling} can be trained given an AMR corpus $C = ((G_1, w_1), \ldots, (G_n, w_n))$. We proceed as follows: As a first step, we derive in Section~\ref{TRAINING:Preparations} how an AMR corpus can be converted into the structure we use for our training process. In Section~\ref{TRAINING:SyntacticAnnotations}, we describe how the models required to estimate the probabilities of syntactic annotations can be learned. Finally, we show in Section~\ref{TRAINING:Transitions} how sequences of training data $(c,t) \in C_\text{AMR} \times T_\text{AMR}$ where $t$ is the right transition to be applied when $c$ is the current configuration can be extracted from $C$ to train the remaining maximum entropy models required for our transition system. We also describe the sequences of features to be used by all these models.

\subsubsection{Preparations}
\label{TRAINING:Preparations}

Let $C = ((G_1, w_1), \ldots, (G_n, w_n))$ be an AMR corpus. We extend this corpus to a sequence $C_\text{ext}$ from which both syntactic annotations and required transition steps can be inferred more easily. Let $(G,w) \in \mathcal{G}_\text{AMR} \times \Sigma_\text{E}^*$ be some element of $C$ and let $G = (V_G, E_G, L_G, \prec_G )$.
As a first preparation step, we convert $w$ to lower case and remove all punctuation from it, resulting in a new string $w' = w_1 \ldots w_m$, $m \in \mathbb{N}$, $w_i \in \Sigma_\text{E}$ for $i \in [m]$. We then utilize a dependency parser to generate the corresponding dependency tree $D = (V_D,E_D,L_D,\prec_D)$ as well as an alignment $A_{D} \subseteq V_D \times [m]$. As each vertex $v \in V_D$ corresponds to exactly one word of $w$, $A_D$ is guaranteed to be a bijective function. Next, we use a POS tagger to annotate each word $w_i$, $i \in [m]$ with its part of speech $p_i \in \mathcal{V}_\anno{POS}$; we abbreviate the obtained sequence $(w_1,p_1) \ldots (w_m, p_m)$ by $w^\anno{POS}$. 

As a final step, we try to obtain an alignment $A_{G} \subseteq V_G \times [m]$ that links each vertex $v \in V_G$ to its realization. To this end, we make use of two methods: Firstly, we use the aligner by \cite{pourdamghani2014aligning} which bijectively converts AMR graphs into strings and aligns the latter to realizations using the word alignment model described in \citet{brown1993mathematics}; the so obtained string-to-string alignment can then easily be converted into the desired format, resulting in the first candidate alignment ${A_\text{wa} \subseteq V_G \times [m]}$. Secondly, we use the rule-based greedy aligner by \cite{flanigan2014discriminative} to obtain another candidate alignment $A_\text{rb} \subseteq V_G \times [m]$. An important difference between these two approaches is that the aligner of \cite{flanigan2014discriminative} aligns each vertex to a \emph{contiguous} sequence of words. In other words, for each $v \in V_G$ that is aligned to at least one word, there are some $k, l \in \mathbb{N}$ such that 
\[
A_\text{rb}(v) = \{k, k+1, k+2, \ldots, k+l-1, k+l \}\,.
\]
This property is useful for our generator as the realization assigned to each vertex through \textsc{Realize} transitions is as well a contiguous sequence of words. Therefore, we also enforce this property upon $A_\text{wa}$ by removing from it for each vertex $v$ all tuples $(v,i)$ that do not belong to the first contiguous sequence aligned to $v$, beginning from the left; we denote the resulting alignment by $A_\text{wa}'$. As it is desirable for our generator that as many words as possible are aligned to some vertex, we construct a joint alignment $A$ by fusing both alignments. To this end, we take $A_\text{wa}'$ as a baseline; for every vertex that is not aligned to any word, we adopt the alignment assigned by $A_\text{rb}$, resulting in the alignment
\[
A = A_\text{wa}' \cup \{ (v,i) \in A_\text{rb} \mid \nexists j \in [m] \colon (v,j) \in  A_\text{wa}'  \} \,.
\]
We further improve upon this alignment by adding a small number of handwritten rules. For example, for unaligned vertices $v \in V_G$ whose concept consists of several words separated by hyphens (such as ``at-least''), we search for a contiguous sequence of precisely those words in the reference realization. If such a subsequence $w_i \ldots w_{i+j}$ of $w'$ is found and none of the corresponding words is already aligned to some vertex, we add $\{ (v, k) \mid i \leq k \leq i+j  \}$ to $A$. Also, we remove alignments to articles, auxiliary verbs and adpositions as these words should almost always be handled through \textsc{Insert-Child} and \textsc{Insert-Between} transitions and thereby get assigned their own, new vertices.

For a complete list of all handwritten alignment rules, we refer to Section~\ref{IMPLEMENTATION:Packages:dag}. We denote by $A_G$ the alignment obtained from $A$ by applying all handwritten rules to it. The components obtained during the preparation process can be joined together into a bigraph $\mathcal{B} = (G, D, w^\anno{POS}, A_G, A_D)$. Doing so for all elements of $C$ results in the desired extended corpus
\[ 
C_\text{ext} = ((G_1, D_1, w_1^\anno{POS}, A_{G_1}, A_{D_1}), \ldots, (G_n, D_n, w_n^\anno{POS}, A_{G_n}, A_{D_n}))
\]
which we require for our training process.

\subsubsection{Syntactic Annotations}
\label{TRAINING:SyntacticAnnotations}

Throughout this section, let $\mathcal{B} = (G, D, w^\anno{POS}, A_G, A_D)$ be an element of the extended corpus $C_\text{ext}$ as defined above where $G = (V_G, E_G, L_G, \prec_G)$ and $w^\anno{POS} = (w_1, p_1) \ldots (w_m, p_m)$. In the following, we first derive how for each vertex $v \in V_G$, the \emph{gold syntactic annotation} $\alpha_v \in \mathcal{A}_\text{syn}$ can be obtained from $\mathcal{B}$ and then describe how a maximum entropy model can be trained from the resulting sequence of tuples $(v, \alpha_v) \in V_G \times \mathcal{A}_\text{syn}$. 

In order to assign to some vertex $v \in V_G$ a meaningful syntactic annotation $\alpha_v$, the latter should somehow be inferred from the words to which $v$ is aligned; if there are no such words, i.e. $A_G(v) = \emptyset$, we ignore vertex $v$ during the training process. If~there are multiple such words, i.e. $|A_G(v)| \geq 2$, and these words differ with regards to their syntactic properties, we must somehow decide from which of them to infer the syntactic annotation of $v$. We do so in a very simple way by using a function $\text{bestPrefix}_\mathcal{B}: V_G \times \mathcal{P}([m]) \pfun [m]$ that, given a vertex $v$ and a nonempty set of word indices $S \subseteq [m]$, returns the index $i \in S$ such that $w_i$ has the longest common prefix with $L_G(v)$; if multiple such indices exist, the lowest one is chosen.

\begin{example}
\label{example:bestPrefix}
Let $\mathcal{B}_1 = (G_1, D_1, w_1^\anno{POS}, A_{G_1}, A_{D_1})$ be an element of the extended corpus $C_\text{ext}$ where $G_1 = (V,E,L,\prec)$, $V = \{ v_1, v_2, v_3, v_4\}$ and 
\begin{align*}
L & = \{ (v_1, \text{person}), (v_2, \text{develop-02}), (v_3, \text{delight-01}), (v_4, -) \} \\
w_1^\anno{POS} & = (\text{the, DT}) (\text{developer}, \text{NN}) (\text{is}, \text{VB}) (\text{not}, \text{RB}) (\text{delighted}, \text{JJ})\,.
\end{align*}
The following statements are true:
\[
\text{bestPrefix}_{\mathcal{B}_1}(v_2, \{2,5 \}) = 2 \quad
\text{bestPrefix}_{\mathcal{B}_1}(v_2, \{4,5 \}) = 5 \quad
\text{bestPrefix}_{\mathcal{B}_1}(v_1, \{1,2 \}) = 1 \,.
\]
Note that the last of the above statements is true although the longest common prefix of $L(v_1)$ with both $w_1$ and $w_2$ is equal to $\varepsilon$ because index $1$ is lower than $2$.
\end{example}

For the syntactic annotation key \anno{POS}, we consider only a subset of the POS tags used in the \emph{Penn Treebank Project} \citep{marcus1993building}.\footnote{A list of all POS tags used in the Penn Treebank Project can be found at \url{www.ling.upenn.edu/courses/Fall_2003/ling001/penn_treebank_pos.html}.} This subset is obtained by aggregating POS tags whenever a distinction between them is not relevant to our use case or can be inferred from the value assigned to some other syntactic annotation key. The function $\text{simplify} \colon \mathcal{V}_\anno{POS} \rightarrow \mathcal{V}_\anno{POS}$ that maps each POS tag to the simplified version we are interested in is defined by 
\[
\text{simplify}(p) = 
\begin{cases}
\text{NN} & \text{if } p \in \{ \text{NN, NNS, NNP, NNPS, FW} \} \\
\text{VB} & \text{if } p \in \{ \text{VB, VBD, VBP, VBZ} \} \\
\text{JJ} & \text{if } p \in \{ \text{JJ, JJR, JJS, RB, RBR, RBS, WRB} \} \\
p & \text{otherwise.}
\end{cases}
\]
In order to obtain gold syntactic annotations, we will sometimes be required to check whether a word $w$ is close to another word from some set $S \subseteq \Sigma_\text{E}$; for example, to find out a noun's denominator, we must check whether it has one of the words ``the'', ``a'' and ``an'' to its left. However, this word is not necessarily directly adjacent to $w$. We therefore define the mapping $\text{left}^\mathcal{B}_S \colon [m] \mapsto \{\text{true}, \text{false}\}$ as
\[
\text{left}^\mathcal{B}_S(i) = 
\begin{cases}
\text{true} & \text{if } w_{i-1} \in S \vee (w_{i-2} \in S \wedge \text{simplify}(p_{i-1}) = \text{JJ}) \\
\text{false} & \text{otherwise}
\end{cases}
\]
so that $\text{left}^\mathcal{B}_S(i)$ is true if and only if $w_i$ has some word from the set $S$ to its left, possibly with some adjective or adverb between them.

\begin{example}
We consider once again the bigraph $\mathcal{B}_1 = (G_1, D_1, w_1^\anno{POS}, A_{G_1}, A_{D_1})$  as introduced in Example~\ref{example:bestPrefix} where 
\[
w_1^\anno{POS} = (\text{the}, \text{DT}) (\text{developer}, \text{NN}) (\text{is}, \text{VB}) (\text{not}, \text{RB}) (\text{delighted}, \text{JJ}) = (w_1, p_1) \ldots (w_5, p_5)\,.
\]
The statements $\text{left}^{\mathcal{B}_1}_{ \{ \text{the, a, an} \} }(2)$ and $\text{left}^{\mathcal{B}_1}_{ \{ \text{is} \} }(5)$ are both true.
The first statement is true because $w_{i-1} = w_1 \in \{ \text{the, a, an} \}$; the second statement is true because $w_{i-2} = w_3 \in \{\text{is}\}$ and $\text{simplify}(p_{i-1}) = \text{simplify}(\text{RB}) = \text{JJ}$. 
\end{example}

Using the above prerequisites, we now describe how the gold syntactic annotation $\alpha_v$ for each vertex $v \in V_G$ can be obtained from $\mathcal{B}$. For this purpose, let $v \in V_G$ be a vertex that is aligned to at least one word, i.e. $A_G(v) \neq \emptyset$, and let $i = \text{bestPrefix}_\mathcal{B}(v, A_G(v))$. Furthermore, let
\begin{align*}
\langle \text{be} \rangle & = \{ \text{be, am, is, are, was, were, being, been} \} \\
\langle \text{have} \rangle & = \{ \text{have, has, had, having} \}
\end{align*}
be two sets containing all forms of the verbs ``be'' and ``have'', respectively. The gold syntactic annotation values $\alpha_v(k)$ for all syntactic annotation keys $k \in \mathcal{K}_\text{syn}$ can be determined independently as follows:

\begin{itemize}
\item \anno{POS}: We assign to $v$ the POS tag $\text{simplify}(p_i)$; the only exception to this rule is that when $w_i$ is a participle and has some form of ``be'' or ``have'' to its left, we treat $v$ like an actual verb:
\begin{align*}
\alpha_v(\anno{POS}) & = 
	\begin{cases}
	\text{VB} & \text{if } p_i \in \{\text{VBN, VBG}\} \wedge \text{left}^\mathcal{B}_{ \langle \text{be} \rangle \cup \langle \text{have} \rangle }(i)  \\
	\text{simplify}(p_i) & \text{otherwise.} \\
 	\end{cases}
\intertext {
\item \anno{NUMBER}: The number of $v$ can be inferred from its non-simplified POS tag:
}
{\alpha}_v(\anno{NUMBER}) & = 
	\begin{cases}
	\text{singular} & \text{if } p_i \in \{ \text{NN, NNP, FW} \} \\
	\text{plural} & \text{if } p_i \in \{ \text{NNS, NNPS} \} \\
	\text{--} & \text{otherwise.}
 	\end{cases}
\intertext{
\item \anno{VOICE}: To determine whether a vertex has passive voice, we check whether its realization is a past participle that has some form of the verb ``be'' close to its left:
}
{\alpha}_v(\anno{VOICE}) & = 
	\begin{cases}
	\text{active} & \text{if } \text{simplify}(p_i) = \text{VB} \\
	\text{passive} & \text{if } p_i = \text{VBN} \wedge \text{left}^\mathcal{B}_{\langle \text{be} \rangle}(i) = 1 \\
	\text{--} & \text{otherwise.}
 	\end{cases}	
\intertext{
\item \anno{TENSE}: To determine the tense of a vertex, we must take into account both its non-simplified POS tag and its left context:}
{\alpha}_v(\anno{TENSE}) & = 
	\begin{cases}
	\text{present} & \text{if } p_i \in \{\text{VBP, VBZ}\} \\
	\text{past} & \text{if } p_i = \text{VBD} \\
	\text{future} & \text{if } p_i = \text{VB} \wedge \text{left}^\mathcal{B}_{ \{ \text{will} \} }(i) = 1 \\
	\text{--} & \text{otherwise.}
 	\end{cases}
\intertext{
\item \anno{DENOM}: We devise two different approaches to assign a denominator to a vertex. While the first approach is purely based upon the AMR graph and the reference realization, the second one makes use of the dependency tree $D$.
For the first approach, we simply check whether the currently considered vertex represents a noun and, if so, whether some article can be found close to its left:
}
{\alpha}_v(\anno{DENOM}) & = 
	\begin{cases}
	\text{the} & \text{if } \text{simplify}(p_i) = \text{NN} \wedge \text{left}^\mathcal{B}_{ \{ \text{the} \} }(i) \\
	\text{a} & \text{if }   \text{simplify}(p_i) = \text{NN} \wedge \text{left}^\mathcal{B}_{ \{ \text{a, an} \} }(i) \\
	\text{--} & \text{otherwise.}
 	\end{cases}
\intertext{
For the second approach, let $D = (V_D, E_D, L_D, \prec_D)$. We consider $v' = A_D^{-1}(i)$, the vertex of the dependency tree that corresponds to $w_i$, and simply check whether one of its children is an article:
}
{\alpha}_v(\anno{DENOM}) & = 
	\begin{cases}
	\text{the} & \text{if } \exists v'' \in \ch[D]{v'} \colon L_D(v'') = \text{the} \\
	\text{a} & \text{if } \exists v'' \in \ch[D]{v'} \colon L_D(v'') \in \{ \text{a, an} \} \\
	\text{--} & \text{otherwise.}
 	\end{cases}
\end{align*}
\end{itemize}

An example of how gold syntactic annotations can be obtained using the above procedures can be seen in Figure~\ref{fig:bigraph-synanno}, where the gold syntactic annotations extracted from a POS-annotated version of the bigraph introduced in Example~\ref{example:bigraph} are shown. 


\begin{figure}[]
\centering
\subfloat[Graphical representation of the bigraph $\mathcal{B} = (G_1, G_2, w^\anno{POS}, A_1, A_2)$, a POS-annotated version of the bigraph introduced in Example~\ref{example:bigraph}. For $i \in \{1, 2\}$, each node $v$ of $G_i$ is inscribed with $v$\,:\,$L_i(v)$; each alignment $(u,j) \in A_i$ is represented by a dashed arrow line connecting $u$ and $w^\anno{POS}(j)$.]{
\scalebox{0.8} {
\begin{tikzpicture}
\tikzstyle{amr-node}=[shape=ellipse,draw, inner sep=0.2, minimum height=0.8cm, text height=1.5ex, text depth=.25ex]
\tikzstyle{dep-node}=[shape=ellipse,draw, inner sep=0.2, minimum height=0.8cm, minimum width=1.5cm, text height=1.5ex, text depth=.25ex]
\tikzstyle{text-node}=[text height=1.5ex, text depth=.25ex]
\tikzstyle{amr-align}=[-latex,color=red, dashed]
\tikzstyle{dep-align}=[-latex,color=red, dashed]
\tikzstyle{amr-edge}=[text height=1.5ex, text depth=.25ex, fill=white]
\tikzstyle{frame}=[draw, dotted, color=gray, inner sep=0.3cm]

	\node (PAD1) at (-8,0){};
	\node (PAD2) at (8,0){};

    \node[amr-node] (amr-want) at (0,0) {$1$\,:\,want-01};
    \node[amr-node] (amr-person) at (-3,-2) {$2$\,:\,person};
    \node[amr-node] (amr-sleep) at (4,-1.25) {$3$\,:\,sleep-01};
    \node[amr-node] (amr-develop) at (-5,-4) {$4$\,:\,develop-02};

    \path [-latex](amr-want) edge node[fill=white] {ARG0} (amr-person);
    \path [-latex](amr-want) edge node[fill=white] (amr-want-arg1) {ARG1} (amr-sleep);
    \path [-latex](amr-sleep) edge node[fill=white, pos=0.38] {ARG0} (amr-person);
    \path [-latex](amr-person) edge node[fill=white] {ARG0-of} (amr-develop);
    
    \node (amr-pad1) at (-6.5,0.25){};
    \node (amr-pad2) at (6.5,-4.25){};
    \node[frame, fit=(amr-pad1)(amr-pad2)] (amr-frame) {};
    
    \node[text-node, align=center] (text-the) at (-5.5, -5.5) {(the, DT)};
    \node[text-node] (text-developer) at (-3, -5.5) {(developer, NN)};
    \node[text-node] (text-wants) at (0, -5.5) {(wants, VBZ)};
    \node[text-node] (text-to) at (2.5, -5.5) {(to, PRT)};
    \node[text-node] (text-sleep) at (5, -5.5) {(sleep, VB)};
    
    \node (text-pad1) at (-6.5,-5.4){};
    \node (text-pad2) at (6.5,-5.6){};
    \node[frame, fit=(text-pad1)(text-pad2)] (text-frame) {};
    
    \path[amr-align] (amr-person) edge node[] {} (text-developer);
    \path[amr-align] (amr-develop) edge node[] {} (text-developer);
    \path[amr-align] (amr-want) edge node[] {} (text-wants);
    \path[amr-align] (amr-sleep) edge node[] {} (text-sleep);
    
    \node[dep-node] (dep-the) at (-3, -7) {$4$\,:\,the};
    \node[dep-node] (dep-to) at (3, -7) {$5$\,:\,to};
    \node[dep-node] (dep-developer) at (-3, -9) {$2$\,:\,developer};
    \node[dep-node] (dep-sleep) at (3, -9) {$3$\,:\,sleep};
    \node[dep-node] (dep-wants) at (0, -10.5) {$1$\,:\,wants};

    \path [-latex](dep-wants) edge node[amr-edge] {nsubj} (dep-developer);
    \path [-latex](dep-wants) edge node[amr-edge, pos=0.47] {xcomp} (dep-sleep);
    \path [-latex](dep-developer) edge node[fill=white] {det} (dep-the);
    \path [-latex](dep-sleep) edge node[fill=white] {mark} (dep-to); 
    
    \node (dep-pad1) at (-6.5,-6.75){};
    \node (dep-pad2) at (6.5,-10.75){};
    \node[frame, fit=(dep-pad1)(dep-pad2)] (dep-frame) {};
    
    \path[dep-align] (dep-the) edge node[] {} (text-the);
    \path[dep-align, bend angle=45, bend right] (dep-developer) edge node[] {} (text-developer);
    \path[dep-align] (dep-wants) edge node[] {} (text-wants);
    \path[dep-align, bend angle=25, bend right] (dep-sleep) edge node[] {} (text-sleep);
    \path[dep-align] (dep-to) edge node[] {} (text-to);   
    
    \node[] at (-7.6, -2) {$G_1$}; 
    \node[] at (-7.6, -5.5) {$w^\anno{POS}$}; 
    \node[] at (-7.6, -8.75) {$G_2$};
    \node[below = -0.1cm of dep-frame] {};
\end{tikzpicture}
}} \\[0.5cm]

\subfloat[ Gold syntactic annotation $\alpha_i$ for each vertex $i \in \{1,2,3,4\}$ of graph $G_1$ shown above]{
\begin{tikzpicture}
	\tikzstyle{frame}=[draw, dashed, inner sep=0.3cm]
    \node[align=center] (syn1) at(-4.2, 0) { 
    \begin{tabular}{c}
   
    $\alpha_{1}$ \\
    \midrule
    {$\begin{aligned}
         \anno{POS} &\mapsto \text{VB}\\
         \anno{NUMBER} &\mapsto \text{--}\\
         \anno{VOICE} &\mapsto \text{active}\\
         \anno{TENSE} &\mapsto \text{present}\\
         \anno{DENOM} &\mapsto \text{--}\\
      \end{aligned}$}\\
      \midrule
       \end{tabular} 
     }; 
     
     \node[align=center] (syn2) at(0, 0) { 
     \begin{tabular}{c}
     
     $\alpha_{2} = \alpha_4$ \\
     \midrule
     {$\begin{aligned}
  		  \anno{POS} &\mapsto \text{NN}\\
          \anno{NUMBER} &\mapsto \text{singular}\\
          \anno{VOICE} &\mapsto \text{--}\\
          \anno{TENSE} &\mapsto \text{--}\\
          \anno{DENOM} &\mapsto \text{the}\\
       \end{aligned}$} \\
       \midrule
        \end{tabular} 
      }; 
      
       \node[align=center] (syn2) at(4.2, 0) { 
       \begin{tabular}{c}
       
             $\alpha_{3}$ \\
                    \midrule
                    {$\begin{aligned}
               		  \anno{POS} &\mapsto \text{VB}\\
                       \anno{NUMBER} &\mapsto \text{--}\\
                       \anno{VOICE} &\mapsto \text{active}\ \\
                       \anno{TENSE} &\mapsto \text{--}\\
                       \anno{DENOM} &\mapsto \text{--}\\
                      \end{aligned}$}\\
                      \midrule
          \end{tabular} 
        }; 
  
\end{tikzpicture}
}
\caption{A bigraph and the gold syntactic annotations inferred from it}
\label{fig:bigraph-synanno}
\end{figure}


By extracting the correct syntactic annotation $\alpha_v$ for each $v \in V_G$ and doing so for every graph contained within our extended corpus $C_\text{ext}$, we obtain a sequence of training data that can be used to train the maximum entropy models $p_k$, $k \in \mathcal{K}_\text{syn}$ required in Section~\ref{GENERATION:SyntacticAnnotation}; the only remaining task is to specify the sequence of features used by these models.
To fulfill this task, we first define a set $\mathcal{F}$ of \emph{feature candidates} where each feature candidate is itself a sequence of features. We then automatically select the best working feature candidates using a greedy algorithm that works as follows:\footnote{Feature selection is also performed through the training algorithm itself by setting corresponding weights to zero. We nonetheless narrow down the choice of feature candidates to improve efficiency.} We start with an empty sequence of features $\mathbf{f}_0 = \varepsilon$ and check for each of the feature candidates $\mathbf{f} \in \mathcal{F}$ whether and by how much adding the contained features to $\mathbf{f}_0$ improves the number of  vertices correctly annotated by the fully trained model on a development data set. We then update $\mathbf{f}_0$ by adding to it the best performing feature candidate $\hat{\mathbf{f}}$ to obtain $\mathbf{f}_1 = \hat{\mathbf{f}} {:} \mathbf{f}_0$ and set $\mathcal{F} \gets \mathcal{F} \setminus \{ \hat{\mathbf{f}} \}$. We continue this procedure to obtain $\mathbf{f}_2, \ldots, \mathbf{f}_n$ until either $\mathcal{F} = \emptyset$ or no more feature candidate is found which improves the result and we take the resulting sequence $\mathbf{f}_n$ as the feature vector of our maximum entropy model.
Before describing how $\mathcal{F}$ is obtained, we require two auxiliary definitions.

\begin{definition}[Gold parent]\label{def:gold-parent}
Let $G = (V, E, L, \prec)$ be a rooted, acyclic graph and $v \in V \setminus \{ \text{root}(G) \}$. The \emph{gold parent of $v$}, denoted by $\widehat{\text{pa}}_G(v)$, is defined as
\[
\widehat{\text{pa}}_G(v) = \argmin_{ v' \in \pa[G]{v} } \text{dist}(\text{root}(G), v')
\]
where for all $v_1, v_2 \in V$, $\text{dist}(v_1, v_2) = 0$ if $v_1 = v_2$ and otherwise, $\text{dist}(v_1, v_2)$ denotes the number of vertices in the shortest walk starting at $v_1$ and ending at $v_2$.
\end{definition}

\begin{definition}[Empirical POS tag]\label{def:empirical-pos}
Let $l \in L_\text{C}$ be an AMR concept. The \emph{empirical POS tag of $l$}, denoted by $\overline{\text{pos}}(l)$, is defined as
\[
\overline{\text{pos}}(l) = \begin{cases} \texttt{PROP} & \text{ if } l \text{ is a PropBank frameset}  \\
\widehat{\text{pos}}(l) & \text{ otherwise}
\end{cases}
\] 
where $\widehat{\text{pos}}(l)$ denotes the POS tag observed most often for concept $l$ in a set of training data.
\end{definition}

Table~\ref{tab:syntactic-annotation-features} lists the indicator features from which $\mathcal{F}$ is derived. Most of these features are parametrized with a single vertex $v$; when computing the feature vector for some vertex $v'$, we set this parameter not only to $v'$, but also to $\widehat{\text{pa}}_G(v')$ and $\widehat{\text{pa}}_G(\widehat{\text{pa}}_G(v'))$, if they exist. In other words, we extract features not only from vertex $v'$ itself, but also from its gold parent and grandparent. We collect all so-obtained indicator features in a set $S = \{s_1, \ldots, s_m \}$, $m \in \mathbb{N}$. The set $\mathcal{F}$ of feature candidates is then derived in a one-to-one manner from the indicator features in $S$ and all pairwise combinations $s_i \circ s_j$, $1 \leq i < j \leq m$ thereof; 
the details of this composition and the conversion from indicator features to actual features can be found in Section~\ref{PRELIMINARIES:LinearClassification}.

\begin{table}
\centering
\bgroup
\def\arraystretch{1.15}
\small
\begin{tabularx}{\textwidth}{|l|X|}
\hline
\textbf{Indicator Feature} & \textbf{Value} \\
\hline
Concept$(v)$ & $L(v)$ \\
Concept$_{S}(v)$, $S \subseteq L_\text{C}$ & A flag indicating whether $L(v) \in S$ \\
Lemma$(v)$ & $L(v)$ with all PropBank sense tags removed \\
WordNetPos$(v)$ & The most likely POS tag for Lemma$(v)$ according to the \emph{use count} provided by WordNet \citep{miller1995wordnet,fellbaum1998wordnet} \\
Pos$(v)$ & The POS tag assigned to $v$, if already determined \\
Number$(v)$ & The number assigned to $v$, if already determined \\
InLabel$(v)$ & If $v \neq \text{root}(G)$, this is the label of the edge connecting $\widehat{\text{pa}}_G(v)$ and $v$; otherwise, it is set to a special value \texttt{ROOT} \\
InLabelInv$(v)$ & A flag indicating whether InLabel$(v)$ ends with $\text{-of}$ \\
InLabelArg$(v)$ & A flag indicating whether InLabel$(v)$ starts with $\text{ARG}$ \\
$\text{HasChild}_l(v)$, $l \in L_\text{C}$ & A flag indicating whether there is some $v' \in \ch[G]{v}$ with $L(v') = l$ \\
$\text{HasEdge}_l(v)$, $l \in L_\text{R}$ & A flag indicating whether there is some $v' \in V$ such that $(v, l, v') \in E$ \\
OutSize$(v)$ & $|\ch[G]{v}|$ \\
OutEmpty$(v)$ & A flag indicating whether $|\ch[G]{v}| = 0$ \\
OutLabels$(v)$ & $\{ l \in L_\text{R} \mid \exists v' \in V \colon (v, l, v') \in E \}$ \\
InLabels$(v)$ & $\{ l \in L_\text{R} \mid \exists v' \in V \colon (v', l, v) \in E \}$ \\
OutLabelsPos$(v)$ & $\{ (l,p) \in L_\text{R} \times \mathcal{V}_\anno{POS} \mid \exists v' \in V \colon (v, l, v') \in E \wedge \overline{\text{pos}}(L(v')) = p \}$ \\
InLabelsPos$(v)$ & $\{ (l,p) \in L_\text{R} \times \mathcal{V}_\anno{POS} \mid \exists v' \in V \colon (v', l, v) \in E \wedge \overline{\text{pos}}(L(v')) = p \}$ \\
Children$(v)$ & $\{ L(v') \mid v' \in \ch[G]{v} \}$ \\
Parents$(v)$ & $\{ L(v') \mid v' \in \pa[G]{v} \}$ \\
OutLabelsChildren$(v)$ & $\{ (l_r,l_c) \in L_\text{R} \times L_\text{C} \mid \exists v' \in V \colon (v, l_r, v') \in E \wedge L(v') = l_c \}$ \\
NonLinkChildren$(v)$ &  $\{ L(v') \mid v' \in \ch[G]{v} \wedge v = \widehat{\text{pa}}_G(v') \}$\\
ChildrenPos$(v)$ & $\{ \overline{\text{pos}}(L(v')) \mid v' \in \ch[G]{v} \}$ \\
Name$(v)$ & The name assigned to $v$, if $\text{name} \in \text{OutLabels}(v)$ \\
Mod$(v)$ & $\{ L(v') \mid v' \in V, (v, \text{mod}, v') \in E \}$ \\
ModPos$(v)$ & $\{ \overline{\text{pos}}(L(v')) \mid v' \in V, (v, \text{mod}, v') \in E \}$ \\
Height$(v)$ & The height of $\restr{G}{v}$, if the latter is a tree \\
Depth$(v)$ & The length of the shortest path from $\text{root}(G)$ to $v$ \\
NrOfArgs$(v)$ & $| \{ e \in E \mid \exists v' \in V, i \in \mathbb{N} \colon e = (v, \text{ARG}i, v') \} |$ \\
ArgFlags$(v)$ &  $\{ (\text{ARG}i, *(i)) \mid 1 \leq i \leq 5  \}$ where $*(i)$ is a flag indicating whether $v$ has an outgoing edge labeled $\text{ARG}i$ \\
ArgLinkFlags$(v)$ & $\{ (\text{ARG}i, *(i)) \mid 1 \leq i \leq 5  \}$ where $*(i)$ is a flag indicating whether $v$ has an outgoing edge $(v, \text{ARG}i, v')$ such that $v = \widehat{\text{pa}}_G(v')$ \\
ArgOfFlags$(v)$ & $\{ (\text{ARG}i\text{-of}, *(i)) \mid 1 \leq i \leq 5  \}$ where $*(i)$ is a flag indicating whether $v$ has an incoming edge labeled $\text{ARG}i\text{-of}$ \\
AllEdgeLabels & $\{ l \in L_\text{R} \mid \exists v_1, v_2 \in V \colon (v_1, l, v_2) \in E \}$ \\
AllCombinedLabels & $\{ (l_r,l_c) \in L_\text{R} \times L_\text{C} \mid \exists v_1, v_2 \in V \colon (v_1, l_r, v_2) \in E \wedge L(v_2) = l_c \}$ \\

\hline
\end{tabularx}
\caption{Indicator features used for modeling the probability of syntactic annotations given an AMR graph $G = (V,E,L, \prec)$. For $v \in V$ and $l \in L_\text{C}$, $\widehat{\text{pa}}_G(v)$ denotes $v$'s gold parent and  $\overline{\text{pos}}(l)$ denotes the empirical POS tag of $l$. For each indicator feature $s$, the value $s(G)$ is either explained textually or formally defined. If $s(G)$ is a singleton, delimiting brackets are omitted.}
\label{tab:syntactic-annotation-features}
\egroup
\end{table}

\subsubsection{Transitions}
\label{TRAINING:Transitions}

We now describe how the parameters required for estimating the probability distribution $P(t \mid c)$ for $t \in T_\text{AMR}$, $c \in C_\text{AMR}$ with maximum entropy models can be obtained from an extended  corpus $C_\text{ext}$ as defined in Section~\ref{TRAINING:Preparations}. To this end, we first show how each element of $C_\text{ext}$ can be turned into a sequence of training data $T = (c_1, t_1), \ldots, (c_m, t_m) \in (C_\text{AMR} \times T_\text{AMR})^*$ consisting of configurations and corresponding \emph{gold transitions}.

We again focus on one element $\mathcal{B} = (G, D, w^\anno{POS}, A_G, A_D)$ of $C_\text{ext}$. To extract the desired sequence $T$ from $\mathcal{B}$, we require two auxiliary procedures: Firstly, we need a function $\text{gold}_\mathcal{B} \colon C_\text{AMR} \setminus \ct \rightarrow T_\text{AMR}$ that maps each non-terminal configuration $c$ to the correct transition $\text{gold}_\mathcal{B}(c)$ to be applied next; we call this function an \emph{oracle}. Secondly, we require a procedure to update $\mathcal{B}$ whenever some transition $t$ is applied to $c$ in order to reflect this application on $\mathcal{B}$. We denote the result of updating the bigraph according to this procedure by $\text{update}(\mathcal{B}, c, t)$. Using these procedures, the sequence $T$ can be obtained through Algorithm~\ref{alg:training}, a simple modification of Algorithm~\ref{alg:greedy-generation} to which we refer as the \emph{training data algorithm}. At the very end of the current section, a comprehensive exemplary application of the training data algorithm and the subroutines used therein is given.

\begin{algorithm}
\DontPrintSemicolon
\KwIn{bigraph $\mathcal{B} = (G, D, w^\anno{POS}, A_G, A_D)$ from $C_\text{ext}$}
\KwOut{sequence of training data $T \in (C_\text{AMR} \times T_\text{AMR})^*$}
\SetKwBlock{Begin}{function}{end function}
\Begin($\text{trainingData} {(} \mathcal{B} {)}$)
{
	$T \gets \varepsilon$\; 
	$c \gets \cs (G)$\; \label{alg:training-line:initial}
    \While{$c \notin \ct$} 
    { \label{alg:training-line:while}
        $t^* \gets \text{gold}_\mathcal{B}(c)$\; \label{alg:training-line:gold}
        $T \gets (c,t^*)\,{:}\,T$\; \label{alg:training-line:new_td}
        $\mathcal{B} \gets \text{update}(\mathcal{B}, c, t^*)$\; \label{line:update} \label{alg:training-line:update}
        $c \gets t^*(c)$\; \label{alg:training-line:new_c}
    }
    \Return{$T$}
}
\caption{Training data algorithm}\label{alg:training}
\end{algorithm}

In the following, we first devise an algorithm to determine $\text{gold}_\mathcal{B}(c)$ and then describe the procedure required to obtain $\text{update}(\mathcal{B},c,t)$.
Given a configuration $c \in C_\text{AMR}$, we compute $\text{gold}_\mathcal{B}(c)$ by first checking for each class $\tau \in \mathcal{C}(T_\text{AMR})$ whether some instance thereof, i.e. some transition $t$ such that $\mathcal{C}(t) = \tau$, needs to be applied. As soon as a class $\tau$ is found of which an instance needs to be applied, we distinguish two cases: If $\tau$ is not parametrized, i.e. $\tau \in \{$\textsc{Keep, Delete, Swap, No-Insertion}$\}$, then $\tau$ is returned immediately. Otherwise, the actual instance of $\tau$ that needs to be applied is determined by calling yet another subroutine $\text{gold}'_\mathcal{B} \colon \mathcal{C}(T_\text{AMR}) \times C_\text{AMR} \pfun T_\text{AMR}$ that is defined such that $\text{gold}'_\mathcal{B}(\tau, c)$ always belongs to class $\tau$.\footnote{In the definition of $\text{gold}'_\mathcal{B}(\tau, c)$, we will sometimes use nondeterminism. It is therefore not a function in the strict mathematical sense; we will view it as a function nonetheless.} The only exception to this rule is that if $\tau \in \{$\textsc{Insert-Child, Insert-Between}$\}$, we also allow $\text{gold}'_\mathcal{B}(\tau,c)$ to be a $\textsc{No-Insertion}$ transition. The idea outlined above is implemented in Algorithm~\ref{alg:gold-transition}, to which we will refer as the \emph{oracle algorithm}.

\begin{algorithm}[h!]
\DontPrintSemicolon
\SetArgSty{textrm}
\KwIn{configuration $c = (G, \sigma_1{:}\sigma, \beta, \rho) \in C_\text{AMR}$ where $G = (V,E, L, \prec)$, bigraph~$\mathcal{B} = (G, D, w^\anno{POS}, A_G, A_D)$ from $C_\text{ext}$ }
\KwOut{gold transition $t \in T_\text{AMR}$}
\SetKwBlock{Begin}{function}{end function}
\Begin($\text{gold}_\mathcal{B} {(} c {)}$)
{	
	\uIf{$\sigma_1 \notin \text{dom}(\rho(\anno{DEL}))$}
	{	
		\If{$|\text{in}_G(\sigma_1)| \geq 2$} { \label{alg:oracle-line:checkdelre}
			\Return{$\text{gold}'_\mathcal{B}(\textsc{Delete-Reentrance}, c)$}
		}
		let $\pa[G]{\sigma_1} = \{ p_{\sigma_1} \}$\;
		\uIf{ $A_G(\sigma_1) = \emptyset$ }
		{ \label{alg:oracle-line:checkdel}
			\Return{\textsc{Delete}}
		}
		\uElseIf{$A_G(\sigma_1) \cap A_G(p_{\sigma_1}) \neq \emptyset$}{
			\label{alg:oracle-line:checkmerge}
		    \Return{$\text{gold}'_\mathcal{B}(\textsc{Merge}, c)$}
		}
		\uElseIf{$ A_G(p_{\sigma_1}) \neq \emptyset \wedge \forall i \in \text{span}_\mathcal{B}^1(p_{\sigma_1}) \colon \min(\text{span}_\mathcal{B}^1(\sigma_1)) \leq i \leq \max(\text{span}_\mathcal{B}^1(\sigma_1))$}
		{ \label{alg:oracle-line:checkswap}
			\Return{\textsc{Swap}}
		}
		\Else 
		{
			\Return{\textsc{Keep}} \label{alg:oracle-line:keep}
		}
	}
	
	\uElseIf{$\sigma_1 \notin \text{dom}(\rho(\anno{REAL}))$}
	{
		\Return{$\text{gold}'_\mathcal{B}(\textsc{Realize}, c)$}
	}
	\uElseIf{$\sigma_1 \notin \text{dom}(\rho(\anno{INS-DONE})) \wedge \rho(\anno{DEL})(\sigma_1) = 0$}
	{
		\Return{$\text{gold}'_\mathcal{B}(\textsc{Insert-Child}, c)$} \label{alg:oracle-line:ins-child}
	}
	\ElseIf{$\beta = \varepsilon$} 
	{
		\Return{$\text{gold}'_\mathcal{B}(\textsc{Reorder}, c)$}
	}
	\Return{$\text{gold}'_\mathcal{B}(\textsc{Insert-Between}, c)$}
}
\caption{Oracle algorithm}\label{alg:gold-transition}
\end{algorithm}

We now describe how the subroutine $\text{gold}'_\mathcal{B} \colon \mathcal{C}(T_\text{AMR}) \times C_\text{AMR} \pfun T_\text{AMR}$ is defined. For some classes $\tau \in \mathcal{C}(T_\text{AMR})$, we devise two different approaches for obtaining the best transition: one that is purely based upon the AMR graph, its realization and the alignment between them and one that additionally makes use of dependency trees. 

Let $\mathcal{B} = (G, D, w^\anno{POS}, A_G, A_D)$ be an element of $C_\text{ext}$ as above, $c = (G, \sigma_1{:}\sigma, \beta, \rho) \in C_\text{AMR}$, $G = (V_G, E_G, L_G, \prec_G)$, $D = (V_D, E_D, L_D, \prec_D)$ and $w^\anno{POS} = (w_1, p_1) \ldots (w_n, p_n)$. For $i \in [n]$, we denote $w_i$ also by $w(i)$ and $p_i$ also by $p(i)$. The required gold transitions can be obtained as follows:

\begin{itemize}
\item $\text{gold}'_\mathcal{B}(\textsc{Delete-Reentrance}, c)$: A \emph{gold incoming edge} $\hat{e} \in \text{in}_G(\sigma_1)$ for vertex $\sigma_1$ is determined; we view this edge as the only incoming edge that is not to be removed. Given $\hat{e}$, some non-gold edge $(v, l, \sigma_1) \in \text{in}(\sigma_1) \setminus \{ \hat{e} \}$ is chosen nondeterministically and the transition $\textsc{Delete-Reentrance-}(v,l)$ is returned. We are guaranteed that such an edge exists as $|\text{in}_G(\sigma_1)| \geq 2$. 

For our first approach -- which makes no use of $D$ --, we simply take the edge connecting $v$ and its gold parent $\widehat{\text{pa}}_G(v)$ (see Definition~\ref{def:gold-parent}) as the gold incoming edge $\hat{e}$. If there are multiple such edges, we choose any of them but we favor edges with non-inverted labels. We note that this approach does not even make use of $w^\anno{POS}$ or $A_G$. Therefore, $\hat{e}$ can also unambiguously be inferred from an AMR graph $G$ during test time.

For the second approach, we use $D$ to compute a set of candidates $C \subseteq V_G$ containing every parent of $\sigma_1$ for which some corresponding dependency tree vertex is also a parent of some dependency tree vertex corresponding to $\sigma_1$:
\[
C = \{ p_{\sigma_1} \in \pa[G]{\sigma_1} \mid \exists p_\text{dep} \in \pi_\mathcal{B}^1(p_{\sigma_1}), \sigma_\text{dep} \in \pi_\mathcal{B}^1(\sigma_1) \colon p_\text{dep} \in \pa[D]{\sigma_\text{dep}} \}\,.
\]
If $C$ consists of only one parent candidate $\hat{p}$ and there is exactly one edge $\hat{e}$ connecting $\hat{p}$ and $\sigma_1$, we simply take $\hat{e}$ to be the gold incoming edge. Otherwise, we determine $\hat{e}$ using the first approach, but with the additional constraint that it must originate from some vertex contained within $C$. 

\item $\text{gold}'_\mathcal{B}(\textsc{Merge}, c)$: Whenever this subroutine is called, we are guaranteed that $\sigma_1$ has exactly one parent; we denote this parent by $p_{\sigma_1}$. As the alignments $A_G(\sigma_1)$ and $A_G(p_{\sigma_1})$ are contiguous and $A_G(\sigma_1) \cap A_G(p_{\sigma_1}) \neq \emptyset$, their union $A = A_G(\sigma_1) \cup A_G(p_{\sigma_1})$ must as well be contiguous. Let  $(a_1, \ldots, a_n)$ be the  $A$-sequence induced by $<_\mathbb{N}$. The gold transition returned is \textsc{Merge-}$(\text{real},\text{pos})$ where $\text{real} = w(a_1) \ldots w(a_n)$ and $\text{pos} = \text{simplify}(p(a_1))$.

\item $\text{gold}'_\mathcal{B}(\textsc{Realize}, c)$: Let $(a_1, \ldots, a_n)$ be the  $A_G(\sigma_1)$-sequence induced by $<_\mathbb{N}$. We set $\text{real} = w(a_1) \ldots w(a_n)$ and return $\textsc{Realize-}(\text{real}, \alpha_{\sigma_1})$ where $\alpha_{\sigma_1}$ is the gold syntactic annotation for node $\sigma_1$ as derived in Section~\ref{TRAINING:SyntacticAnnotations}.

\item $\text{gold}'_\mathcal{B}(\textsc{Reorder}, c)$: We adapt the method by~\citet{pourdamghani2016generating} to obtain the \emph{gold order} among $\ch[G]{\sigma_1} \cup \{ \sigma_1 \}$. To this end, all children of $\sigma_1$ are first divided into a left and right half:
\begin{align*}
\text{left} & = 
\{ v \in \ch[G]{\sigma_1} \mid \text{med}(\text{span}_\mathcal{B}^1(v)) \leq \text{med}( A_G(\sigma_1) ) \} \\ \text{right} & = \ch[G]{\sigma_1} \setminus \text{left}
\end{align*}
where $\text{med}$ denotes the median of a set of natural numbers and $\text{med}(\emptyset) = -\infty$.
For all $S \in \{\text{left, right}\}$, let
\[\lessdot_S = \{ (v_1, v_2) \in S \times S \mid \text{med}(\text{span}_\mathcal{B}^1(v_1)) < \text{med}(\text{span}_\mathcal{B}^1(v_2)) \}\,.\]
We turn $\lessdot_S$ into a total order ${\lessdot_S}'$ on $S$ by fixing some arbitrary order among all nodes $v_1, v_2 \in S$ with $\text{med}(\text{span}_\mathcal{B}^1(v_1)) = \text{med}(\text{span}_\mathcal{B}^1(v_2))$. Let $x_S$ denote the $S$-sequence induced by ${\lessdot_S}'$. We return $\textsc{Reorder-}(x_\text{left} \cdot \sigma_1 \cdot x_\text{right})$.

\item $\text{gold}'_\mathcal{B}(\textsc{Insert-Child}, c)$: 
For the approach disregarding $D$, we restrict ourselves to left insertions and utilize a handwritten set $\Sigma_{\textsc{ic}} \subseteq \Sigma_\text{E}$ of allowed concepts for child insertions. This set consists mostly of auxiliary verbs and articles; for details, we refer to Section~\ref{IMPLEMENTATION:Packages:misc}. We require that articles can only be inserted as children of nouns whereas auxiliary verbs can only be assigned to verbs and adjectives. Let $i = \min(A_G(\sigma_1))$ and let $k \in \mathbb{N}$ be some hyperparameter. For $j = {i-1}, {i-2}, \ldots, {i-k}$ we check whether $w_j$ is an element of $\Sigma_\textsc{ic}$ and the following conditions hold: 
\[ 
( \nexists v' \in V_G \colon j \in A_G(v') ) \wedge ( \nexists j' \in \mathbb{N} \colon j < j' < i \wedge \text{simplify}(p_{j'}) = \text{simplify}(p_i) )\,.
\]
In other words, we only consider such words as candidates for \textsc{Insert-Child} transitions that are not aligned to any vertex and we demand that each such word is inserted as a child of the vertex aligned to the closest word to its right with fitting POS tag. As soon as some $j$ is found such that all of the above conditions hold, $\textsc{Insert-Child-}(\text{lem}(w_j), \mathsf{left})$ is returned where for each $e \in \Sigma_\text{E}$, $\text{lem}(e)$ denotes the base form of $e$; for example, $\text{lem}(\text{is}) = \text{be}$ and $\text{lem}(\text{houses}) = \text{house}$. If no such $j$ is found, we return $\textsc{No-Insertion}$.

For our alternative approach using the dependency tree $D$, we consider the set
\[
C = \{ v \in V_D \mid \exists v' \in \pi_\mathcal{B}^1(\sigma_1) \colon v \in \ch[D]{v'} \wedge \ch[D]{v} = \emptyset \}
\]
of dependency tree leaves that are children of some vertex corresponding to $\sigma_1$. For all $v \in C$, we note that $\pi_\mathcal{B}^2(v) = \emptyset$ means that the word at index $A_D(v)$ has no representation in the AMR graph. Therefore, we assume
\[
I = \{ i \in [n] \mid \exists v \in C \colon \pi_\mathcal{B}^2(v) = \emptyset \wedge i = A_D(v) \}
\]  
to be the set of indices of all words that need to be inserted as children of $\sigma_1$. If $I = \emptyset$, we return $\textsc{No-Insertion}$. Otherwise, let $j = \min(I)$. We return
$\textsc{Insert-Child-}(\text{lem}(w(j)),d)$ where $\text{lem}$ is defined as above and
\[
d = 
	\begin{cases}
	\mathsf{left} & \text{if } j < \min(A_G(\sigma_1)) \\
	\mathsf{right} & \text{otherwise.}
	\end{cases}
\]
For both approaches, if $\text{gold}'_\mathcal{B}(\textsc{Insert-Child}, c) \neq \textsc{No-Insertion}$, we denote by $\text{ind}_\mathcal{B}(\textsc{Insert-Child}, c)$ the index $j$ of the word which triggered the insertion.

\item $\text{gold}'_\mathcal{B}(\textsc{Insert-Between}, c)$: As $\beta \neq \varepsilon$ whenever this subroutine is called, we are guaranteed that there are $\beta_1 \in \ch[G]{\sigma_1}$ and $\beta' \in \ch[G]{\sigma_1}^*$ such that $\beta = \beta_1{:}\beta'$.

For the first approach, we again make use of a handwritten set $\Sigma_\textsc{ib} \subseteq \Sigma_\text{E}$ of allowed concepts, this time consisting mostly of adpositions, and we consider only cases where $\min(A_G(\sigma_1)) < \min(A_G(\beta_1))$. Furthermore, we require that the word to be inserted is located between the phrase corresponding to $\sigma_1$ and the phrase corresponding to $\beta_1$ in the reference realization. That means, we consider only words with indices in the range $( \max(A_G(\sigma_1)) , \min(A_G(\beta_1))$ as insertion candidates. 
From right to left, we check for each index $i$ in the above range whether $w_i$ is not aligned to any vertex (i.e. $\{ v \in V_G \mid (v,i) \in A_G \} = \emptyset$) and $w_i \in \Sigma_\textsc{ib}$. If this is the case, we return $\textsc{Insert-Between-}(w_i, \mathsf{left})$; if no such index is found, we return $\textsc{No-Insertion}$. However, as soon as we encounter some word $w_i$ that is aligned to some other child $\beta'$ of $\sigma_1$ (i.e. $\beta' \in \{ v \in \ch[G]{\sigma_1} \mid (v,i) \in A_G  \}$) while iterating over $i$, we assume that all words to the left of $w_i$ should be inserted between $\sigma_1$ and $\beta'$ rather than between $\sigma_1$ and $\beta_1$ and immediately return $\textsc{No-Insertion}$.


For our alternative approach, we use the dependency tree $D$ to align edges to corresponding insertions in advance and store these alignments in a set $A_\textsc{ib} \subseteq E \times [|w^\anno{POS}|]$. This is done as follows: For each vertex $v \in V_D$ with $\pa[D]{v} \neq \emptyset$ and $\ch[D]{v} \neq \emptyset$ that does not correspond to any vertex of $G$, i.e. $\pi_\mathcal{B}^2(v) = \emptyset$, we check whether there is some pair $(p_v, c_v) \in \pa[D]{v} \times \ch[D]{v}$ such that the AMR vertices corresponding to $p_v$ and $c_v$ are connected through some edge. In other words, we search for some edge $e = (v_1, l, v_2) \in E_G$ such that
\[
\exists (p_v, c_v) \in \pa[D]{v} \times \ch[D]{v} \colon v_1 \in \pi_\mathcal{B}^2(p_v) \wedge v_2 \in \pi_\mathcal{B}^2(c_v)\,.
\]
If such an edge is found, then we add $(e, A_D(v))$ to $A_\textsc{ib}$ and continue with the next dependency tree vertex. Otherwise, we check whether some edge $e' = (v_2, l, v_1)$ with the required property exists and, if so, add $(e', A_D(v))$ to $A_\textsc{ib}$. If this is also not the case, we extend our search radius and consider not only all parents and children of $v$, but also its grandparents and grandchildren.
At runtime, we must then simply check whether the edge $e$ connecting $\sigma_1$ and $\beta_1$ is aligned to some word index $i$ through $A_\textsc{ib}$. If this is not the case, $\textsc{No-Insertion}$ is returned; otherwise, we return $\textsc{Insert-Between-}(w(i), d)$ where
\[
d = 
	\begin{cases}
	\mathsf{left} & \text{if } i < \min(A_G(\beta_1)) \\
	\mathsf{right} & \text{otherwise.}
	\end{cases}
\]
For both approaches, if $\text{gold}'_\mathcal{B}(\textsc{Insert-Between}, c) \neq \textsc{No-Insertion}$, we denote by $\text{ind}_\mathcal{B}(\textsc{Insert-Between}, c)$ the index $i$ of the word which triggered the insertion.
\end{itemize}
This concludes our discussion of the oracle algorithm; we are now able to extract the correct transition to be applied next from a bigraph $\mathcal{B}$ of the extended corpus and a corresponding configuration $c$. As a next step, we describe how the bigraph $\mathcal{B}$ is updated after applying this gold transition.
For this purpose, let $\mathcal{B} = (G, D, w^\anno{POS}, A_G, A_D)$, $c =(G, \sigma_1{:}\sigma, \beta, \rho) \in C_\text{AMR}$, $t \in T_\text{AMR}$ and $G = (V, E, L, \prec)$. Furthermore, let $t(c) = (G', \sigma', \beta', \rho')$ where $G' =(V', E', L', \prec')$. Then 
\[
\text{update}(\mathcal{B},c,t) = (G', D, w^\anno{POS}, A_G', A_D)
\]
where depending on the class $\mathcal{C}(t)$ of the transition applied, the new alignment $A_G'$ between $G'$ and $w^\anno{POS}$ can be obtained by distinguishing the following cases:
\begin{itemize}
\item If $\mathcal{C}(t) = \textsc{Merge}$, then $\sigma_1$ must have exactly one parent $p_{\sigma_1}$ and the application of $t$ merges $\sigma_1$ and $p_{\sigma_1}$ into a single vertex. To reflect this in the alignment, we set
\[
A_G' = A_G \setminus \{ (\sigma_1, i) \mid i \in [|w^\anno{POS}|] \} \cup \{ (p_{\sigma_1}, i) \mid (\sigma_1, i) \in A_G \}\,.
\]
\item If $\mathcal{C}(t) \in \{ \textsc{Insert-Child}, \textsc{Insert-Between} \}$, then a new vertex is inserted into the graph, so $V' = V \cup \{ \tilde{\sigma} \}$ for some vertex $\tilde{\sigma} \in V_\text{ins}$. This vertex must be aligned to the word which triggered its insertion. We set
\[
A_G' = A_G \cup \{ (\tilde{\sigma}, \text{ind}_\mathcal{B}(\mathcal{C}(t), c) ) \}\,.
\]
\item If $\mathcal{C}(t) \notin \{\textsc{Merge}, \textsc{Insert-Child}, \textsc{Insert-Between} \}$, i.e. none of the above cases applies, we leave the alignment unchanged and set $A_G' = A_G$.
\end{itemize}

The procedures used by the training data algorithm are now fully specified. In order to obtain a complete sequence $T_\text{comp}$ of training data, we join together the sequences $T = \text{trainingData}(\mathcal{B})$ for each element $\mathcal{B}$ of $C_\text{ext}$. As probabilities for \textsc{Realize} and \textsc{Reorder} transitions are modeled slightly different from the rest, two final modifications must be made to this sequence $T_\text{comp}$: Firstly, each tuple $(c, \textsc{Realize-}(w,\alpha))$ is removed from $T_\text{comp}$ and the tuple $( (c,\alpha), \textsc{Realize-}(w,\alpha))$ is added to a new sequence $T_\textsc{Real}$. This is done because the probabilities of \textsc{Realize} transitions are estimated by a separate maximum entropy model $p_\textsc{Real}$ introduced in Eq.~(\ref{equ:rea4}) and in accordance with this model, we may assume the correct syntactic annotation for \textsc{Realize} transitions to be known. Secondly, we remove each pair $(c, t)$ with $\mathcal{C}(t) = \textsc{Reorder}$ from $T_\text{comp}$ and extract from it the sequences of training data required for training the maximum entropy models introduced in Eq.~(\ref{equ:reo3}). To this end, let $t = \textsc{Reorder-}(v_1, \ldots, v_n)$ and $c = (G, \sigma_1{:}\sigma, \beta, \rho)$. Then there is some $k \in [n]$ such that $\sigma_1 = v_k$. The following sets containing pairs of contexts and corresponding outputs are extracted from $(c, t)$:
\begin{align*}
S_* & = \{ (c, v_i \lessdot \sigma_1) \mid 1 \leq i < k \} \cup \{ (c, \sigma_1 \lessdot v_i) \mid k < i \leq n \} \\
S_l & = \{ ((c, v_i \lessdot \sigma_1, v_j \lessdot \sigma_1), v_i \lessdot v_j) \mid 1 \leq i < j < k \} \\
S_r & = \{ ((c, \sigma_1 \lessdot v_i, \sigma_1 \lessdot v_j), v_i \lessdot v_j) \mid k < i < j \leq n \} 
\end{align*}
For $i \in \{ *, l, r \}$, the sets $S_i$ extracted from all tuples in $T_\text{comp}$ of the above form are collected and joined to a new sequence $T_i$; this sequence is then used to train the maximum entropy model $p_i$ introduced in Eq.~(\ref{equ:reo3}). Analogously, the sequence $T_\textsc{Real}$ is used to train $p_\textsc{Real}$. For the maximum entropy model $p_\text{TS}$ introduced in Eq.~(\ref{equ:p_ts}), which handles all remaining transitions, the tuples remaining in $T_\text{comp}$ are used as training data.

\begin{table}[t!]
\centering
\bgroup
\def\arraystretch{1.15}
\small
\begin{tabularx}{\textwidth}{|l|X|}
\hline
\textbf{Indicator Feature} & \textbf{Value} \\
\hline
$\text{Rho}_k(v)$, $k \in \mathcal{K}$ & $\rho(k)(v)$ \\
RealizationLemma$(v)$ & The base form of $\rho(\anno{REAL})(v)$ \\
RelativePosition$(v)$ & If $v \prec p_v$ and $\rho(\anno{DEL})(p_v) = 0$, this is set to ``$\textsf{left}$''. Otherwise, if ${p_v \prec v}$ and $\rho(\anno{DEL})(p_v) = 0$, this is set to ``$\textsf{right}$''. If none of the above holds, this feature is set to ``$\textsf{del}$''. \\
$\text{OutLabels}_S(v)$, $S \subseteq L_\text{R}$ & A flag indicating whether $\text{OutLabels}(v) \subseteq S$ \\

SameSideSize$(v)$ & $|\{ v' \in V \mid p_v = p_{v'} \wedge (v \prec p_v \Leftrightarrow v' \prec p_v ) \}|$ \\
SameSideLabels$(v)$ & $\{ l \in L_\text{R} \mid \exists v' \in V \colon (p_v, l, v') \in E \wedge (v \prec p_v \Leftrightarrow v' \prec p_v ) \}$ \\
SameSideLabelsPos$(v)$ & $\{ (l,p) \in L_\text{R} \times \mathcal{V}_\anno{POS} \mid \exists v' \in V \colon (p_v, l, v') \in E \wedge \overline{\text{pos}}(L(v')) = p \wedge ({v \prec p_v} \Leftrightarrow {v' \prec p_v}) \}$ \\
SameSidePos$(v)$ & $\{ \overline{\text{pos}}(L(v')) \mid v' \in V \wedge p_v = p_{v'} \wedge (v \prec p_v \Leftrightarrow v' \prec p_v ) \}$ \\

Mergeable$(v)$ & A flag indicating whether some \textsc{Merge} transition has been applied to any vertex with the same concept and parent concept as $v$ during training \\

ComplexPos$(v)$ & For $\rho(\anno{POS})(v) \notin \{ \text{NN, VB} \}$, this is equal to $\rho(\anno{POS})(v)$. For nouns, the value of $\rho(\anno{NUMBER})(v)$ is added and for verbs, this feature is a composition of $\rho(\anno{TENSE})(v)$, $\rho(\anno{VOICE})(v)$, $\text{HasChild}_l(v)$ for all grammatical mood indicators $l$ and the most likely grammatical number $n \in \mathcal{V}_\anno{NUMBER}$ for the first child of $v$ connected through an edge with label $\text{ARG}i$, $i \in \mathbb{N}$, if such a child exists. \\

\hline
\end{tabularx}
\caption{Additional indicator features used for modeling the probabilities of transitions $P(t \mid c)$ where $c = (G, \sigma, \beta, \rho)$ with $G = (V,E,L, \prec)$. For $v \in V$ and $l \in L_\text{C}$, $p_v$ denotes the parent of $v$ if $|\pa[G]{v}| = 1$ and  $\overline{\text{pos}}(l)$ denotes the empirical POS tag of $l$ (see Definition~\ref{def:empirical-pos}). For each indicator feature $s$, the value $s(G)$ is either explained textually or formally defined. If $s(G)$ is a singleton, delimiting brackets are omitted.}
\label{tab:transition-features}
\egroup
\end{table}

To train all of the above maximum entropy models, we proceed exactly the same as for the syntactic annotation models (see Section~\ref{TRAINING:SyntacticAnnotations}). That is, we specify a set of indicator features from which we extract feature candidates that are then greedily composed to a final feature sequence with which the model is trained. As indicator features, we use the same features as for our syntactic annotation models (see Table~\ref{tab:syntactic-annotation-features}) as well as some additional ones. These additional indicator features can be found in Table~\ref{tab:transition-features}; all of them are parametrized with some vertex $v$. 
It is important to note that both the relevance and the definiteness of all our features depends heavily on the transitions whose probability is to be obtained. For instance, we may be interested in properties of both the node $\sigma_1$ on top of the node buffer and its parent when considering \textsc{Merge} transitions, whereas for \textsc{Insert-Between} transitions, properties of $\sigma_1$ and the node $\beta_1$ on top of the child buffer are of relevance. Furthermore, available context information varies due to the order in which transitions are applied. For example, the POS tag assigned to a vertex is only known \emph{after} its realization has been determined; it can therefore only be used as an indicator feature for transitions applied to it after a \textsc{Realize} transition. To handle both problems, we use varying sets of parameters for each parametrized indicator feature, depending on the considered transition; as is done by \citet{wang2015transition}, we also set each indicator feature to a special value $\textsf{NONE}$ whenever it is not relevant or not properly defined in the current context. The actual list of relevant features for each class of transitions $\tau \in \mathcal{C}(T_\text{AMR})$ can be found in the implementation (see Section~\ref{IMPLEMENTATION:Packages:ml}).

We are now able to train all maximum entropy models required to estimate $P(t\,{\mid}\,c)$, but we make one final modification to the training procedure: To compensate for errors made by our model $p_\text{TS}$ in an early stage of processing a node, we carry out the training procedure twice. In a first iteration, we train all models exactly as described above. In a second iteration, we slightly modify Algorithm~\ref{alg:training}: Whenever the transition to be applied next is contained within the set $T_\text{restr}$, we replace the call to $\text{gold}_\mathcal{B}(c)$ in line~\ref{alg:training-line:gold} with
\[
t^* \gets \argmax_{t \in T_\text{restr} \colon c \in \text{dom}(t) } P(t \mid c)
\] 
where $P$ is estimated by the model trained in the first iteration. In other words, we replace gold transitions from $T_\text{restr}$ with the actual output of our pretrained model. We then fuse the so-obtained training data sequence with the sequence obtained in the first run and retrain all maximum entropy models using this combined sequence.

We conclude this section with a comprehensive exemplary application of the training data algorithm; this application also includes several runs of the oracle algorithm. As this requires frequent switching between both algorithms, we abbreviate each line $l$ of an algorithm $a$ by ($a{:}l$); for example, (6:3) refers to the third line of Algorithm~6.    

\begin{example}
\label{example:training}
We consider a POS-annotated and lowercased version of the bigraph $\mathcal{B}_1$ introduced in Example~\ref{example:bigraph}. For reasons of consistency with the notation used throughout this section, we  additionally rename its components and obtain the bigraph $\mathcal{B} = (G, D, w^\anno{POS}, A_G, A_D)$ with $G = (V_G, E_G, L_G, \prec_G)$ and $D = (V_D, E_D, L_D, \prec_D)$ shown in Figure~\ref{fig:bigraph-pos}. 
We walk through Algorithm~\ref{alg:training} with $\mathcal{B}$ as an input step by step and show how the set $\text{trainingData}(\mathcal{B})$ is obtained. 

\begin{figure}[]
\centering
\scalebox{0.8} {
\begin{tikzpicture}
\tikzstyle{amr-node}=[shape=ellipse,draw, inner sep=0.2, minimum height=0.8cm, text height=1.5ex, text depth=.25ex]
\tikzstyle{dep-node}=[shape=ellipse,draw, inner sep=0.2, minimum height=0.8cm, minimum width=1.5cm, text height=1.5ex, text depth=.25ex]
\tikzstyle{text-node}=[text height=1.5ex, text depth=.25ex]
\tikzstyle{amr-align}=[-latex,color=red, dashed]
\tikzstyle{dep-align}=[-latex,color=red, dashed]
\tikzstyle{amr-edge}=[text height=1.5ex, text depth=.25ex, fill=white]
\tikzstyle{frame}=[draw, dotted, color=gray, inner sep=0.3cm]

	\node (PAD1) at (-8,0){};
	\node (PAD2) at (8,0){};

    \node[amr-node] (amr-want) at (0,0) {$v_1$\,:\,want-01};
    \node[amr-node] (amr-person) at (-3,-2) {$v_2$\,:\,person};
    \node[amr-node] (amr-sleep) at (4,-1.25) {$v_3$\,:\,sleep-01};
    \node[amr-node] (amr-develop) at (-5,-4) {$v_4$\,:\,develop-02};

    \path [-latex](amr-want) edge node[fill=white] {ARG0} (amr-person);
    \path [-latex](amr-want) edge node[fill=white] (amr-want-arg1) {ARG1} (amr-sleep);
    \path [-latex](amr-sleep) edge node[fill=white, pos=0.38] {ARG0} (amr-person);
    \path [-latex](amr-person) edge node[fill=white] {ARG0-of} (amr-develop);
    
    \node (amr-pad1) at (-6.5,0.25){};
    \node (amr-pad2) at (6.5,-4.25){};
    \node[frame, fit=(amr-pad1)(amr-pad2)] (amr-frame) {};
    
    \node[text-node, align=center] (text-the) at (-5.5, -5.5) {(the, DT)};
    \node[text-node] (text-developer) at (-3, -5.5) {(developer, NN)};
    \node[text-node] (text-wants) at (0, -5.5) {(wants, VBZ)};
    \node[text-node] (text-to) at (2.5, -5.5) {(to, PRT)};
    \node[text-node] (text-sleep) at (5, -5.5) {(sleep, VB)};
    
    \node (text-pad1) at (-6.5,-5.4){};
    \node (text-pad2) at (6.5,-5.6){};
    \node[frame, fit=(text-pad1)(text-pad2)] (text-frame) {};
    
    \path[amr-align] (amr-person) edge node[] {} (text-developer);
    \path[amr-align] (amr-develop) edge node[] {} (text-developer);
    \path[amr-align] (amr-want) edge node[] {} (text-wants);
    \path[amr-align] (amr-sleep) edge node[] {} (text-sleep);
    
    \node[dep-node] (dep-the) at (-3, -7) {$d_4$\,:\,the};
    \node[dep-node] (dep-to) at (3, -7) {$d_5$\,:\,to};
    \node[dep-node] (dep-developer) at (-3, -9) {$d_2$\,:\,developer};
    \node[dep-node] (dep-sleep) at (3, -9) {$d_3$\,:\,sleep};
    \node[dep-node] (dep-wants) at (0, -10.5) {$d_1$\,:\,wants};

    \path [-latex](dep-wants) edge node[amr-edge] {nsubj} (dep-developer);
    \path [-latex](dep-wants) edge node[amr-edge, pos=0.47] {xcomp} (dep-sleep);
    \path [-latex](dep-developer) edge node[fill=white] {det} (dep-the);
    \path [-latex](dep-sleep) edge node[fill=white] {mark} (dep-to); 
    
    \node (dep-pad1) at (-6.5,-6.75){};
    \node (dep-pad2) at (6.5,-10.75){};
    \node[frame, fit=(dep-pad1)(dep-pad2)] (dep-frame) {};
    
    \path[dep-align] (dep-the) edge node[] {} (text-the);
    \path[dep-align, bend angle=45, bend right] (dep-developer) edge node[] {} (text-developer);
    \path[dep-align] (dep-wants) edge node[] {} (text-wants);
    \path[dep-align, bend angle=25, bend right] (dep-sleep) edge node[] {} (text-sleep);
    \path[dep-align] (dep-to) edge node[] {} (text-to);   
    
    \node[] at (-7.6, -2) {$G$}; 
    \node[] at (-7.6, -5.5) {$w^\anno{POS}$}; 
    \node[] at (-7.6, -8.75) {$D$};         
\end{tikzpicture}
}
\caption{Graphical representation of the bigraph $\mathcal{B} = (G, D, w^\anno{POS}, A_G, A_D)$ introduced in Example~\ref{example:training}. For $i \in \{G, D\}$, each node $v \in V_i$ is inscribed with $v$\,:\,$L_i(v)$; each alignment $(u,j) \in A_i$ is represented by a dashed arrow line connecting $u$ and $w^\anno{POS}(j)$.}
\label{fig:bigraph-pos}
\end{figure}

The first step of the training data algorithm is to initialize $T = \varepsilon$ and to compute
\[
\cs(G) = (G, (v_4,v_2,v_3,v_1), \varepsilon, \rho) \text{ where } \rho = \{ (k, \emptyset) \mid k \in \mathcal{K} \}
\]
which is stored in a variable $c$ (\ref{alg:training}:\ref{alg:training-line:initial}). As $c$ is not a terminal state, the algorithm calls routine $\text{gold}_\mathcal{B}(c)$ to obtain the gold transition to be applied next. In this subroutine, it is first determined that node $v_4$ has only one parent and thus, no \textsc{Delete-Reentrance} transition needs to be applied (\ref{alg:gold-transition}:\ref{alg:oracle-line:checkdelre}). Also, as $v_4$ is aligned to some word, it must not be deleted (\ref{alg:gold-transition}:\ref{alg:oracle-line:checkdel}). It is then tested whether $v_4$ and its parent node $v_2$ have a common realization (\ref{alg:gold-transition}:\ref{alg:oracle-line:checkmerge}). As this is the case, the gold transition to be applied next belongs to the class \textsc{Merge} and as $\text{gold}'_\mathcal{B}(\textsc{Merge}, c) = \textsc{Merge-}(\text{developer}, \text{NN})$, the value returned by $\text{gold}_\mathcal{B}(c)$ is likewise $t^* =\textsc{Merge-}(\text{developer}, \text{NN})$. The training tuple $(c, t^*)$ is appended to $T$ (\ref{alg:training}:\ref{alg:training-line:new_td}), $\mathcal{B}$ is updated by removing all alignments involving $v_4$ (\ref{alg:training}:\ref{alg:training-line:update}) and $c$ is updated by applying $t^*$ (\ref{alg:training}:\ref{alg:training-line:new_c}), resulting in the new configuration
\[
c \gets (G_1, (v_2,v_3,v_1), \varepsilon, \rho_1)
\]
where $\rho_1 = \rho[\anno{POS}(v_2) \mapsto \text{NN}, \anno{INIT-CONCEPT}(v_2) \mapsto \text{person}]$ and $G_1$ is shown in Figure~\ref{fig:g1}. 

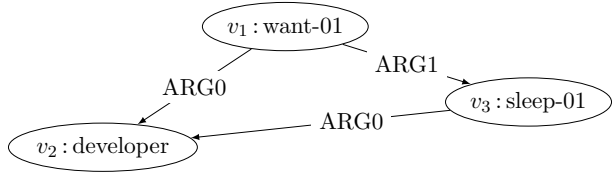
\begin{figure}[]
\centering
\scalebox{0.8} {
\begin{tikzpicture}
\tikzstyle{amr-node}=[shape=ellipse,draw, inner sep=0.2, minimum height=0.8cm, text height=1.5ex, text depth=.25ex]

    \node[amr-node] (amr-want) at (0,0) {$v_1$\,:\,want-01};
    \node[amr-node] (amr-person) at (-3,-2) {$v_2$\,:\,developer};
    \node[amr-node] (amr-sleep) at (4,-1.25) {$v_3$\,:\,sleep-01};

    \path [-latex](amr-want) edge node[fill=white] {ARG0} (amr-person);
    \path [-latex](amr-want) edge node[fill=white] (amr-want-arg1) {ARG1} (amr-sleep);
    \path [-latex](amr-sleep) edge node[fill=white, pos=0.38] {ARG0} (amr-person);
\end{tikzpicture}
}
\caption{Graphical representation of the AMR graph $G_1 = (V_{G_1}, E_{G_1}, L_{G_1}, \prec_{G_1})$. Each node $v \in V_{G_1}$ is inscribed with $v$\,:\,$L_{G_1}(v)$.}
\label{fig:g1}
\end{figure}

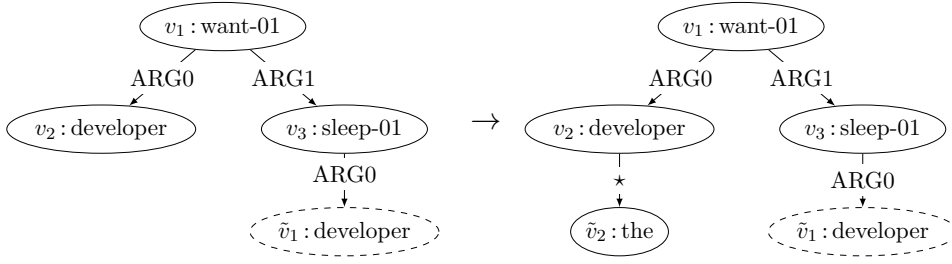
\begin{figure}[]
\centering
\scalebox{0.8} {
\begin{tikzpicture}[baseline=(current bounding box.center)]
\tikzstyle{amr-node}=[shape=ellipse,draw, inner sep=0.2, minimum height=0.8cm, text height=1.5ex, text depth=.25ex]

    \node[amr-node] (amr-want) at (0,0) {$v_1$\,:\,want-01};
    \node[amr-node] (amr-person) at (-2,-1.7) {$v_2$\,:\,developer};
    \node[amr-node] (amr-sleep) at (2,-1.7) {$v_3$\,:\,sleep-01};
    \node[amr-node, dashed] (amr-person-clone) at (2,-3.4) {$\tilde{v}_1$\,:\,developer};

    \path [-latex](amr-want) edge node[fill=white] {ARG0} (amr-person);
    \path [-latex](amr-want) edge node[fill=white] (amr-want-arg1) {ARG1} (amr-sleep);
    \path [-latex](amr-sleep) edge node[fill=white, pos=0.38] {ARG0} (amr-person-clone);
\end{tikzpicture}
\quad{\Large $\rightarrow$}\quad
\begin{tikzpicture}[baseline=(current bounding box.center)]
\tikzstyle{amr-node}=[shape=ellipse,draw, inner sep=0.2, minimum height=0.8cm, text height=1.5ex, text depth=.25ex]

    \node[amr-node] (amr-want) at (0,0) {$v_1$\,:\,want-01};
    \node[amr-node] (amr-person) at (-2,-1.7) {$v_2$\,:\,developer};
    \node[amr-node] (amr-the) at (-2,-3.4) {$\tilde{v}_2$\,:\,the};
    \node[amr-node] (amr-sleep) at (2,-1.7) {$v_3$\,:\,sleep-01};
    \node[amr-node, dashed] (amr-person-clone) at (2,-3.4) {$\tilde{v}_1$\,:\,developer};

    \path [-latex](amr-want) edge node[fill=white] {ARG0} (amr-person);
    \path [-latex](amr-person) edge node[fill=white] {$\star$} (amr-the);
    \path [-latex](amr-want) edge node[fill=white] (amr-want-arg1) {ARG1} (amr-sleep);
    \path [-latex](amr-sleep) edge node[fill=white] {ARG0} (amr-person-clone);
\end{tikzpicture}
}
\caption{Graphical representation of the AMR graph $G_2 = (V_{G_2}, E_{G_2}, L_{G_2}, \prec_{G_2})$ and the graph $G_3 = (V_{G_3}, E_{G_3}, L_{G_3}, \prec_{G_3})$ obtained from $G_2$ through a \textsc{Insert-Child}-$(\text{the}, \mathsf{left})$ transition. For $i \in \{2,3\}$, each node $v \in V_{G_i}$ is inscribed with $v$\,:\,$L_{G_i}(v)$.}
\label{fig:g2}
\end{figure}

As $c$ is still no terminal configuration, the next transition is determined by calling $\text{gold}_\mathcal{B}(c)$. Because $v_2$ has two parent nodes, $v_1$ and $v_3$, a \textsc{Delete-Reentrance} transition needs to be applied (\ref{alg:gold-transition}:\ref{alg:oracle-line:checkdelre}). For both the text-based and the dependency-tree-based approach, $\text{gold}'_\mathcal{B}(\textsc{Delete-Reentrance}, c)$ returns $\textsc{Delete-Reentrance-}(v_3, \text{ARG0})$, indicating that $e = (v_3, \text{ARG0}, v_2)$ needs to be removed from $E_{G_1}$. For the text-based approach, this is the case because the path from $v_3$ to $\text{root}(G_1)$ is longer than the path from $v_1$, making $v_1$ the gold parent of $v_2$ (see Definition~\ref{def:gold-parent}). For the approach using $D$, the reason is that $d_2$, the dependency tree vertex corresponding to $v_2$, is a child of $d_1$ (which corresponds to $v_1$), but not a child of $d_3$ (which corresponds to $v_3$).
After $t^* = \textsc{Delete-Reentrance-}(v_3, \text{ARG0})$ is returned, $(c, t^*)$ is added to the sequence $T$ of training data (\ref{alg:training}:\ref{alg:training-line:new_td}), $\mathcal{B}$ is updated (\ref{alg:training}:\ref{alg:training-line:update}) and by application of $t^*$ (\ref{alg:training}:\ref{alg:training-line:new_c}), the new configuration
\[
c \gets (G_2, (v_2, \tilde{v}_1, v_3, v_1), \varepsilon, \rho_2)
\]
is obtained where $\rho_2 = \rho_1[ \anno{LINK}(\tilde{v}_1) = v_2]$ and $G_2$ is shown in Figure~\ref{fig:g2} on the left.

In the next iteration, neither \textsc{Delete-Reentrance} nor $\textsc{Delete}$ transitions are applicable for the same reasons as in the very first iteration. There is no need for a \textsc{Merge} transition as $v_2$ and $v_1$ do not have a common realization (\ref{alg:gold-transition}:\ref{alg:oracle-line:checkmerge}). No \textsc{Swap} is required because no word aligned to $v_1$ is between two words belonging to the span of $v_2$ (\ref{alg:gold-transition}:\ref{alg:oracle-line:checkswap}). The oracle algorithm therefore returns $t^* = \textsc{Keep}$ (\ref{alg:gold-transition}:\ref{alg:oracle-line:keep}). Again, $(c,t^*)$ is added to $T$, the bigraph is updated and $t^*$ is applied whereby the new configuration
\[
c \gets (G_2, (v_2, \tilde{v}_1, v_3, v_1), \varepsilon, \rho_3)
\] 
with $\rho_3 = \rho_2[\anno{DEL}(v_2) \mapsto 0]$ is obtained; as \textsc{Keep} only modifies the \anno{DEL} flag, this configuration is almost identical to the previous one.

At its next call, the oracle algorithm returns $t^* = \textsc{Realize-}(\text{developer}, \sigma_{v_2})$ where in accordance with Figure~\ref{fig:bigraph-synanno} (Section~\ref{TRAINING:SyntacticAnnotations}),
\[
\sigma_{v_2} = \{ (\anno{POS}, \text{NN} ), (\anno{DENOM}, \text{the}), ( \anno{TENSE}, \text{--} ), ( \anno{NUMBER}, \text{singular} ), ( \anno{VOICE}, \text{--} ) \} 
\] 
is the gold syntactic annotation for $v_2$. The tuple $(c, t^*)$ is added to $T$, $\mathcal{B}$ is updated and $t^*$ is applied, resulting in the configuration
\[
c \gets (G_2, (v_2, \tilde{v}_1, v_3, v_1), \varepsilon, \rho_4)
\]
where $\rho_4$ is obtained from $\rho_3[ \anno{REAL}(v_2) \mapsto \text{developer}]$ by setting $\rho_4(k)(\sigma_1) = \sigma_{v_2}(k)$ for all $k \in \mathcal{K}_\text{syn}$. 
Yet another call of the oracle algorithm returns $t^* = \textsc{Insert-Child-}(\text{the}, \mathsf{left})$, regardless of which approach for $\text{gold}'_\mathcal{B}(\textsc{Insert-Child}, c)$ is chosen (\ref{alg:gold-transition}:\ref{alg:oracle-line:ins-child}). For the text-based approach, this is the case because $w_1$ (``the'') is not aligned to any vertex and occurs directly left of $w_2$ (``developer''), the first word aligned to $v_2$ in the reference realization. For the approach using $D$, the sets 
\begin{align*}
C & = \{ v \in V_D \mid \exists v' \in \pi_\mathcal{B}^1(v_2) \colon v \in \ch[D]{v'} \} = \{ d_4 \} \\
I & = \{ i \in [n] \mid \exists v \in C \colon \pi_\mathcal{B}^2(v) = \emptyset \wedge i = A_D(v) \} = \{ 1 \}
\end{align*}
are computed and $t^* = \textsc{Insert-Child-}(\text{lem}(w(j)), d)$ is returned where $j = \min(I) = 1$, $\text{lem}(w(1)) = \text{lem}(\text{the}) = \text{the}$ and $d = \textsf{left}$ as $1 < \min(A_G(v_2)) = 2$. 

 As before, we update $T$ and $\mathcal{B}$ and apply $t^*$ to obtain
\[
c \gets (G_3, (\tilde{v}_2, v_2, \tilde{v}_1, v_3, v_1), \varepsilon, \rho_5)
\]
where $\rho_5 = \rho_4[\anno{DEL}(\tilde{v}_2) \mapsto 0, \anno{INS-DONE}(\tilde{v}_2) = 1]$ and $G_3$ is shown in Figure~\ref{fig:g2} on the right. We leave further study of the remaining steps to the reader, but we provide in Table~\ref{tab:gold-transitions} a list of all gold transitions returned by the oracle algorithm in subsequent calls, assuming that in each call of $\text{gold}'_\mathcal{B}$, the approach which makes no use of the dependency tree $D$ is chosen to obtain the gold transition whenever two alternative approaches are defined.
\end{example}

\begin{table}
\centering
\bgroup
\def\arraystretch{1.5}
\small
\begin{tabularx}{\textwidth}{|r|r|Z|}
\hline
\multicolumn{1}{|c|}{$\sigma$} & \multicolumn{1}{c|}{$\beta$} & \textbf{Gold Transition} \\
\hline
$\tilde{v}_2:(v_2, \tilde{v}_1, v_3, v_1)$ & $\varepsilon$ & \textsc{Realize}-$(\text{the}, \sigma_{ \tilde{v}_2 })$ where $\sigma_{ \tilde{v}_2 } = \{ (\anno{POS}, \text{DT} )$, $(\anno{DENOM}, \text{--})$, $( \anno{TENSE}, \text{--} )$, $( \anno{NUMBER}, \text{--} )$, $( \anno{VOICE}, \text{--} ) \}$  \\
$\tilde{v}_2:(v_2, \tilde{v}_1, v_3, v_1)$ & $\varepsilon$ & \textsc{Reorder}-$(\tilde{v}_2)$ \\
$v_2:(\tilde{v}_1, v_3, v_1)$ & $\varepsilon$ & \textsc{No-Insertion} \\
$v_2:(\tilde{v}_1, v_3, v_1)$ & $\varepsilon$ & \textsc{Reorder}-$(\tilde{v}_2, v_2)$ \\
$v_2:(\tilde{v}_1, v_3, v_1)$ & $\tilde{v}_2$ & \textsc{No-Insertion} \\
$\tilde{v}_1 : (v_3, v_1)$ & $\varepsilon$ & \textsc{Delete} \\
$\tilde{v}_1 : (v_3, v_1)$ & $\varepsilon$ & \textsc{Reorder}-$(\tilde{v}_1)$ \\
$v_3:(v_1)$ & $\varepsilon$ & \textsc{Keep} \\
$v_3:(v_1)$ & $\varepsilon$ & \textsc{Realize}-$(\text{sleep}, \sigma_ {v_3})$  where $\sigma_{ v_3 } = \{ (\anno{POS}, \text{VB} )$, $(\anno{DENOM}, \text{--})$, $( \anno{TENSE}, \text{--} )$, $( \anno{NUMBER}, \text{--} )$, $( \anno{VOICE}, \text{active} ) \}$ \\
$v_3:(v_1)$ & $\varepsilon$ & \textsc{No-Insertion} \\
$v_3:(v_1)$ & $\varepsilon$ & \textsc{Reorder}-$(\tilde{v}_1, v_3)$ \\
$v_3:(v_1)$ & $\tilde{v}_1$ & \textsc{No-Insertion} \\
$v_1$ & $\varepsilon$ & \textsc{Keep} \\
$v_1$ & $\varepsilon$ & \textsc{Realize}-$(\text{wants}, \sigma_ {v_1})$ where $\sigma_{ v_1 } = \{ (\anno{POS}, \text{VB} )$, $(\anno{DENOM}, \text{--})$, $( \anno{TENSE}, \text{present} )$, $( \anno{NUMBER}, \text{--} )$, $( \anno{VOICE}, \text{active} ) \}$ \\
$v_1$ & $\varepsilon$ & \textsc{No-Insertion} \\
$v_1$ & $\varepsilon$ & \textsc{Reorder}-$(v_2, v_1, v_3)$ \\
$v_1$ & $v_2 : (v_3)$ & \textsc{No-Insertion} \\
$v_1$ & $v_3$ & \textsc{Insert-Between}-$(\text{to}, \mathsf{left})$ \\
$\varepsilon$ & $\varepsilon$ & -- \\
\hline
\end{tabularx}
\caption{Gold transitions returned by the oracle algorithm when processing the configuration $c = (G_3, (\tilde{v}_2,v_2, \tilde{v}_1, v_3, v_1), \varepsilon, \rho_5)$. The contents of the node buffer $\sigma$ and the child buffer $\beta$ before application of each transition are specified.}
\label{tab:gold-transitions}
\egroup
\end{table}


\subsection{Postprocessing}
\label{GENERATION:Postprocessing}

To further improve the quality of the realizations produced by our generator, we carry out several postprocessing steps. For doing so, we make use of both the actual realization $\tilde{w} = \text{generate}(G)$ obtained from the input AMR graph $G$ and the final configuration from which this realization is inferred. While there may be several more useful postprocessing steps, we restrict ourselves here to revising inserted articles, adding punctuation and removing duplicate words from the realization.

In the following, let $\hat{c} = (\hat{G}, \varepsilon, \varepsilon, \hat{\rho})$ with $\hat{G} = (\hat{V}, \hat{E}, \hat{L}, \hat{\prec})$ be the final configuration obtained in line~\ref{alg:gen-line:genpartial} of Algorithm~\ref{alg:generation} for input $G$.
As a first postprocessing step, we revise all inserted articles and check whether further articles need to be inserted. It makes sense to perform this revision as articles are added through \textsc{Insert-Child} transitions; at the time these transitions are applied to a node, its context (i.e. the words to its left and right in the final realization) is generally still unknown. We therefore simply check for each $v \in \hat{V}$ with $\hat{\rho}(\anno{POS})(v) = \text{NN}$ whether removing or inserting an article improves the score assigned to $f_\text{AMR}(\hat{c})$ through our language model.
To this end, we first remove from $\hat{G}$ each child of $v$ whose label is an element of the set $\langle \text{art} \rangle = \{ \text{a}, \text{an}, \text{the} \}$. We then compute a linear combination of the language model score and the syntactic annotation probabilities of the so-obtained graph $\hat{G}'$ and compare this score with the scores of the graphs obtained from  $\hat{G}'$ by inserting a new vertex $\tilde{v}$ with some realization from the set $\langle \text{art} \rangle$ as the leftmost child of $v$. From all of these graphs, we choose the one with the highest score and update the final configuration $\hat{c}$ accordingly.

Since all punctuation marks are removed from the AMR corpus during preparation in Section~\ref{TRAINING:Preparations}, our generator does not learn to insert them. To fix this problem, we use a rather simple, non-probabilistic approach for which we consider the set
\[
\hat{R} = 
\begin{cases}
	\ch[\hat{G}]{\text{root}(\hat{G})} & \text{if } \hat{L}(\text{root}(\hat{G})) = \text{multi-sentence} \\
	\{ \text{root}(\hat{G}) \} & \text{otherwise}
\end{cases}
\]
that, in most cases, just contains the root of $\hat{G}$. However, some AMR graphs encode not just one, but multiple sentences; this is indicated through a special concept ``multi-sentence'' for the root node. Therefore, whenever the root of $\hat{G}$ is labeled ``multi-sentence'', we process the subgraphs $\restr{\hat{G}}{v}$ for all $v \in \ch[\hat{G}]{\text{root}(\hat{G})}$ as if they were separate graphs.
For every vertex $v \in \hat{R}$, we define two predicates 
\begin{align*}
	\phi_v(\texttt{?}) & = \exists v' \in \ch[\hat{G}]{v} \colon \hat{L}(v') \in \{ \text{interrogative, amr-unknown} \} \\
	\phi_v(\texttt{,}) & = v \neq \text{root}(\hat{G}) \wedge \exists v' \in \ch[\hat{G}]{\text{root}(\hat{G})} \colon v \, \hat{\prec}\, v' 
\end{align*}
from which we infer the punctuation mark for the subgraph $\restr{\hat{G}}{v}$ as follows:
\begin{align*}
\text{punc}(v) = 
	\begin{cases} 
	\text{?} & \text{if } \phi_v(\texttt{?}) \\
	\text{,} & \text{if } \neg\phi_v(\texttt{?}) \wedge \phi_v(\texttt{,}) \\
	\text{.} & \text{if } \neg\phi_v(\texttt{?}) \wedge \neg\phi_v(\texttt{,}) \wedge |\hat{V}| \geq 5 \\
	\varepsilon & \text{otherwise.} \\ 
	\end{cases}
\end{align*}
In other words, we assign to each subgraph $\restr{\hat{G}}{v}$ the punctuation mark ``?'' if $v$ has a child labeled ``interrogative'' or ``amr-unknown'' as these are the concepts used by AMR to indicate questions. We assign the punctuation mark ``,'' if $\restr{\hat{G}}{v}$ does not encode a question and its span does not contain the rightmost word of the generated sentence. If none of the above conditions holds and $\hat{G}$ has at least five vertices, the punctuation mark ``.'' is assigned to it. We do not append a full stop to AMR graphs with less than five vertices because these often do not represent complete sentences. 

Using the above definitions, we construct a new terminal configuration $c'$ that includes the punctuation marks to be inserted. To this end, we require a set of new vertices $V_\text{punc} = \{ v_\text{punc} \mid v \in \hat{R} \}$ such that $V_\text{punc} \cap \hat{V} = \emptyset$. We set the realization of each vertex $v_\text{punc}$ to the punctuation mark assigned to $\restr{\hat{G}}{v}$ and modify $\hat{\prec}$ such that this punctuation mark is the rightmost word of the subgraph's realization. More formally, we define $c' = (G', \varepsilon, \varepsilon, \rho')$ where
\begin{align*}
G' & = (\hat{V} \cup V_\text{punc}, E', L', \prec') \\
E' & = \hat{E} \cup \{ (v, \star, v_\text{punc}) \mid v \in \hat{R} \} \\
L' & = \hat{L} \cup \{ (v_\text{punc}, \text{punc}(v)) \mid v \in \hat{R} \} \\
\prec' & = (\hat{\prec} \cup  \{ (v', v_\text{punc}) \mid v \in \hat{R}, v' \in \ch[\hat{G}]{v} \cup \{v \} \})^+ \\
\rho' & = \hat{\rho}[\anno{REAL} \mapsto \hat{\rho}(\anno{REAL}) \cup \{ (v_\text{punc}, \text{punc}(v)) \mid v \in \hat{R} \}]
\end{align*}
and compute $\tilde{w} = f_\text{AMR}(c')$.

\begin{figure}
\centering
\scalebox{0.8} {
\begin{tikzpicture}
\tikzstyle{amr-node}=[shape=ellipse,draw, inner sep=0.2, minimum height=0.8cm, text height=1.5ex, text depth=.25ex]
\tikzstyle{text-node}=[text height=1.5ex, text depth=.25ex, align=center]

	\node (PAD2) at (-2.5,0){};
	\node (PAD2) at (2.5,0){};

    \node[amr-node] (amr-island) at (0,0) {$v_1$\,:\,island};
    \node[amr-node] (amr-name) at (0,-1.7) {$v_2$\,:\,name};
    \node[amr-node] (amr-easter) at (-1.6,-3.4) {$v_3$\,:\,`Easter'};
    \node[amr-node] (amr-is2) at (1.6,-3.4) {$v_4$\,:\,`Island'};

    \path [-latex](amr-island) edge node[fill=white] {name} (amr-name);
    \path [-latex](amr-name) edge node[fill=white] {op1} (amr-easter);
    \path [-latex](amr-name) edge node[fill=white] {op2} (amr-is2);    
\end{tikzpicture}
}
\caption{AMR representation of Easter Island}
\label{fig:easterisland}
\end{figure}
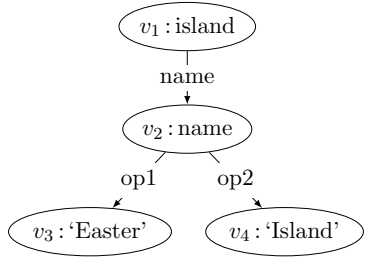

As a final postprocessing step, we remove duplicate words from $\tilde{w}$. That is, whenever a word appears twice in a row in $\tilde{w}$, one of both instances is discarded. Such realizations with duplicate words are occasionally generated by our system due to named instances whose concept shares a common word with its name. An example of such a named instance can be seen in Figure~\ref{fig:easterisland}, where the English word ``island'' is both the concept of vertex $v_1$ and part of its name, possibly resulting in the lower-case realization ``easter island island'' for the whole AMR graph.

\subsection{Hyperparameter Optimization}
\label{GENERATION:HyperparameterOptimization}

Throughout the previous sections, we have introduced several hyperparameters. These parameters include, for example, real-valued weights $\theta_\tau$, $\tau \in \mathcal{C}(T_\text{AMR})$ for transitions and tuples $(n,r) \in \mathbb{N}^+ \times \mathbb{R}^+_0$ for pruning. 
In this section, we will give a short overview on how these parameters can be obtained. 

To simplify the optimization task, we regard each $k$-dimensional hyperparameter $\theta \in \mathbb{R}^k$, $k \in \mathbb{N}$, as a sequence of $k$ one-dimensional hyperparameters. Let $n \in \mathbb{N}$ be the total number of such one-dimensional hyperparameters used in our generation pipeline. As $\mathbb{N} \subseteq \mathbb{R}$, we can write each possible assignment of values to all hyperparameters as a sequence $\Theta = (\theta_1, \ldots, \theta_n) \in \mathbb{R}^n$. To evaluate a particular such assignment $\Theta$, we simply use the development set of an AMR corpus and calculate the Bleu score that the generation algorithm achieves if for all $i \in [n]$, the $i$-th hyperparameter is set to $\theta_i$; we denote the obtained score by $\text{score}_\text{Bleu}(\Theta)$. We are then interested in the highest-scoring assignment
\[
\hat{\Theta} = \argmax_{\Theta \in \mathbb{R}^n }\, \text{score}_\text{Bleu}(\Theta)\,.
\] 
Two commonly used algorithms to approximate the solution to the above equation are \emph{grid search} and \emph{random search}. While the first algorithm defines a set $V_i = \{v_i^1, \ldots, v_i^m \}$, $m \in \mathbb{N}$ of possible values for each hyperparameter $i$ and then performs an exhaustive search over all possible assignments, the latter samples random assignments for a predefined number of times. As reported by \citet{bergstra2012random}, random search is in general the more efficient of both approaches, especially if the number of hyperparameters is high or the evaluation of a hyperparameter set is an expensive operation. We therefore first perform a random search and then try to locally optimize single hyperparameters in the best assignment found during random search. 

To reduce the search space, we introduce for each $i \in [n]$ an interval $r_i = [\text{min}_i, \text{max}_i]$ with $ \text{min}_i \leq \text{max}_i$ and $ \text{min}_i, \text{max}_i \in \mathbb{R}$ that specifies both the minimum and the maximum value that can be assigned to the $i$-th hyperparameter. We then sample several uniformly distributed vectors $(\theta_1, \ldots, \theta_n) \in r_1 \times \ldots \times r_n$ and take the highest-scoring such vector $(\hat{\theta}_1, \ldots, \hat{\theta}_n)$ as an initial assignment. Afterwards, we iterate over all $i \in [n]$ and look whether the total score of vector $(\hat{\theta}_1, \ldots, \hat{\theta}_n)$ can be improved by changing only $\hat{\theta}_i$. To this end, we introduce yet another parameter $s \in \mathbb{N}^+$ and try replacing $\hat{\theta}_i$ by all values contained within the set 
\[
V_i = \{ \text{min}_i + j \cdot \frac{\text{max}_i - \text{min}_i}{s} \mid 0 \leq j \leq s \}\,.
\]
In other words, we try $s+1$ values uniformly distributed between $\min(i)$ and $\max(i)$. 

For a list of all required hyperparameters and further details on the implementation of this hyperparameter optimization algorithm, we refer to Section~\ref{IMPLEMENTATION:Packages:gen}.

\clearpage


\section{Implementation}
\label{IMPLEMENTATION}

We now describe our implementation of the transition-based generator.\footnote{Our implementation can be found at \url{github.com/timoschick/amr-gen}.} This implementation is written entirely in Java, a relatively fast high-level programming language that is also used by most of the external libraries required by our generator. It is worth nothing that our implementation occasionally differs to some extent from the algorithms and formal definitions given in Section~\ref{GENERATION}. While some modifications actually improve the output of our generator, the vast majority thereof is solely due to reasons of efficiency. For example, we do not train a single maximum entropy model $p_\text{TS}$ to estimate $P(t \mid c)$ for all transitions $t \in T_\text{AMR}$ with $\mathcal{C}(t) \notin \{ \textsc{Reorder, Realize}\}$, but instead train independent models for each of the stages identified in Figure~\ref{fig:transition-sequences} (Section~\ref{GENERATION:Decoding}); this makes the training process both faster and more memory efficient by reducing the number of training data per model. 
However, the most important changes in terms of the generator's actual output are that firstly, we enforce several constraints with regards to the applicability of transitions and secondly, we provide \emph{default realizations} in order to cope with AMR concepts not seen during training. 

In the following, we will first discuss all enforced transition constraints in Section~\ref{IMPLEMENTATION:TransitionConstraints} and the embedding of default realizations in Section~\ref{IMPLEMENTATION:DefaultRealizations}. Subsequently, we provide a description of the implementation's overall structure and selective Java classes in Section~\ref{IMPLEMENTATION:Packages}. An overview of external libraries used by our generator is given in Section~\ref{IMPLEMENTATION:ExternalLibraries}. For a more quick and practical introduction on how to use the generator, we refer to the instructions found in the implementation's \texttt{README} file.

\subsection{Transition Constraints}
\label{IMPLEMENTATION:TransitionConstraints}

For each class $\tau \in \mathcal{C}(T_\text{AMR})$, we implement several constraints limiting the number of configurations given which transitions from $\tau$ are applicable. For our discussion of these constraints, let $c = (G, \sigma_1{:}\sigma, \beta, \rho)$ be the current configuration of our transition system where $G = (V,E,L,\prec)$. If $\sigma_1$ has only a single parent node, we denote the latter by $p_{\sigma_1}$. The constraints for each class of transitions are as follows:

\begin{itemize}

\item \textsc{Swap}: We allow this transition only if $\sigma_1$ is not a copy of some other node, i.e. $\sigma_1 \notin \text{dom}(\rho(\anno{LINK}))$. We do so because copies created through \textsc{Delete-Reentrance} transitions can not have any children of their own and thus, the projectivity of $\text{yield}$ does not constitute a problem. Furthermore, we demand that $\sigma_1$ is not a named entity; this can be verified by checking whether there is some $v \in \ch[G]{\sigma_1}$ with $L(v) = \text{name}$. As a final constraint, we demand that $\sigma_1$ and $p_{\sigma_1}$ have not already been swapped in any previous transition step. 

\item \textsc{Merge}: During training, we store for each pair $(p_{\sigma_1}, \sigma_1)$ of merged vertices all assigned concepts and POS tags. From these data, we construct a lookup table 
\[
L_\text{M} \colon L_\text{C} \times L_\text{C} \pfun \Sigma_\text{E}^* \times \mathcal{V}_\anno{POS}
\] mapping each pair of parent and child labels to the tuple of concept and POS tag observed most often. For instance, the lookup table obtained from training with LDC2014T12 (see Section~\ref{PRELIMINARIES:Corpora}) contains, among others, the following entries:
\begin{align*}
& L_\text{M}( \text{early}, \text{more} ) = (\text{earlier}, \text{JJ}) & &
L_\text{M}( \text{likely}, {-} )  = (\text{unlikely}, \text{JJ}) \\
& L_\text{M}( \text{thing}, \text{achieve-01}) = (\text{achievement}, \text{NN}) & & 
L_\text{M}( \text{person}, \text{hunt-01}) = (\text{hunter}, \text{NN})
\end{align*}
We then restrict the number of allowed \textsc{Merge} transitions as follows: Whenever $(L(p_{\sigma_1}), L(\sigma_1)) \notin \text{dom}(L_\text{M})$, i.e. vertices with the same labels as $\sigma_1$ and $p_{\sigma_1}$ have never been merged during training, we disallow all kinds of \textsc{Merge} transitions. Otherwise, we allow only \textsc{Merge}-$L_\text{M}(L(p_{\sigma_1}), L(\sigma_1))$, the \textsc{Merge} transition observed most often for the given pair of labels.  As in the case of \textsc{Swap} transitions, we additionally disallow \textsc{Merge} transitions whenever $\sigma_1$ is a copy of some other node or a named entity.

\item \textsc{Delete}: Again, we disallow \textsc{Delete} transitions for named entities. Although copies created through \textsc{Delete-Reentrance} are often not represented in the generated sentences, we also disallow $\textsc{Delete}$ transitions if $\sigma_1 \in \text{dom}(\rho(\anno{LINK}))$. This is because the realization of such copies is handled exclusively through default realizations as described in Section~\ref{IMPLEMENTATION:DefaultRealizations}.

\item \textsc{Realize}: We implement several restrictions with regards to syntactic annotations; the main purpose of these restrictions is to make the process of computing and storing syntactic annotations more efficient. Whenever a \textsc{Realize-}$(w,\alpha)$ transition is applied, the following must hold:
\begin{align*}
\alpha(\anno{POS}) \neq \text{VB} & \ \Rightarrow\ \alpha(\anno{TENSE}) = \alpha(\anno{VOICE}) = \text{--} \\
\alpha(\anno{POS}) \neq \text{NN} & \ \Rightarrow\ \alpha(\anno{NUMBER}) = \alpha(\anno{DENOM}) = \text{--} \\
\alpha(\anno{NUMBER}) = \text{plural} & \ \Rightarrow\ \alpha(\anno{DENOM}) \neq \text{a}\,.
\end{align*}
To further improve the efficiency of our implementation, whenever the concept represented by $\sigma_1$ is not a PropBank frameset,\footnote{Whether a vertex $v \in V$ represents a PropBank frameset can easily be determined by checking whether $L(v)$ matches the regular expression $\texttt{[A-z]}^+\texttt{-[0-9]}^+$.} we require that $\alpha(\anno{POS}) = \widehat{\text{pos}}(L(\sigma_1))$, i.e. we assign to $\sigma_1$ the POS tag most frequently observed for concept $L(\sigma_1)$ during training (see Definition~\ref{def:empirical-pos}). This restriction stems from the observation that for most concepts which are not PropBank framesets, almost all reasonable realizations have the same simplified part of speech. For  example, it is almost always the case that instances of the concepts ``boy'', ``city'' and ``world'' are realized as nouns and instances of ``early'', ``rich'' and ``fast'' are realized as adverbs or adjectives.
If $\sigma_1 \in \text{dom}(\rho(\anno{LINK}))$, we only allow \textsc{Realize-}$(w,\alpha)$ if $w$ is one of the default realizations assigned to $c$ and $\alpha$ (see Section~\ref{IMPLEMENTATION:DefaultRealizations}).

In our implementation of Algorithm~\ref{alg:lm-generation}, we do not consider all possible syntactic annotations when computing the $n_1$-best \textsc{Realize} transitions. Instead, we only consider the $n_k$-best values for each syntactic annotation key $k \in \mathcal{K}_\text{syn}$ where $n_k \in \mathbb{N}$ is some hyperparameter.

\item \textsc{Insert-Child}: We allow at most one \textsc{Insert-Child} transition per vertex and we only allow vertices to be inserted left of $\sigma_1$; both restrictions are purely on grounds of efficiency. Furthermore, we manually handle insertions of articles and auxiliary verbs required by passive constructions as these can directly be inferred from the syntactic annotation values $\rho(\anno{DENOM})(\sigma_1)$ and $\rho(\anno{VOICE})(\sigma_1)$, respectively.

\item \textsc{Reorder}: As the number of possible reorderings for some vertex $v$ grows superexponentially with the number of its children, we implement several constraints to reduce the number of reorderings to be considered. Let $\textsc{Reorder-}(v_1, \ldots, v_n)$ be the \textsc{Reorder} transition whose applicability is to be checked and let
\[
{\lessdot} = \{ (v_i, v_j) \mid 1 \leq i < j \leq n \}
\]
denote the total order such that $(v_1, \ldots, v_n)$ is the $(\ch{\sigma_1} \cup \{ \sigma_1 \})$-sequence induced by $\lessdot$.
If $\sigma_1$ has some child $c_{\sigma_1}$ with $L(c_{\sigma_1}) \in \{\text{the, a, an}\}$, we demand that $c_{\sigma_1}$ occurs before $\sigma_1$ and all of its other children, i.e. $c_{\sigma_1} = v_1$. For enumerations and listings, we require that the order defined through edge labels of the form $\text{OP}i$, $i \in \mathbb{N}$ be preserved. In other words, if $\sigma_1$ has children $c_1, \ldots, c_m$ where each child $c_i$ is connected to $\sigma_1$ through an edge with label $\text{OP}i$, we demand that $c_j \lessdot c_k$ for all $1 \leq j < k \leq m$. 
We implement several more such restrictions; for a full list thereof, we refer to Section~\ref{IMPLEMENTATION:Packages:gen}. 
 
\item \textsc{Insert-Between}: We restrict the allowed labels for vertices inserted through left and right \textsc{Insert-Between} transitions to two handwritten sets $W_\textsf{left}$ and $W_\textsf{right}$, containing the insertions observed most frequently during training as well as common English prepositions (see Section~\ref{IMPLEMENTATION:Packages:misc}). As children connected to $\sigma_1$ through an edge with label ``$\text{domain}$'' almost always require a \textsc{Insert-Between-}$(w, \mathsf{right})$ transition with $w \in \langle \text{be} \rangle$, we handle this special case manually.

\end{itemize}

\subsection{Default Realizations}
\label{IMPLEMENTATION:DefaultRealizations}

As some AMR concepts are either not observed at all during training or only some specific forms thereof are observed (for example, a verb may occur in the training corpus only in past tense), we provide default realizations $\tilde{r}_{(c, \alpha)}$ for some pairs $(c, \alpha) \in C_\text{AMR} \times \mathcal{A}_\text{syn}$. Given some configuration $c = (G, \sigma_1{:}\sigma, \varepsilon, \rho)$ in which \textsc{Realize} transitions are applicable, we then set
\[ 
P(\textsc{Realize-}(\tilde{r}_{(c, \alpha)},\alpha) \mid c, \alpha) = \tilde{p}
\] for all $\alpha \in \mathcal{A}_\text{syn}$ where $\tilde{p} \in [0,1]$ is some hyperparameter; in order to assure that $P$ is still a valid probability measure, we subtract a small amount $\delta$ from the probabilities of all other applicable $\textsc{Realize}$ transitions. 

Let the current configuration be of the form $c = (G, \sigma_1{:}\sigma, \varepsilon, \rho)$ with $G = (V,E,L,\prec)$ and let $\alpha \in \mathcal{A}_\text{syn}$ be a syntactic annotation for $\sigma_1$.
If $\sigma_1$ is a noun, verb, adjective or adverb according to $\alpha$ and not a copy of some other node, i.e. $\alpha(\anno{POS}) \in \{ \text{NN, VB, JJ} \}$ and $\sigma_1 \notin \text{dom}(\rho(\anno{LINK}))$, we determine $\tilde{r}_{(c, \alpha)}$ as follows: If $L(\sigma_1)$ is a PropBank frameset, we first remove the frameset id from it; for example, we turn the instances ``want-01'' and ``develop-02'' into ``want'' and ``develop'', respectively. Let $l_{\sigma_1}$ denote the so-obtained truncated label. We query WordNet \citep{fellbaum1998wordnet,miller1995wordnet} to find out whether a word with lemma $l_{\sigma_1}$ and POS tag $\alpha(\anno{POS})$ exists; if this is not the case, no default realization $\tilde{r}_{(c, \alpha)}$ can be found. Otherwise, we use SimpleNLG \citep{gatt2009simplenlg} to turn $l_{\sigma_1}$ into the required word form according to $\alpha$. This is done by first instantiating a \emph{phrase} consisting only of $l_{\sigma_1}$ and then specifying \emph{features} of this phrase. For example, the number of a noun can be set to some value \texttt{num} as follows:
\begin{code}
phrase.setFeature(Feature.NUMBER, num);
\end{code}
The so-obtained word is then returned as a default realization $\tilde{r}_{(c, \alpha)}$. For $\alpha(\anno{POS}) = \text{JJ}$, if $l_{\sigma_1}$ can serve as both an adjective and an adverb, both forms are used as default realizations with probabilities of $\tilde{p}/2$ each. For example, given $l_{\sigma_1} = \text{quick}$, both ``quick'' and ``quickly'' are returned.

If  $\alpha(\anno{POS}) \notin \{ \text{NN, VB, JJ} \}$, we check whether $l_{\sigma_1}$ is a pronoun and if so, we provide both the corresponding personal pronoun and possessive pronoun forms as default realizations, each with probability $\tilde{p}/2$. Importantly, this is also done if $\sigma_1$ is a copy of some other vertex, but in this case, we make use of yet another hyperparameter $p_\varepsilon \in [0, \tilde{p}]$, set the probabilities of both realizations to $(\tilde{p}-p_\varepsilon)/2$ and add $\varepsilon$ as another default realization with probability $p_\varepsilon$.
If none of the above applies and $\sigma_1 \in \text{dom}(\rho(\anno{LINK}))$, we return only $\varepsilon$ as a default realization.

\begin{figure}
\centering
\scalebox{0.8} {
\begin{tikzpicture}
\tikzstyle{amr-node}=[shape=ellipse,draw, inner sep=0.2, minimum height=0.8cm, text height=1.5ex, text depth=.25ex]
\tikzstyle{text-node}=[text height=1.5ex, text depth=.25ex, align=center]

	\node (PAD2) at (-2.5,0){};
	\node (PAD2) at (2.5,0){};

    \node[amr-node] (amr-project) at (0,0) {1\,:\,project};
    \node[amr-node] (amr-name) at (0,-1.65) {2\,:\,name};
    \node[amr-node] (amr-three) at (-1.6,-3.3) {3\,:\,`Three'};
    \node[amr-node] (amr-gorges) at (1.6,-3.3) {4\,:\,`Gorges'};

    \path [-latex](amr-project) edge node[fill=white] {name} (amr-name);
    \path [-latex](amr-name) edge node[fill=white] {op1} (amr-three);
    \path [-latex](amr-name) edge node[fill=white] {op2} (amr-gorges);    
\end{tikzpicture}
}
\caption{AMR representation of the ``Three Gorges'' project}
\label{fig:prepare-gorges}
\end{figure}
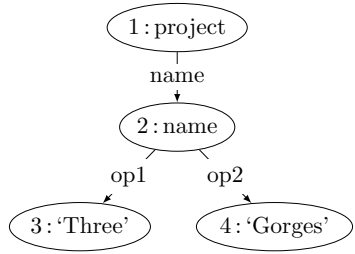

Apart from this basic handling of unknown instances and pronouns, we also provide special realization rules for named entities (i.e. vertices with a child labeled ``name''), dates and numbers. For named entities, we remove all vertices encoding the name from the AMR graph and keep only the concept itself, for which we allow three different kinds of default realizations: nothing but the name, the name followed by the concept and the concept followed by the name. For instance, consider the AMR graph shown in Figure~\ref{fig:prepare-gorges}. As this graph represents a named entity, we remove from it all vertices but the root, for which we provide the three default realizations ``Three Gorges'', ``Three Gorges project'' and ``project Three Gorges''. If the named entity has already been observed during training, we choose from these three candidates the realization assigned to it most often to be the default realization. Otherwise, if at least the concept of the named entity has already been observed during training, we choose the arrangement observed most often for this concept. If neither the name nor the concept were observed during training, we take only the name itself as the default realization. An exception to the above rules are countries, world regions and continents, for which the default realizations are both the name and the corresponding adjective, each with probability $\tilde{p}/2$.\footnote{The adjective forms corresponding to countries and nations are extracted from \url{en.wikipedia.org/wiki/List_of_adjectival_and_demonymic_forms_for_countries_and_nations}.} For example, an instance of the AMR concept ``country'' with name ``France'' gets assigned the default realizations ``France'' and ``French''.

Date entities are converted to month-day-year format, resulting in strings like ``April 2 2016'' or ``July 24 2011''. Finally, numbers that are not part of a date are converted to ordinal numbers if their parent is an instance of the concept ``ordinal-entity'' and otherwise left as is, but if they end with six or nine zeros, the latter are replaced by the string ``million'' or ``billion'', respectively.

\subsection{Packages}
\label{IMPLEMENTATION:Packages}

Our implementation of the transition-based generator is divided into five packages \texttt{main}, \texttt{dag}, \texttt{ml}, \texttt{gen} and \texttt{misc}. For each of these packages, we discuss here only the most important classes contained therein and the functionality they provide; for a thorough description of all classes and functions, we refer to the \emph{Javadoc} documentation available in the \texttt{javadoc} subdirectory of our implementation. 

\subsubsection{\texttt{main}} 

The \texttt{main} package consists only of the two classes \texttt{PathList} and \texttt{AmrMain}. While the former contains nothing but string constants referring to the paths of training, development and test data, trained maximum entropy models and various external resources, the latter provides wrapper functions for the most important tasks to be performed by our implementation: Generation, training and hyperparameter optimization can be performed using the methods \texttt{generate()}, \texttt{train()} and \texttt{optimizeHyperparams()}, respectively. While the first method can be called with an arbitrary list of AMR graphs as parameter, the other methods require the training and development corpora to be found in the directories specified in \texttt{PathList}. Assuming that they are stored in official AMR format,\footnote{See \url{github.com/amrisi/amr-guidelines/blob/master/amr.md} for a description of this format.} AMR graphs can be read from a file using the \texttt{loadAmrGraphs()} function.

To train the generator using \texttt{train()}, each subdirectory of the training directory (specified in \texttt{PathList.AMR\_SUBDIRECTORIES} and \texttt{PathList.TRAINING\_DIR}, respectively) must contain all information required to build an extended corpus (see Section~\ref{TRAINING:Preparations}), but this information is to be distributed among several files. These files must go by the following names specified in \texttt{PathList} and should contain the following information:
 
 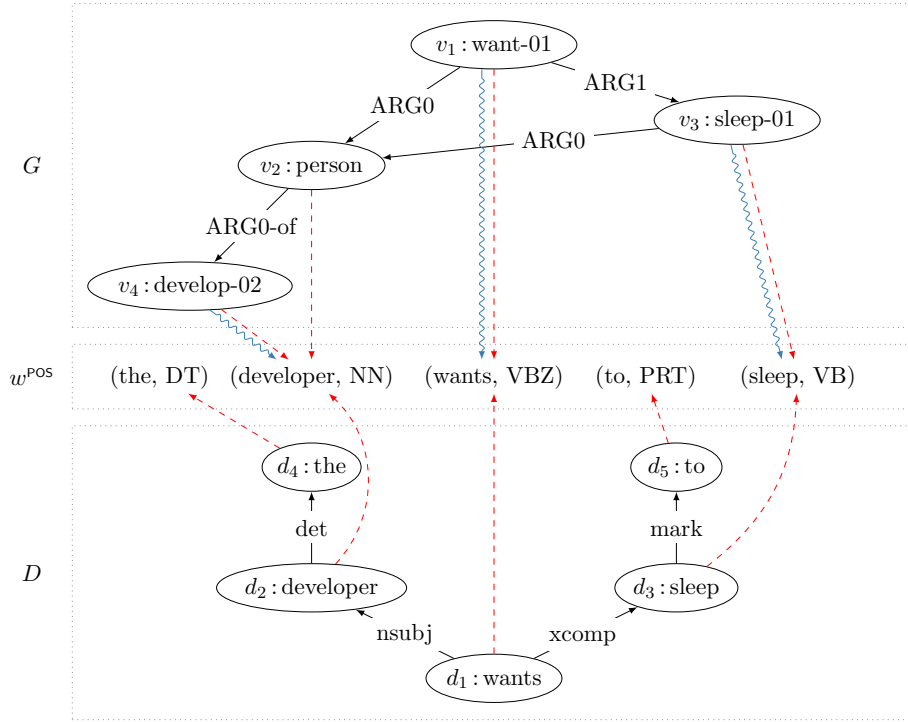
\begin{figure}[t!]
 \centering
 \scalebox{0.8} {
 \begin{tikzpicture}
 \tikzstyle{amr-node}=[shape=ellipse,draw, inner sep=0.2, minimum height=0.8cm, text height=1.5ex, text depth=.25ex]
 \tikzstyle{dep-node}=[shape=ellipse,draw, inner sep=0.2, minimum height=0.8cm, minimum width=1.5cm, text height=1.5ex, text depth=.25ex]
 \tikzstyle{text-node}=[text height=1.5ex, text depth=.25ex]
 \tikzstyle{amr-align}=[-latex,color=red, dashed]
 \tikzstyle{amr-align-alt}=[-latex,color=plot1, decorate,
     decoration={%
       snake,
       segment length=1.64mm,
       amplitude=0.4mm,
       pre length=4pt,
       post length=4pt,
     }]
 \tikzstyle{dep-align}=[-latex,color=red, dashed]
 \tikzstyle{amr-edge}=[text height=1.5ex, text depth=.25ex, fill=white]
 \tikzstyle{frame}=[draw, dotted, color=gray, inner sep=0.3cm]
 
 	\node (PAD1) at (-8,0){};
 	\node (PAD2) at (8,0){};
 
     \node[amr-node] (amr-want) at (0,0) {$v_1$\,:\,want-01};
     \node[amr-node] (amr-person) at (-3,-2) {$v_2$\,:\,person};
     \node[amr-node] (amr-sleep) at (4,-1.25) {$v_3$\,:\,sleep-01};
     \node[amr-node] (amr-develop) at (-5,-4) {$v_4$\,:\,develop-02};
 
     \path [-latex](amr-want) edge node[fill=white] {ARG0} (amr-person);
     \path [-latex](amr-want) edge node[fill=white] (amr-want-arg1) {ARG1} (amr-sleep);
     \path [-latex](amr-sleep) edge node[fill=white, pos=0.38] {ARG0} (amr-person);
     \path [-latex](amr-person) edge node[fill=white] {ARG0-of} (amr-develop);
     
     \node (amr-pad1) at (-6.5,0.25){};
     \node (amr-pad2) at (6.5,-4.25){};
         \node[frame, fit=(amr-pad1)(amr-pad2)] (amr-frame) {};
     
     \node[text-node, align=center] (text-the) at (-5.5, -5.5) {(the, DT)};
     \node[text-node] (text-developer) at (-3, -5.5) {(developer, NN)};
     \node[text-node] (text-wants) at (0, -5.5) {(wants, VBZ)};
     \node[text-node] (text-to) at (2.5, -5.5) {(to, PRT)};
     \node[text-node] (text-sleep) at (5, -5.5) {(sleep, VB)};
     
     \node (text-pad1) at (-6.5,-5.4){};
     \node (text-pad2) at (6.5,-5.6){};
         \node[frame, fit=(text-pad1)(text-pad2)] (text-frame) {};
     
     \path[amr-align] (amr-person) edge node[] {} (text-developer);
     \path[amr-align] (amr-develop) edge node[] {} (text-developer);
     \path[amr-align] (amr-want) edge node[] {} (text-wants);
     \path[amr-align] (amr-sleep) edge node[] {} (text-sleep);
     
       \draw[amr-align-alt, transform canvas={xshift=-2mm}] (amr-develop) -- (text-developer);
       \draw[amr-align-alt, transform canvas={xshift=-2mm}] (amr-want) -- (text-wants);
       \draw[amr-align-alt, transform canvas={xshift=-2mm}] (amr-sleep) -- (text-sleep);
     

         \node[dep-node] (dep-the) at (-3, -7) {$d_4$\,:\,the};
         \node[dep-node] (dep-to) at (3, -7) {$d_5$\,:\,to};
         \node[dep-node] (dep-developer) at (-3, -9) {$d_2$\,:\,developer};
         \node[dep-node] (dep-sleep) at (3, -9) {$d_3$\,:\,sleep};
         \node[dep-node] (dep-wants) at (0, -10.5) {$d_1$\,:\,wants};
     
         \path [-latex](dep-wants) edge node[amr-edge] {nsubj} (dep-developer);
         \path [-latex](dep-wants) edge node[amr-edge, pos=0.47] {xcomp} (dep-sleep);
         \path [-latex](dep-developer) edge node[fill=white] {det} (dep-the);
         \path [-latex](dep-sleep) edge node[fill=white] {mark} (dep-to); 
         
         \node (dep-pad1) at (-6.5,-6.75){};
         \node (dep-pad2) at (6.5,-10.75){};
         \node[frame, fit=(dep-pad1)(dep-pad2)] (dep-frame) {};
         
         \path[dep-align] (dep-the) edge node[] {} (text-the);
         \path[dep-align, bend angle=45, bend right] (dep-developer) edge node[] {} (text-developer);
         \path[dep-align] (dep-wants) edge node[] {} (text-wants);
         \path[dep-align, bend angle=25, bend right] (dep-sleep) edge node[] {} (text-sleep);
         \path[dep-align] (dep-to) edge node[] {} (text-to); 
         
             \node[] at (-7.6, -2) {$G$}; 
             \node[] at (-7.6, -5.5) {$w^\anno{POS}$}; 
             \node[] at (-7.6, -8.75) {$D$};   
     
 \end{tikzpicture}
 }
 \caption{Graphical representation of the bigraph  $\mathcal{B} = (G, D, w^\anno{POS}, A_G, A_D)$ as described in Example~\ref{example:training}. For $i \in \{G, D\}$, each node $v \in V_i$ is inscribed with $v$\,:\,$L_i(v)$; each alignment $(u,j) \in A_i$ is represented by a dashed arrow line connecting $u$ and $w^\anno{POS}(j)$. An additional alignment $A_G' \subseteq V_G \times [|w^\anno{POS}|]$ is indicated through wavy arrow lines.}
 \label{fig:amr-encoding}
 \end{figure}
 
\begin{itemize}
\item \texttt{PathList.AMR\_FILENAME}: This file must contain a list of aligned and tokenized AMR graphs, separated by empty lines and encoded using the official AMR format. The alignments must be stored in the format used by \citet{flanigan2014discriminative}.\footnote{See \url{github.com/jflanigan/jamr/blob/Generator/docs/Alignment_Format.md} for a description of this format.}
Above each AMR graph, there must be a line starting with \texttt{\# ::tok} containing a tokenized reference realization and a line starting with \texttt{\# ::alignments} containing the alignments. Additional annotations -- such as the non-tokenized reference realization -- are allowed, but ignored during the training procedure. For example, the AMR graph shown in Figure~\ref{fig:amr-encoding}, its reference realization and the corresponding alignment $A_G$ may be represented like this:

\begin{code}
# ::tok the developer wants to sleep
# ::alignments 1-2|0.0+0.0.0 2-3|0 4-5|0.1	
(v1 / want-01
      :ARG0 (v2 / person
      		:ARG0-of (v4 / develop-02))
      :ARG1 (v3 / sleep-01
      		:ARG0 v2))
\end{code}
\vspace{-\medskipamount}
\item \texttt{PathList.DEPENDENCIES\_FILENAME}: This file must contain a list of dependency trees which correspond to the AMR graphs found in the above file in a one-to-one manner. The dependency trees must be separated by empty lines and encoded in \emph{Stanford dependencies} (SD) format.\footnote{See \url{nlp.stanford.edu/software/stanford-dependencies.shtml} for a description of this format.} To give an example, the dependency tree shown in Figure~\ref{fig:amr-encoding} can be encoded as follows: 

\begin{code}
root(ROOT-0, wants-3)
nsubj(wants-3, developer-2)
xcomp(wants-3, sleep-5)
det(developer-2, the-1)
mark(sleep-5, to-4)
\end{code}
\vspace{-\medskipamount}
\item \texttt{PathList.POS\_FILENAME}: This file should contain a newline-separated list of POS sequences where POS tags are separated by tabs. The $i$-th sequence of POS tags must correspond to the reference realization of the $i$-th AMR graph found in the \texttt{PathList.AMR\_FILENAME} file. The following entry corresponds to the reference realization shown in Figure~\ref{fig:amr-encoding}:

\begin{code}
DT	NN	VBZ	PRT	VB
\end{code}
\vspace{-\medskipamount}
\item \texttt{PathList.EM\_ALIGNMENTS\_FILENAME}: This file should contain a newline-separated list of alignments in the format used by the string-to-string aligner described in \citet{pourdamghani2014aligning}.\footnote{Note that this format differs slightly from the one used by \citet{flanigan2014discriminative}.} The $i$-th alignment must correspond to the reference realization of the $i$-th AMR graph found in the \texttt{PathList.AMR\_FILENAME} file. For example, the entry encoding the additional alignment $A_G'$ shown in Figure~\ref{fig:amr-encoding} may look like this:

\begin{code}
1-1.1.1 2-1 4-1.2 
\end{code}
\vspace{-\medskipamount}
\end{itemize}
The training procedure requires at least 8GB of RAM and may take several hours to days, depending on the used hardware. It is important to note that when training the generator with the \texttt{train()} method on a different corpus than LDC2014T12, some of the resources found in directory \texttt{res} must also be rebuilt using the corresponding methods provided by \texttt{misc.StaticHelper}. For more information on this process, we refer to the Javadoc documentation of the latter class and to the \texttt{README} file.

Our implementation also supports the command-line based generation of English sentences from AMR graphs. For generation using the command line, the following parameters may be specified:

\begin{itemize}

\item \texttt{--input} (\texttt{-i}): The file in which the input graphs are stored in official AMR format. If this parameter is not specified, it is assumed that the required AMR graphs can be found in the subdirectories of the \texttt{PathList.TEST\_DIR} file.

\item \texttt{--output} (\texttt{-o}): The file in which the generated sentences should be saved. This is the only mandatory parameter.

\item \texttt{--bleu} (\texttt{-b}): If this flag is set, the Bleu score achieved by the generator on the given data set is printed to the standard output stream. This is only possible if the AMR graphs are stored with tokenized reference realizations in the input file.

\item \texttt{--show-output} (\texttt{-s}): If this flag is set, pairs of reference realizations and corresponding generated sentences are printed to the standard output stream once the generator is finished. Again, this can only be done if the AMR graphs are stored with tokenized reference realizations in the input file.

\end{itemize}

As the generation process requires around 8GB of RAM, the generator should always be run with parameter \texttt{-Xmx8g}. For example, the command

\begin{code}
java -jar -Xmx8g AmrGen.jar --input in.txt --output out.txt --bleu
\end{code}
can be used to generate sentences from all AMR graphs found in \texttt{in.txt}, write them to \texttt{out.txt} and print the obtained Bleu score to the standard output stream.

\subsubsection{\texttt{dag}}
\label{IMPLEMENTATION:Packages:dag}

This package contains classes that are closely related to labeled ordered graphs as introduced in Definition~\ref{def:graph}. Most importantly, the class \texttt{DirectedGraph} is used to model actual graphs; their vertices and edges are represented by instances of \texttt{Vertex} and \texttt{Edge}, respectively. 

Although they could theoretically be modeled using just the above classes, a wrapper class \texttt{DependencyTree} is used to represent dependency trees and a class \texttt{Amr} is used to represent AMR graphs. Bigraphs are not explicitly modeled; instead, AMR graphs simply store a reference to the corresponding dependency tree. If given, the \texttt{Amr} class also stores the reference realization of the graph and the corresponding alignment as well as POS tags. Furthermore, it provides some convenient methods and functions for the handling of AMR graphs. For example, the \texttt{calculateSpan()} method can be used to calculate the span of each vertex and \texttt{yield()} implements both $\text{yield}_{(G,\rho)}$ and $\text{yield}_{(G,\rho)}^\text{par}$. Another important method provided by this class is \texttt{prepare()} and its subroutines \texttt{prepareForTesting()} and \texttt{prepareForTraining()}, which prepare an AMR graph either for training or testing; this preparation includes, among others, collapsing named entities into a single node for more efficient processing, converting the reference realization to lower case and computing the span of each vertex. The \texttt{prepareForTraining()} method also defines all alignment rules mentioned in Section~\ref{TRAINING:Preparations}.

In addition to the above functionality, the package \texttt{dag} provides two classes \texttt{AmrFrame} and \texttt{DependencyTreeFrame} which provide means of visualizing both dependency trees and AMR graphs; these classes are also capable of showing alignments between graphs and their realizations as well as annotations assigned to vertices.

\subsubsection{\texttt{gen}}
\label{IMPLEMENTATION:Packages:gen}

This package constitutes the core of our generator. The actual generation algorithm is implemented in the classes \texttt{FirstStageProcessor} and \texttt{SecondStageProcessor}. The former contains a method \texttt{processFirstStage()} which implements the restricted version of the greedy generation algorithm, applying only transitions from the set $T_\text{restr}$ to its input; the latter contains the rest of the logic required by the generation algorithm. Most importantly, it contains a function \texttt{getBest()}, which is a straightforward implementation of Algorithm~\ref{alg:lm-generation}, the best transition sequence algorithm. Default realizations as defined in Section~\ref{IMPLEMENTATION:DefaultRealizations} and required by this method are provided by the \texttt{getDefaultRealizations()} function of class \texttt{DefaultRealizer}. A full list of restrictions for \textsc{Reorder} transitions can be found in class \texttt{PositionHelper}, which also contains a method to compute $n$-best reorderings. Finally, the \texttt{postProcess()} method of class \texttt{PostProcessor} can be used to perform postprocessing as described in Section~\ref{GENERATION:Postprocessing}.

For training the various maximum entropy models required by our generator, the non-instantiable classes \texttt{GoldSyntacticAnnotations} and \texttt{GoldTransitions} contain static methods to obtain gold syntactic annotation values and gold transitions, respectively. These classes implement all approaches devised in Sections~\ref{TRAINING:SyntacticAnnotations}~and~\ref{TRAINING:Transitions}, with the sole exception of \textsc{Delete-Reentrance} transitions, for which only the text-based approach is implemented. This is the case because a qualitative analysis of several dozen AMR graphs from the LDC2014T12 corpus showed both approaches to give almost identical results, but this approach performed slightly better than the dependency-tree-based approach and is much easier to implement. 

Hyperparameters used throughout the generation process are managed by the classes \texttt{Hyperparam} and \texttt{IntHyperparam}; the former also contains methods to perform random search and grid search as explained in Section~\ref{GENERATION:HyperparameterOptimization}. For a list of all hyperparameters and a short explanation thereof, we refer to the documentation of the \texttt{Hyperparams} class.

\subsubsection{\texttt{ml}}
\label{IMPLEMENTATION:Packages:ml}

This package contains all classes related to maximum entropy modeling. 
As mentioned before, we do not use a single maximum entropy model $p_\text{TS}$ to estimate $P(t \mid c)$ for all transitions $t \in T_\text{AMR}$, but instead train independent such models for each stage identified in Figure~\ref{fig:transition-sequences} (Section~\ref{GENERATION:Decoding}). On grounds of efficiency, we additionally use two different maximum entropy models for \textsc{Insert-Between} transitions: The model implemented by \texttt{ArgInsertionMaxentModel} is queried whenever the vertex on top of the node buffer is connected to its child through a PropBank semantic role (i.e. the edge connecting both vertices has a label of the form ARG$i$ for some $i \in \mathbb{N}$); in all other cases, we use the model implemented by \texttt{OtherInsertionMaxentModel}.
 
All classes representing maximum entropy models can be identified by their common suffix \texttt{MaxentModel}; they are subclasses of either \texttt{OpenNlpMaxentModelImplementation}, an implementation of maximum entropy models based on the \texttt{GISModel} class provided by OpenNLP, or \texttt{StanfordMaxentModelImplementation}, an implementation using the Stanford Classifier.\footnote{For further details on OpenNLP and the Stanford Classifier, we refer to \url{opennlp.apache.org} and \url{nlp.stanford.edu/software/classifier.shtml}, respectively.} The \texttt{IndicatorFeature} interface and its two implementations \texttt{StringFeature} and \texttt{ListFeature} provide means of representing features.

\subsubsection{\texttt{misc}}
\label{IMPLEMENTATION:Packages:misc}

The package \texttt{misc} contains miscellaneous classes whose methods are used in various places throughout the implementation. For example, the class \texttt{PosHelper} provides the $\text{simplify}$ mapping defined in Section~\ref{TRAINING:SyntacticAnnotations} and \texttt{PrunedList} implements the function $\text{prune}_n$ as introduced in Definition~\ref{def:prune}. The class \texttt{StaticHelper} contains functions for generating additional resources required by the generator, such as the lookup table $L_\text{M}$ for \textsc{Merge} transitions introduced in Section~\ref{IMPLEMENTATION:TransitionConstraints}. The \texttt{WordNetHelper} class provides an interface to WordNet \citep{fellbaum1998wordnet,miller1995wordnet}. Importantly, the class \texttt{WordLists} contains several collections of words required by the generator; for example, the words allowed for \textsc{Insert-Between} and \textsc{Insert-Child} transitions are defined therein. 

\subsection{External Libraries}
\label{IMPLEMENTATION:ExternalLibraries}

Our implementation makes use of several external libraries for various purposes such as POS tagging, language modeling, maximum entropy modeling and computing Bleu scores.
Below, we list all external libraries embedded into our generator and briefly explain how they are used:

\begin{itemize}

\item The \emph{Extended Java WordNet Library} (available at \url{extjwnl.sourceforge.net}) is used to access WordNet \citep{miller1995wordnet,fellbaum1998wordnet} which, in turn, is required for default realizations and to compute some features of our maximum entropy models.

\item We use both the \emph{Apache OpenNLP} library (available at \url{opennlp.apache.org}) and the \emph{Stanford Classifier} (available at \url{nlp.stanford.edu/software/classifier.shtml}) for maximum entropy modeling; while the training procedure provided by the former library is both faster and more memory-efficient, we achieved slightly better results using the latter. 

\item The \emph{Berkeley Language Model} \citep{pauls2011faster} is used for computing $\text{score}_\text{LM}$, the language model score assigned to generated sentences. It provides methods for efficiently loading and accessing large $n$-gram language models. 

\item For POS tagging of our training and development data, we use the \emph{Stanford Log-linear Part-Of-Speech Tagger} \citep{toutanova2003feature}, a part of the \emph{Stanford CoreNLP} toolkit \citep{manning2014stanford}.

\item \emph{SimpleNLG} \citep{gatt2009simplenlg} is used to determine default realizations.

\item We use the \texttt{BleuMetric} implementation of \emph{Phrasal} \citep{spence2014phrasal} to compute the Bleu score obtained by our generator.

\item To graphically display AMR graphs and dependency trees, we use several classes provided by \emph{JGraphX} (available at \url{github.com/jgraph/jgraphx}).

\item For parsing command line options, we make use of \emph{JCommander} (available at \url{jcommander.org}).

\end{itemize}

\clearpage


\section{Experiments}
\label{EXPERIMENTS}

We evaluate our approach by studying the results of several experiments conducted using the implementation described in Section~\ref{IMPLEMENTATION}. For carrying out these experiments, a single machine with 8GB of RAM and a 2.40GHz Intel\textsuperscript{\textregistered} Core\texttrademark i7-3630QM CPU with eight cores was used; the operating system was Ubuntu 16.10.

All experiments reported in this section were performed using the LDC2014T12 corpus, containing $10,\!313$ training AMR graphs, $1,\!368$ development AMR graphs and $1,\!371$ test AMR graphs (see Table~\ref{tab:corpora}, Section~\ref{PRELIMINARIES:Corpora}). The reference realizations of all AMR graphs in the training and development set were tokenized using \emph{cdec} \citep{dyer2010cdec} and annotated with POS tags using the \emph{Stanford Log-linear Part-of-Speech Tagger} \citep{toutanova2003feature}; dependency trees were obtained using the BLLIP parser \citep{charniak2000maximum,charniak2005coarse} and subsequently converted into the format required by our generator using the \emph{Stanford Dependencies Converter}.\footnote{For further details on the Stanford Dependencies format and the conversion process, see \url{nlp.stanford.edu/software/stanford-dependencies.shtml}.} Alignments between AMR graphs and reference realizations were obtained using the methods by \citet{flanigan2014discriminative} and \citet{pourdamghani2014aligning} and fused as described in Section~\ref{TRAINING:Preparations}. 
For language modeling, we used a 3-gram model with Kneser-Ney smoothing trained on Gigaword v1 (LDC2003T05).\footnote{The used Gigaword $n$-gram counts are available at \url{www.keithv.com/software/giga/}.} The corresponding language model file in binary format can be found in the file \texttt{res/lm.binary} of our implementation.

We manually compared the quality of gold annotations and transitions returned by the alternative approaches devised in Sections~\ref{TRAINING:SyntacticAnnotations}~and~\ref{TRAINING:Transitions} on a small number of development AMR graphs; in the vast majority of cases, both approaches returned exactly the same. However, using dependency trees to determine gold denominators turned out to be slightly more error-prone, the reason being that the automatically generated dependency trees for some realizations were themselves erroneous. For \textsc{Insert-Child} and \textsc{Insert-Between} transitions, it happened occasionally that one of both approaches returned nonsensical transitions, but it was very rarely the case that both approaches failed simultaneously.
Therefore, in all of the experiments discussed below, we used the purely text-based approach to obtain gold denominators during training; for \textsc{Insert-Child} and \textsc{Insert-Between} transitions, we used both approaches concurrently, thus doubling the number of available training data. Hyperparameter optimization was performed as described in Section~\ref{GENERATION:HyperparameterOptimization} with parameter $s = 15$, resulting in the configuration found in the file \texttt{res/hyperparams.txt}. 

As a first experiment, we used the fully trained system to generate realizations for all AMR graphs in the development and test set of LDC2014T12 and computed the corresponding Bleu scores.\footnote{Throughout this section, we implicitly mean the case-insensitive $1\mydots4$-gram Bleu score with scaling factor $s = 100$, rounded to the first decimal place, whenever we speak of Bleu scores.} Our approach achieves a Bleu score of 27.4 on both the development and test set. A comparison of these results with the scores achieved by all other currently published approaches can be seen in Table~\ref{tab:comparison}; therein and throughout this section, we abbreviate the tree-transducer-based approach of \cite{flanigan2016generation} by JAMR-gen, the phrase-based generator of \cite{pourdamghani2016generating} by PBMT-gen, the approach of \citet{song2016amr} based on a traveling salesman problem solver by TSP-gen, the synchronous node replacement grammar approach of \citet{song2017amr} by SNRG-gen and the generator of \citet{konstas2017neural} using a neural network architecture by NEUR-gen. Whenever available, Table~\ref{tab:comparison} lists the results obtained with the LDC2014T12 corpus as this is the corpus used for our experiments, thus allowing for better comparisons than LDC2015E86.

\begin{table}[!t]
\centering
\bgroup
\newcolumntype{Y}{>{\raggedright\arraybackslash}X}
\def\arraystretch{1.5}

\begin{tabularx}{0.795\textwidth}{|l|c|Y|c|c|c|}
\hline
\textbf{System} & \textbf{LM} & \textbf{Corpus} & $l_\text{max}$ & \textbf{Dev} & \textbf{Test} \\
\hline
\multirow{ 2}{*}{Our approach} & \multirow{ 2}{*}{3-gram} & \multirow{ 2}{*}{LDC2014T12} & $\infty$ & 27.4 & 27.4 \\
 & & & 30 & 28.3 & 28.9 \\
 \hdashline
JAMR-gen \hspace{0pt plus 1filll} \small(2016) & 5-gram & LDC2014T12 & $\infty$ & 22.7 & 22.0 \\
PBMT-gen \hspace{0pt plus 1filll} \small(2016) & 5-gram & LDC2014T12 & $\infty$ & 27.2 & 26.9 \\
TSP-gen \hspace{0pt plus 1filll} \small(2016) & 4-gram & LDC2015E86 & 30 & 21.1 & 22.4 \\
SNRG-gen \hspace{0pt plus 1filll} \small(2017) & 4-gram & LDC2015E86 & 30 & 25.2 & 25.6 \\
NEUR-gen \hspace{0pt plus 1filll} \small(2017) & -- & LDC2014T12, LDC2011T07 & $\infty$ & \multicolumn{1}{c|}{--} & 29.7 \\
\hline
\end{tabularx}
\caption{Comparison of our approach with other generators. The ``LM'' column lists the kind of language model used, the ``Corpus'' column contains the used corpora and the ``$l_\text{max}$'' column contains the maximum number of words in the reference realization for an AMR graph to be considered for Bleu score computation. The ``Dev'' and ``Test'' columns show the Bleu scores obtained on the development and test sets, rounded to the first decimal place.}
\label{tab:comparison}
\egroup
\end{table}

In terms of Bleu scores, our approach performs much better than JAMR-gen, TSP-gen and SNRG-gen and slightly better than PBMT-gen, but worse than NEUR-gen. For the comparison with the TSP-gen and SNRG-gen generators, we must take into account that these systems were both trained using the LDC2015E86 corpus; while the test and development sets in this corpus are exactly the same as for LDC2014T12, it contains 6,520 additional training AMR graphs, thus giving TSP-gen and SNRG-gen a noticeable advantage.
It is also important to note that the scores reported in \citet{song2016amr,song2017amr} were obtained after removing from the development and test sets all AMR graphs whose reference realizations have more than $l_\text{max} = 30$ words; this is especially relevant as longer AMR graphs are, generally speaking, more difficult to process. After removal of all AMR graphs with more than 30 words, our approach achieves scores of 28.3 and 28.9 on the development and test set, respectively, whereas TSP-gen achieves scores of 21.1 and 22.4 and SNRG-gen achieves scores of 25.2 and 25.6. 

Except for NEUR-gen, the above-mentioned generators all make use of language models trained on Gigaword; however, JAMR-gen, TSP-gen, SNRG-gen and PBMT-gen use 4- or 5-gram models trained on Gigaword~v5 whereas we consider only 3-grams and use Gigaword~v1. As higher-order $n$-grams can cope with more complex sentence structures and are thus more powerful than a 3-gram model, we believe that our approach would perform even better if we replaced our 3-gram model by some higher-order model. Unfortunately, we are not able to verify this claim as neither Gigaword nor higher-order $n$-gram models trained on it are available free of charge; we thus have to resort to a freely available 3-gram language model trained on Gigaword~v1.

The NEUR-gen system does not include a language model at all; instead, sentences from Gigaword~v5 (LDC2011T07) are annotated with AMR graphs using the text-to-AMR parser described in \citet{konstas2017neural} and directly embedded into the system as additional training data (see Section~\ref{RELATEDWORK}). However, only such sentences from Gigaword are used which contain exclusively words that also occur in LDC2014T12. To obtain the Bleu score of 29.7 on the LDC2014T12 test set, \citet{konstas2017neural} use two million such sentences, increasing the number of training data by a factor of roughly 153. Although many of the automatically generated AMR graphs are likely to contain at least some errors, it is reasonable to assume that the improvement in Bleu score compared to other approaches is mainly due to this enormous enlargement of the training corpus. This claim is supported by the fact that using the LDC2015E86 corpus, the test set results reported by \citet{konstas2017neural} lie between 22.0, when only the AMR graphs from LDC2015E86 are used, and 33.8, when 20 million annotated sentences from Gigaword are factored into the training process. For LDC2014T12, \citet{konstas2017neural} unfortunately do not report the scores for the development set or for any number of included Gigaword sentences other than two million. Naturally, it would make sense to investigate whether including annotated sentences from Gigaword into the training process of our system leads to comparable improvements of our results. As mentioned above, however, Gigaword is not free of charge, making us unable to carry out this investigation. 

As another experiment, we evaluated our generator on several subsets of our development and test sets that contain only AMR graphs for which the number of tokens $l_\text{ref}$ in the reference realization lies within a certain interval. We chose the set of intervals 
\[
\{ [0,10], (10, 20], (20, 30], (30, 40], (40, \infty) \}
\] 
and computed the Bleu score and the average time required to process a single graph for each interval.\footnote{The time measurements do not include the time required to load the language model and all required maximum entropy models into memory.} The results can be seen in Figure~\ref{fig:amr-bleu}~and~\ref{fig:amr-time}; Figure~\ref{fig:amr-number} lists the number of graphs in the LDC2014T12 corpus for each of the above intervals.

\pgfplotsset{compat=1.5}
\captionsetup[subfigure]{oneside,margin={0.1cm,0cm}}

\begin{figure}[]
\centering
\subfloat[Case-insensitive $1,\ldots,4$-gram Bleu score achieved by our generator on the development and test set when only AMR graphs with reference realization lengths $l_\text{ref}$ in the given intervals are considered]{
	\begin{tikzpicture}[trim axis left, trim axis right]
		\begin{axis}[
				 	legend style={draw=none, font=\footnotesize},
				 	scale only axis, 
				 	height=0.25\textheight,
					width=0.7\textwidth,
					xtick pos=left,
					ytick pos=left,
					ymin=22, ymax=31, xmin=0.6, xmax=5.4,
					xlabel={Reference realization length $l_\text{ref}$},
					ylabel={Bleu score},
					xtick={1,2,3,4,5},
					xticklabels={$\leq\!10$, 11--20, 21--30, 31--40, $>\!40$},
					ytick={23,24,25,26,27,28,29,30}
					]
					\addplot[color=plot1, mark=otimes*,mark options=solid] table[x=length, y=bleu] {bleu-dev.csv};
					\addlegendentry{Dev};
					\addplot[color=plot2, mark=otimes*,mark options=solid, dashed] table[x=length, y=bleu] {bleu-test.csv};
					\addlegendentry{Test};
				\end{axis}
	\end{tikzpicture}
	\label{fig:amr-bleu}
	}
	\\[0.5cm]
	
	\subfloat[Average time required to generate a sentence from a single AMR graph in the development and test set when only AMR graphs with reference realization lengths $l_\text{ref}$ in the given intervals are considered.]
	{
	\begin{tikzpicture}[trim axis left, trim axis right]
		\begin{axis}[
		 	legend style={draw=none, font=\footnotesize},
		 	scale only axis, 
		 	height=0.25\textheight,
			width=0.7\textwidth,
			xtick pos=left,
			ytick pos=left,
			ymin=-0.05, ymax=0.95, xmin=0.6, xmax=5.4,
			xlabel={Reference realization length $l_\text{ref}$},
			ylabel={$t$ in $s$},
			xtick={1,2,3,4,5},
			xticklabels={$\leq\!10$, 11--20, 21--30, 31--40, $>\!40$},
			ytick={0, 0.1, 0.2, 0.3, 0.4, 0.5, 0.6, 0.7, 0.8, 0.9}
			]
			\addplot[color=plot1, mark=otimes*,mark options=solid] table[x=length, y=time] {time-dev.csv};
			\addlegendentry{Dev};
			\addplot[color=plot2, mark=otimes*,mark options=solid, dashed] table[x=length, y=time] {time-test.csv};
			\addlegendentry{Test};	
		\end{axis}
	\end{tikzpicture}
	\label{fig:amr-time}
	}\\[0.5cm]
		\subfloat[Number of development and test AMR graphs for some values of $l_\text{ref}$]
		{
		\hspace{-2.7mm}
		\newcolumntype{Y}{>{\centering\arraybackslash}X}
			\begin{tabularx}{0.7\textwidth}{|lYYYYY|}
			\hline
			& \multicolumn{5}{c|}{Reference realization length $l_\text{ref}$} \\
			& $\leq\!10$ & 11--20 & 21--30 & 31--40 & $>\!40$ \\
			\hline
			Dev AMRs & 255 & 485 & 374 & 162 & 92 \\
			Test AMRs & 299 & 441 & 333 & 173 & 125 \\
			\hline
			\end{tabularx}
			\label{fig:amr-number}
		}
\caption{Performance of our transition-based generator when considering only AMR graphs for which the number $l_\text{ref}$ of tokens in the reference realization is within a certain interval}
\label{fig:diagram2}
\end{figure}

Not surprisingly, the processing of AMR graphs takes more time the longer the reference realizations are, with about 0.05s required for graphs with $l_\text{ref} \leq 10$ and up to 0.7s required for graphs with $l_\text{ref} > 40$. However, it is worth noting that our implementation is by no means optimized with respect to algorithmic efficiency. For example, the processing of large graphs could massively be improved through parallelization as for vertices $v$ and $v'$ with $v \notin \text{succ}(v')$ and $v' \notin \text{succ}(v)$, the sets $\text{best}(v)$ and $\text{best}(v')$ required by Algorithm~\ref{alg:generation} can be computed independently. 

With regards to the Bleu scores reported in Figure~\ref{fig:amr-bleu}, it is noteworthy that the results for $l_\text{ref} > 40$ are well below average, supporting our claim that a 4- or 5-gram language model might improve the Bleu score achieved by our generator as such higher order $n$-gram models are especially helpful for long sentences. Interestingly, however, the Bleu score of $22.8$ achieved on the test set for $l_\text{ref} \leq 10$ is even lower than for $l_\text{ref} > 40$. A qualitative analysis of all AMR graphs whose reference realizations have at most ten tokens shows that this low score is mainly due to wrongly guessed punctuation marks -- which can have a great impact on the Bleu score for sentences with relatively few words --, wrong date formats and errors made by our syntactic annotation models.
To illustrate this, consider the following examples, where for each $i \in \mathbb{N}$, $w_\text{r}^i$ denotes a reference realization provided in the LDC2014T12 test set and $w_\text{g}^i$ denotes the output of our generator for the corresponding AMR graph:\\[-2mm]

\bgroup
\def\arraystretch{1.5}
\noindent\begin{tabularx}{\textwidth}{rlrXr}
$\quad w_\text{r}^1 =$& 2004-12-19 & $\quad w_\text{r}^2 =$& a kathmandu police officer reports & \\
$w_\text{g}^1 =$& december 19 2004 & $\quad w_\text{g}^2 =$& a report by the kathmandu police officers . & \\
\end{tabularx}\\
\egroup

\noindent For $w_\text{r}^1$ and $w_\text{g}^1$, there are no matching $n$-grams at all; for $w_\text{r}^2$ and $w_\text{g}^2$, only three unigrams and one bigram match. Nonetheless, $w_\text{g}^1$ and $w_\text{g}^2$ are about equally good realizations of the corresponding AMR graphs as $w_\text{r}^1$ and $w_\text{r}^2$.

Our generator works best for AMR graphs whose reference realizations have between $11$ and $30$ tokens; for an example, consider the following pairs of reference realizations $w_\text{r}^i$ and outputs $w_\text{g}^i$:\\[-2mm]

\bgroup
\def\arraystretch{1.5}
\noindent\begin{tabularx}{\textwidth}{rXr}
$\quad w_\text{r}^3 =$& the story is based on the final report of the attorney general 's office . & \\
$w_\text{g}^3 =$& the story is based on the attorney general 's office final report . & \\
$w_\text{r}^4 =$& wen stated that the chinese government supports plans for peace in the middle east and remains firmly opposed to violent retaliation . & \\
$w_\text{g}^4 =$& wen stated that the chinese government supports the plan for peace in the middle east and remains in firm opposition to the violent retaliation~. & \\
\end{tabularx}\\
\egroup

\noindent However, if there are long range dependencies, our generator often fails to find syntactically correct realizations that transfer the meaning of the corresponding graphs. This is especially the case for AMR graphs with long reference realizations, as can be seen in the below example:\\[-2mm]

\bgroup
\def\arraystretch{1.5}
\noindent\begin{tabularx}{\textwidth}{rXr}
$\quad w_\text{r}^5 =$& the performance of the female competitors of the chinese diving team , mingxia fu and bin chi , in the first 6 rounds of the 10 - meter platform diving competition at the seventh world swimming championships held here today was ideal , and hopes of entering the heats are in sight . & \\
$w_\text{g}^5 =$& the ideal female competitors mingxia fu and bin chi of chinese diving team performance 6 first round of preliminary competition of the 10 meter platform diving at the seventh world swimming championships were held here today and hope to enter the heat is in sight . & \\
\end{tabularx}\\
\egroup

\noindent As a last experiment, we looked into the individual syntactic annotations and transitions used by our generator and investigated how well the prediction of these annotations and transitions works. In accordance with our generation algorithm, we discuss the results of this investigation separately for transitions from the set $T_\text{restr}$ and all remaining transitions.

For transitions contained within $T_\text{restr}$, the confusion matrix shown in Figure~\ref{tab:confusion-matrix} compares the transitions applied by our generator during the processing of all development AMR graphs of LDC2014T12 with the respective gold transitions. Each entry in a row with label $t_\text{a}$ and column with label $t_\text{g}$ denotes the number of times a transition of class $t_\text{a}$ was applied when the gold transition would have been in $t_\text{g}$; accordingly, diagonal entries correspond to correctly applied transitions. For example, 707 \textsc{Merge} transitions were applied correctly and 70 \textsc{Merge} transitions were applied when according to $\text{gold}_\mathcal{B}$, a \textsc{Keep} transition should have been applied. As can be concluded from Figure~\ref{tab:confusion-matrix}, \textsc{Swap} is by far the most error-prone transition for the first stage: It is only applied correctly in 75 cases whereas in 332 cases, a \textsc{Keep} transition is applied when a \textsc{Swap} transition would actually be required.

\newcolumntype{R}[2]{%
    >{\adjustbox{angle=#1,lap=\width-(#2)}\bgroup}%
    l%
    <{\egroup}%
}
\newcommand*\rot{\multicolumn{1}{R{30}{1em}}}

\begin{figure}
\centering
\begin{tabular}{rr|c|c|c|c|rr}
\multicolumn{8}{c}{\textbf{Gold Transition}} \\[0.2cm]
& \multicolumn{1}{c}{} & \rot{\textsc{Merge}} & \rot{\textsc{Swap}} & \rot{\textsc{Delete}} & \rot{\textsc{Keep}} & \multicolumn{1}{c}{} & \\
& &  &  &  &  & & \\[-0.41cm]
\hhline{~~----~~}
\multirow{4}{*}{\rotatebox[origin=c]{90}{\textbf{Applied Transition\ \ }}} &
\textsc{Merge} & \cellcolor{confgreen}\squarebox{707} & \cellcolor{confred!4}\squarebox{7} & \cellcolor{confred!5}\squarebox{11} & \cellcolor{confred!23}\squarebox{78} &\hphantom{\textsc{Merge}} & \multirow{4}{*}{\rotatebox[origin=c]{90}{\phantom{\textbf{Applied Transition}}}} \\
\hhline{~~|-|-|-|-|~~}
& \textsc{Swap} & \squarebox{0} & \cellcolor{confgreen}\squarebox{75} & \cellcolor{confred!2}\squarebox{2} & \cellcolor{confred!8}\squarebox{25} &\hphantom{\textsc{Swap}} & \\
\hhline{~~----~~}
& \textsc{Delete} &\cellcolor{confred!2}\squarebox{2} &\cellcolor{confred!3}\squarebox{4} & \cellcolor{confgreen}\squarebox{865} &\cellcolor{confred!27}\squarebox{90} &\hphantom{\textsc{Delete}} & \\
\hhline{~~----~~}
& \textsc{Keep} & \cellcolor{confred!24}\squarebox{81} & \cellcolor{confred}\squarebox{332} & \cellcolor{confred!70}\squarebox{233} &\cellcolor{confgreen}\squarebox{$\!$\footnotesize{13979}}&\hphantom{\textsc{Keep}} & \\
\hhline{~~----~~}
\end{tabular}
\caption{Confusion matrix for transitions performed in the first phase of our generation algorithm; \textsc{Delete-Reentrance} transitions are not included as they are always applied correctly.}
\label{tab:confusion-matrix}

\end{figure}

\begin{figure}[t]
\centering

\scalebox{0.8} {
\subfloat[][]{
\begin{tikzpicture}
\tikzstyle{amr-node}=[shape=ellipse,draw, inner sep=0.2, minimum height=0.8cm, minimum width=1.2cm, text height=1.5ex, text depth=.25ex]
\tikzstyle{amr-edge}=[text height=1.5ex, text depth=.25ex, fill=white]
\tikzstyle{text-node}=[text height=1.5ex, text depth=.25ex, align=left]

	\node (PAD2) at (-2.5,0){};
	\node (PAD2) at (2.5,0){};

    \node[amr-node] (amr-contain) at (0,0) {contain-01};
    \node[amr-node] (amr-minus) at (-1.5,-2) {$-$};
    \node[amr-node] (amr-ever) at (1.5,-2) {ever};

    \path [-latex](amr-contain) edge node[amr-edge] {polarity} (amr-minus);
    \path [-latex](amr-contain) edge node[amr-edge] {time} (amr-ever);
    
    \node[text-node] at(0,-3.6) {
    $w_r = $ never contained \\ $w_g = $ not ever contained
    };
    
\end{tikzpicture}
\label{fig:amr1-merge}
}
\subfloat[][]{
\begin{tikzpicture}
\tikzstyle{amr-node}=[shape=ellipse,draw, inner sep=0.2, minimum width=1.2cm, minimum height=0.8cm, text height=1.5ex, text depth=.25ex, minimum width=1.2cm]
\tikzstyle{amr-edge}=[text height=1.5ex, text depth=.25ex, fill=white]
\tikzstyle{text-node}=[text height=1.5ex, text depth=.25ex, align=left]

	\node (PAD2) at (-2.5,0){};
	\node (PAD2) at (2.5,0){};

    \node[amr-node] (amr-minister) at (0,0) {minister};
    \node[amr-node] (amr-vice) at (-1.5,-2) {vice};
    \node[amr-node] (amr-prime) at (1.5,-2) {prime};

    \path [-latex](amr-minister) edge node[amr-edge] {mod} (amr-vice);
    \path [-latex](amr-minister) edge node[amr-edge] {mod} (amr-prime);
    
    \node[text-node] at(0,-3.6) {
    $w_r = $ vice-prime minister \\ $w_g = $ vice prime minister
    };
    
\end{tikzpicture}
\label{fig:amr2-merge}
}
\subfloat[][]{
\begin{tikzpicture}
\tikzstyle{amr-node}=[shape=ellipse,draw, inner sep=0.2, minimum width=1.2cm, minimum height=0.8cm, text height=1.5ex, text depth=.25ex, minimum width=1.2cm]
\tikzstyle{amr-edge}=[text height=1.5ex, text depth=.25ex, fill=white]
\tikzstyle{text-node}=[text height=1.5ex, text depth=.25ex, align=left]

	\node (PAD2) at (-2.5,0){};
	\node (PAD2) at (2.5,0){};

    \node[amr-node] (amr-minister) at (0,0) {possible};
    \node[amr-node] (amr-vice) at (-1.5,-2) {$-$};
    \node[amr-node] (amr-prime) at (1.5,-2) {imagine-01};

    \path [-latex](amr-minister) edge node[amr-edge] {polarity} (amr-vice);
    \path [-latex](amr-minister) edge node[amr-edge, pos=0.52] {domain} (amr-prime);
    
    \node[text-node] at(0,-3.6) {$w_r = $ unimaginable \\ $w_g = $ can not imagine};
    
\end{tikzpicture}
\label{fig:amr3-merge}
}
}
\caption{Partial AMR graphs from LDC2014T12 requiring \textsc{Merge} transitions among neighbors. The corresponding reference realization $w_r$ and the output of our generator $w_g$ in the respective contexts is given below each partial graph.}
\label{fig:amrs-merge}
\end{figure}
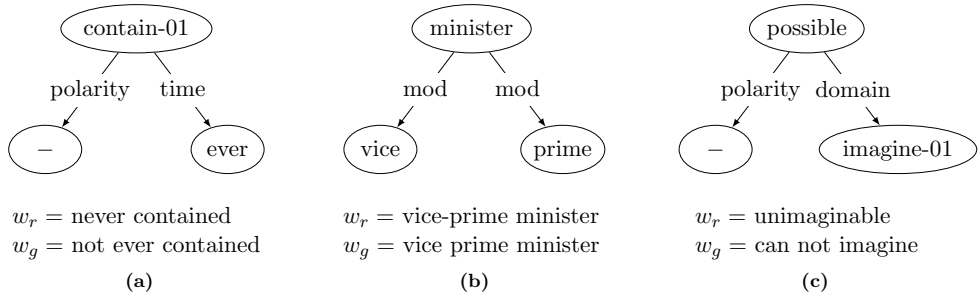

With regards to \textsc{Merge}, it is noteworthy that our definition of this transition -- which only allows merging nodes with their parents -- makes it impossible for our generator to transform several graphs into their reference realizations. This can be seen in the three exemplary partial AMR graphs from LDC2014T12 illustrated in Figure~\ref{fig:amrs-merge}: The graph in Figure~\ref{fig:amr1-merge} requires a \textsc{Merge} transition among the two neighboring nodes with labels ``$-$'' and ``ever'' to obtain the reference realization; similarly, merging the nodes with labels ``vice'' and ``prime'' is necessary for the graph shown in Figure~\ref{fig:amr2-merge}. Even more problematic is the graph illustrated in Figure~\ref{fig:amr3-merge}, which would require us to merge all three vertices simultaneously. These examples suggest that revising the definition of \textsc{Merge} transitions might be a way to improve the results obtained by our generator. 

We finally turn to an evaluation of the maximum entropy models used for syntactic annotations and all remaining transitions. Table~\ref{tab:mem-eval} shows the percentage of times in which the transition with the highest probability according to our models was in fact the gold transition to be applied, divided into three groups. The first of these groups, headed ``Reorderings'' in Table~\ref{tab:mem-eval}, lists the number of times the maximum entropy models $p_*$, $p_l$ and $p_r$ assigned the highest probability to the right order between two vertices. The group captioned ``Insertions'' lists the percentage of correctly predicted transitions in stages 3 and 5 of Figure~\ref{fig:transition-sequences} (Section~\ref{GENERATION:Decoding}). We recall that in stage~3, only \textsc{Insert-Child} and \textsc{No-Insertion} transitions can be applied whereas in stage 5, only \textsc{Insert-Between} and \textsc{No-Insertion} transitions are applicable. The last group, titled ``Realizations'', subsumes the results obtained by all syntactic annotation models $p_k$, $k \in \mathcal{K}_\text{syn}$ and the model $p_\textsc{Real}$ for \textsc{Realize} transitions. The vast majority of values shown in Table~\ref{tab:mem-eval} is above $80 \%$, indicating that in general, the features used to train our models are well-chosen. The percentage of correctly determined POS tags on both the development and test set is comparably low; however, as can be seen in the example outputs $w_g^2$ and $w_g^4$ shown before, this does not necessarily result in bad realizations.  

\begin{table}
\centering
\begin{tabular}[t]{l c c}
Reorderings & Dev & Test \\
\midrule
$p_*$ & 85.34\% & 83.90\% \\
$p_l$ & 84.38\% & 83.96\% \\
$p_r$ & 83.26\% & 78.11\% \\
& & \\
Insertions & Dev & Test \\
\midrule
$p_\text{TS}$ (Stage 3) & 86.32\% & 84.78\% \\
$p_\text{TS}$ (Stage 5) & 89.71\% & 89.55\% \\
\end{tabular}
\quad
\begin{tabular}[t]{l c c}
Realizations & Dev & Test \\
\midrule
$p_\anno{POS}$ & 76.58\% & 74.90\% \\[0.11cm]
$p_\anno{DENOM}$ & 80.61\% & 81.65\% \\[0.11cm]
$p_\anno{TENSE}$ & 74.79\% & 72.49\% \\[0.11cm]
$p_\anno{NUMBER}$ & 84.80\% & 86.00\% \\[0.11cm]
$p_\anno{VOICE}$ & 93.35\% & 93.84\% \\[0.11cm]
$p_\anno{REAL}$ & 82.28\% & 81.83\% \\[0.11cm]
\end{tabular}
\caption{Percentage of times in which the maximum entropy models used by our generator assign the highest probability to the correct outputs when processing the development and test sets of LDC2014T12. Situations in which the correct transition or annotation is uniquely determined through the transition constraints defined in Section~\ref{IMPLEMENTATION:TransitionConstraints} are excluded.}
\label{tab:mem-eval}

\end{table}

\clearpage


\section{Conclusion}
\label{CONCLUSION}

We have devised a novel approach for the challenging task of AMR-to-text generation. Our core idea was to turn input AMR graphs into ordered trees from which sentences can easily be inferred through application of the yield function.
We chose the principle component of our approach to be the
transition system $S_\text{AMR}$, whose set of transitions $T_\text{AMR}$ defines how the transformation from AMR graphs to suitable trees can be performed. Some transitions contained within this set, such as \textsc{Merge}, \textsc{Swap} and \textsc{Delete}, have an equivalent in the likewise transition-based text-to-AMR parser by~\cite{wang2015transition}, which served as a model for our approach. 

In order to turn $S_\text{AMR}$ into a generator, we assigned probabilities to transitions and defined the score of a transition sequence to be a linear combination of the probabilities of all its transitions and the probability assigned to the resulting sentence by a language model. We approximated these probabilities using maximum entropy models that were trained with a set of gold transitions extracted from a large corpus of AMR graphs and corresponding realizations. As an exhaustive search for the highest-scoring transition sequence given some input would be far too time-consuming, we developed an algorithm that approximates this sequence in two phases: In a first phase, only transitions from a subset $T_\text{restr}$ of $T_\text{AMR}$ are greedily applied without taking the language model into consideration; in a second phase, the output of this first phase is processed bottom-up, considering multiple partial transition sequences at each step and factoring in the language model. Through parametrized pruning, we restricted the number of sequences to be considered, allowing us to find a good balance between required time and quality of the generated sentences. We introduced the concepts of syntactic annotations and default realizations to help our system decide which transition to apply next. To further improve our results, we defined some postprocessing steps -- such as the insertion of punctuation marks -- to revise the tree structure obtained from our transition system. 

In experiments carried out using a Java-based implementation of our generator, we obtained a lower-cased $1\ldots 4$-gram Bleu score of 27.4 on the LDC2014T12 test set, the~second best result reported so far and the best without using parsed sentences from an external source such as Gigaword (LDC2011T07) as additional training data. This result strongly suggests that our transition-based transformation of AMR graphs into ordered tree structures is indeed quite a promising approach for the AMR-to-text generation~task. 

Throughout this work, we have highlighted a number of ways in which the results obtained by our system may further be improved upon.  
As outlined in Section~\ref{EXPERIMENTS}, one promising way that could easily be implemented, but would require access to Gigaword, would be to replace the used $3$-gram language model with some higher-order model. One could also follow the idea of \citet{konstas2017neural} and annotate Gigaword sentences with AMR graphs using a parser to augment the number of available training data; as pointed out in Section~\ref{EXPERIMENTS}, it is reasonable to assume that implementing this idea would have a major impact on the quality of our generator. 

Another possible modification shown to be promising in Section~\ref{EXPERIMENTS} is the redefinition of \textsc{Merge} transitions to allow for a merging of neighboring vertices. It is also conceivable to modify this transition in a way that allows for vertex groups of arbitrary size to be merged. In this context, one may also investigate whether the generator could further be tweaked by revising other classes of transitions. Of course, such a revision does not have to be limited to the formal definitions of the transitions themselves, but may also be extended to the extraction of gold transitions from a training corpus as done by the oracle algorithm introduced in Section~\ref{TRAINING:Transitions}.

While we have put plenty of effort into the selection of suitable features for the training of our maximum entropy models, one could of course also try to improve our generator's output by adding new features extracted from the given contexts. In addition, it should be investigated whether the conditional probability $P(t \mid c)$ of a transition $t$ given a configuration $c$ and the various conditional probabilities of syntactic annotations can be predicted more reliably by a model more powerful than maximum entropy models. In view of recent advances in AMR generation and parsing made with neural network architectures \citep[see][]{van2017neural,konstas2017neural}, especially probabilistic neural networks come to mind.

A further way to improve results may be to extend or revise the postprocessing steps introduced in Section~\ref{GENERATION:Postprocessing}. For instance, the assignment of punctuation marks could be refined -- or even be integrated into the actual transition system -- as the current output of punctuation marks by our generator shows some room for improvement, especially with respect to the placement of commas.  

Yet another possibility for enhancing the quality of our generator lies in editing the current implementation in order to make it more resource-friendly and time-efficient; as outlined in Section~\ref{EXPERIMENTS}, the latter could be achieved through parallelization. A~time-optimized implementation may also lead to better results in terms of Bleu score, as it would allow us to both drop some of the transition constraints introduced in Section~\ref{IMPLEMENTATION:TransitionConstraints} and increase the maximum values allowed for performance-relevant hyperparameters used by the best transition sequence algorithm.

Finally, it would also be interesting to investigate in how far our results are, as claimed in Section~\ref{INTRODUCTION}, in fact transferable to other languages. As indicated in Section~\ref{GENERATION:SyntacticAnnotation}, this would require us to revise the concept of syntactic annotations to properly reflect the linguistic peculiarities of the considered language. Unfortunately, however, such an investigation is not feasible at present, as no sufficiently large AMR corpus is available for any other language than English.

\clearpage

\bibliographystyle{apalike}
\bibliography{literatur}

\end{document}